%% file: LearningOptimizedRiskScores.tex
\documentclass[twoside,11pt]{article}
\usepackage{jmlr2e}

\usepackage{amsfonts,bbm,bm,courier}
\usepackage[scaled]{helvet}
\usepackage{textcase,moresize,etex,relsize,setspace,lastpage,needspace}

\usepackage{ifthen,xifthen,xpatch,afterpage}

\usepackage{array,tabularx,multirow,hhline,booktabs,colortbl}

\usepackage{xcolor,hyperref}

\usepackage{enumitem}

\usepackage{placeins,graphicx,grffile}
\usepackage[font=footnotesize,labelfont=bf,width=\textwidth]{caption}

\usepackage{amsmath,amssymb,eqnarray,mathtools}

\newlist{thmlist}{enumerate}{1}
\setlist[thmlist]{label=\raisebox{0.25ex}{\tiny$\bullet$}, topsep=0.5em,itemsep=2pt}

\usepackage{algorithm,algpseudocode}

\usepackage{amsthm}
\RequirePackage[framemethod=TikZ]{mdframed}

\newtheoremstyle{break}  % follow `plain` defaults but change HEADSPACE.
  {\topsep}   % ABOVESPACE
  {\topsep}   % BELOWSPACE
  {\itshape}  % BODYFONT
  {0pt}       % INDENT (empty value is the same as 0pt)
  {\bfseries} % HEADFONT
  {\newline}         % HEADPUNCT
  {0pt\needspace{2\baselineskip}}  % HEADSPACE. `plain` default: {5pt plus 1pt minus 1pt}
  {}          % CUSTOM-HEAD-SPEC

\newtheoremstyle{framed-and-break}  % follow `plain` defaults but change HEADSPACE.
  {0em}   % ABOVESPACE
  {0em}   % BELOWSPACE
  {\itshape}  % BODYFONT
  {0pt}       % INDENT (empty value is the same as 0pt)
  {\bfseries} % HEADFONT
  {\newline}         % HEADPUNCT
  {0pt\needspace{2\baselineskip}}  % HEADSPACE. `plain` default: {5pt plus 1pt minus 1pt}
  {}          % CUSTOM-HEAD-SPEC

\newtheoremstyle{framed-no-break}  % follow `plain` defaults but change HEADSPACE.
  {0em}   % ABOVESPACE
  {0em}   % BELOWSPACE
  {\itshape}  % BODYFONT
  {0pt}       % INDENT (empty value is the same as 0pt)
  {\bfseries} % HEADFONT
  {. }         % HEADPUNCT
  {5pt plus 1pt minus 1pt}%
  {}

\definecolor{thm-color}{RGB}{206, 225, 255}
\definecolor{corollary-color}{RGB}{206, 225, 255}
\definecolor{lemma-color}{RGB}{206, 225, 255}
\definecolor{proposition-color}{RGB}{206, 225, 255}
\definecolor{proof-color}{gray}{0.95}

\definecolor{definition-color}{RGB}{248, 222, 216}
\definecolor{assumption-color}{gray}{0.95}
\definecolor{remark-color}{RGB}{248, 238, 216}

\theoremstyle{framed-no-break}
\newtheorem*{boxproof}{Proof}

\theoremstyle{break}
\newtheorem{assumption}[theorem]{Assumption}

\theoremstyle{framed-and-break}
\newtheorem{boxtheorem}[theorem]{Theorem}
\newtheorem{boxcorollary}[theorem]{Corollary}
\newtheorem{boxproposition}[theorem]{Proposition}
\newtheorem{boxlemma}[theorem]{Lemma}
\newtheorem{boxremark}[theorem]{Remark}
\newtheorem{boxdefinition}[theorem]{Definition}
\newtheorem{boxexample}[theorem]{Example}
\newtheorem{boxassumption}[theorem]{Assumption}

\renewenvironment{definition}%
{%
\begin{mdframed}[%
backgroundcolor=definition-color,
linewidth=1pt,
innerleftmargin=0.5em,
innerrightmargin=0.5em,
innertopmargin=0.5em,
innerbottommargin=0.5em,
skipabove=0.5em, 
skipbelow=0.5em%
]%
\begin{boxdefinition}%
}%
{%
\end{boxdefinition}%
\end{mdframed}%
}

\renewenvironment{remark}%
{%
\begin{mdframed}[%
backgroundcolor=remark-color,
linewidth=1pt,
innerleftmargin=0.5em,
innerrightmargin=0.5em,
innertopmargin=0.25em,
innerbottommargin=0.25em,
skipabove=0.5em, 
skipbelow=0.5em%
]%
\begin{boxremark}}%
{\end{boxremark}%
\end{mdframed}}

{%
\begin{mdframed}[%
backgroundcolor=assumption-color,
linewidth=0pt,
innerleftmargin=0.5em,
innerrightmargin=0.5em,
innertopmargin=0.25em,
innerbottommargin=0.25em,
skipabove=0.25em, 
skipbelow=0.25em%
]%
\begin{boxassumption}%
{\end{boxassumption}%
\end{mdframed}}
}

{%
\begin{mdframed}[%
backgroundcolor=remark-color,
linewidth=1pt,
innerleftmargin=0.5em,
innerrightmargin=0.5em,
innertopmargin=0.5em,
innerbottommargin=0.5em,
skipabove=0.5em, 
skipbelow=0.5em%
]%
\begin{boxexample}%
}%
{%
\end{boxexample}%
\end{mdframed}%
}

\renewenvironment{proof}%
{%
\begin{mdframed}[%
backgroundcolor=proof-color,
linewidth=1pt,
innerleftmargin=0.5em,
innerrightmargin=0.5em,
innertopmargin=0.5em,
innerbottommargin=0.5em,
skipabove=0.5em, 
skipbelow=0.5em%
]%
\begin{boxproof}%
}%
{%
\hfill\rule{1.5ex}{1.5ex}\end{boxproof}%
\end{mdframed}%
}

\renewenvironment{theorem}%
{%
\begin{mdframed}[%
backgroundcolor=thm-color,
linewidth=1pt,
innerleftmargin=0.5em,
innerrightmargin=0.5em,
innertopmargin=0.5em,
innerbottommargin=0.5em,
skipabove=0.5em, 
skipbelow=0.5em%
]%
\begin{boxtheorem}%
}%
{%
\end{boxtheorem}%
\end{mdframed}%
}

\renewenvironment{proposition}%
{%
\begin{mdframed}[%
backgroundcolor=proposition-color,
linewidth=1pt,
innerleftmargin=0.5em,
innerrightmargin=0.5em,
innertopmargin=0.5em,
innerbottommargin=0.5em,
skipabove=0.5em, 
skipbelow=0.5em%
]%
\begin{boxproposition}%
}%
{%
\end{boxproposition}%
\end{mdframed}%
}

{%
\begin{mdframed}[%
backgroundcolor=lemma-color,
linewidth=1pt,
innerleftmargin=0.5em,
innerrightmargin=0.5em,
innertopmargin=0.5em,
innerbottommargin=0.5em,
skipabove=0.5em, 
skipbelow=0.5em%
]%
\begin{boxlemma}%
}%
{%
\end{boxlemma}%
\end{mdframed}%
}

\renewenvironment{corollary}%
{%
\begin{mdframed}[%
backgroundcolor=corollary-color,
linewidth=1pt,
innerleftmargin=0.5em,
innerrightmargin=0.5em,
innertopmargin=0.5em,
innerbottommargin=0.5em,
skipabove=0.5em, 
skipbelow=0.5em%
]%
\begin{boxcorollary}%
}%
{%
\end{boxcorollary}%
\end{mdframed}%
}

\input{preamble.tex}
\jmlrheading{20}{2019}{1-10}{9/18; Revised 4/19}{6/19}{18-615}{}
\ShortHeadings{Ustun and Rudin}{Learning Optimized Risk Scores}

\begin{document}

\title{Learning Optimized Risk Scores}

\author{%
\name Berk Ustun \email berk@seas.harvard.edu\\
\addr Center for Research in Computation and Society\\
Harvard University
\AND
\name Cynthia Rudin \email cynthia@cs.duke.edu\\
\addr Department of Computer Science\\
Department of Electrical and Computer Engineering\\
Department of Statistical Science\\
Duke University
}
\editor{Russ Greiner}
\maketitle

\begin{abstract}
Risk scores are simple classification models that let users make quick risk predictions by adding and subtracting a few small numbers.
These models are widely used in medicine and criminal justice, but are difficult to learn from data because they need to be calibrated, sparse, use small integer coefficients, and obey application-specific operational constraints.
In this paper, we present a new machine learning approach to learn risk scores.
We formulate the risk score problem as a mixed integer nonlinear program, and present a cutting plane algorithm for non-convex settings to efficiently recover its optimal solution.
We improve our algorithm with specialized techniques to generate feasible solutions, narrow the optimality gap, and reduce data-related computation.
Our approach can fit risk scores in a way that scales linearly in the number of samples, provides a certificate of optimality, and obeys real-world constraints without parameter tuning or post-processing.
We benchmark the performance benefits of this approach through an extensive set of numerical experiments, comparing to risk scores built using heuristic approaches.
We also discuss its practical benefits through a real-world application where we build a customized risk score for ICU seizure prediction in collaboration with the Massachusetts General Hospital.
\end{abstract}
\vspace{1em}
\begin{keywords}
scoring systems;
classification;
constraints; 
calibration;
interpretability;
cutting plane methods;
discrete optimization;
mixed integer nonlinear programming.
\end{keywords}

\section{Introduction}
\label{Sec::Introduction}

\emph{Risk scores} are linear classification models that let users assess risk by adding, subtracting, and multiplying a few small numbers (see Figure \ref{Fig::ExampleRiskScore}). These models are widely used to support decision-making in domains such as:
\begin{itemize}

\item \emph{Medicine}: to assess the risk of mortality in intensive care ~\citep[e.g.,][]{moreno2005saps}, critical physical conditions~\citep[e.g., adverse cardiac events,][]{six2008chest, than2014development} and mental illnesses \citep[e.g., adult ADHD in~][]{kessler2005world, ustun2016adhd}.

\item \emph{Criminal Justice}: to assess the risk of recidivism when setting bail, sentencing, and release on parole~\citep[see e.g.,][]{latessa2009creation,austin2010kentucky,psc2017final}. 

\item \emph{Finance}: to assess the risk of default on a loan \citep[see e.g., credit scores in][]{fico2011scorecard,siddiqi2012credit}, and to guide investment decisions~\citep[][]{piotroski2000value,beneish2013earnings}.
\end{itemize}

The adoption of risk scores in these areas stems from the fact that decision-makers often find them easy to use and understand. In comparison to other kinds of classification models, risk scores let users make quick predictions by simple arithmetic, without a computer or calculator. Users can gauge the effect of changing multiple input variables on the predicted outcome, and override predictions in an informed manner if needed. In comparison to scoring systems for decision-making~\citep[see e.g., the models considered in][]{ustun2016slim,ustun2015recidivism,carrizosa2013strongly,van2013risk,billiet2016interval,billiet2017interval,sokolovska2017fused,sokolovska2018provable}, which predict a yes-or-no outcome at a fixed operating point, risk scores output risk estimates at multiple operating points. Thus, users can choose an operating point while the model is deployed. Further, they are provided with risk estimates that -- if calibrated -- can inform this choice, and support decisions in other ways~\citep[see e.g.,][]{shah2018big}. We provide more background on risk scores in Appendix \ref{Appendix::RiskScoreBackground}.

\begin{figure}[htbp]
\centering
{\scoringsystem{}
\begin{tabular}{|l l  r | l |}
   \hline
1.   & \textssm{\textbf{C}ongestive Heart Failure} & 1 point & $\phantom{+}\prow{}$ \\ 
  2. & \textssm{\textbf{H}ypertension}  & 1 point & $+\prow{}$ \\ 
  3. & \textssm{\textbf{A}ge $\geq$ 75} & 1 point & $+\prow{}$ \\ 
  4. & \textssm{\textbf{D}iabetes Mellitus} & 1 point & $+\prow{}$ \\ 
  5. & \textssm{Prior \textbf{S}troke or Transient Ischemic Attack}  & \textbf{2} points & $+\phantom{\prow{}}$ \\[0.1em]
   \hline
 & \instruction{1}{5} & \scorelabel{} & $=\phantom{\prow{}}$ \\ 
   \hline
\end{tabular}
}

{%
\risktable{}
\begin{tabular}{|r|c|c|c|c|c|c|c|}
\hline \rowcolor{scorecolor}\scorelabel{} & 0 & 1 & 2 & 3 & 4 & 5 & 6 \\
\hline \rowcolor{riskcolor}\risklabel{} & 1.9\% & 2.8\% & 4.0\% & 5.9\% & 8.5\% & 12.5\% & 18.2\%\\ 
\hline
\end{tabular}
}
\caption{CHADS$_2$ risk score of~\citet{gage2001validation} to assess stroke risk (see \href{http://www.mdcalc.com}{www.mdcalc.com} for other medical scoring systems). The variables and points of this model were determined by a panel of experts, and the risk estimates were computed empirically from data. %The risk estimates in CHADS$_2$ increase with the score, but they are not necessarily calibrated.
}
\label{Fig::ExampleRiskScore}
\end{figure}

Although risk scores have existed for nearly a century \citep[see e.g.,][]{burgess1928factors}, many of these models are still built \emph{ad hoc}. This is partly because risk scores are often developed for applications where models must adhere to constraints related to interpretability and usability~\citep[see e.g., requirements on ``face validity" and ``user friendliness" in][]{than2014development}. Handling such constraints necessitates precise control over multiple elements of a model, from its choice of features %~\citep{kessler2005world}, 
to their relationship with the predicted outcome~\citep[e.g., monotonicity of the predictions with respect to the feature values, see][]{gupta2016monotonic}, 
to performance on specific subgroups~\citep{feldman2015certifying,pleiss2017fairness}. Since existing classification methods do not provide control over all these elements, risk scores are typically built using heuristics and expert judgment~\citep[e.g., preliminary feature selection, followed by logistic regression on the chosen features, scaling, and rounding as outlined by][]{antman2000timi}. In some cases, risk scores are hand-crafted by a panel of experts~\citep[see e.g., the CHADS$_2$ score in Figure \ref{Fig::ExampleRiskScore}, or the National Early Warning Score of][]{mcginley2012national}. As we will show, such ad hoc approaches may produce a model that violates important requirements, or that performs poorly relative to the best risk score that can be built using the same dataset. In such cases, the lack of a formal guarantee further complicates model development: when a risk score performs poorly, one cannot tell if this is due to the use of heuristics, or due to overly restrictive constraints.

In this paper, we present a new machine learning approach to learn risk scores from data. Our approach aims to train risk scores in a single-shot procedure -- by solving a mixed-integer nonlinear program (MINLP), which minimizes the logistic loss for calibration and AUC, penalizes the $\ell_0$-norm for sparsity, and restricts coefficients to small integers. We refer to this optimization problem as the \emph{risk score problem}, and refer to the risk score built from its solution as a \emph{Risk-calibrated Supersparse Linear Integer Model} (\RiskSLIM{}). The same term is used for our algorithmic framework for building these models. We aim to recover a certifiably optimal solution (i.e., a global optimum \emph{and} a certificate of optimality). This requires solving a difficult optimization problem, but has three major benefits for our setting:
\begin{romanlist}

\item \emph{Performance}: Since the MINLP directly penalizes and constrains discrete quantities, it can produce a risk score that is fully optimized for feature selection and small integer coefficients, and that obeys all application-specific requirements. Thus, models will not suffer in training performance due to the use of heuristics or post-processing.

\item \emph{Direct Customization}: Practitioners can address application-specific requirements by adding discrete constraints to the MINLP formulation, which can be solved with a generic solver (that is called by our algorithm as a subroutine). In this way, they can customize risk scores without parameter tuning, post-processing, or implementing a new algorithm.

\item \emph{Evaluating the Impact of Constraints}: Our approach pairs risk scores with a certificate of optimality. By definition, a certifiably optimal solution to the risk score problem attains the best performance among risk scores that satisfy a particular set of constraints. Once we recover a certifiably optimal solution, we therefore end up with a risk score with acceptable performance, or a risk score with unacceptable performance \emph{and} a certificate proving that the constraints were too restrictive. By comparing certifiably optimal risk scores for different sets of constraints, we can make informed choices between models that obey different sets of requirements.
\end{romanlist} 

Considering these potential benefits, a key goal of this work is to recover certifiably optimal solutions to the risk score problem for the largest possible datasets. As we will show, solving the risk score problem with a commercial MINLP solver is time-consuming even on small datasets, as generic MINLP algorithms are slowed down by excessive data-related computation. Accordingly, we aim to solve the risk score problem with a \emph{cutting plane algorithm}, which reduces data-related computation by iteratively solving a surrogate problem with a linear approximation of the loss function that is much cheaper to evaluate.
Cutting plane algorithms have an impressive track record on large supervised learning problems, as they scale linearly with the number of samples and provide precise control over data-related computation~\cite[see e.g.,][]{teo2009bundle, franc2009optimized, joachims2009cutting}. 

However, prior cutting plane algorithms were designed under the assumption that the surrogate problem can be solved to optimality at each iteration.
This assumption, which is perfectly reasonable in a convex setting, leads cutting plane algorithms to \emph{stall} on non-convex problems, as the time to solve the surrogate to optimality increases exponentially with each iteration. To overcome this issue, we present a new cutting plane algorithm for non-convex settings. We then improve its performance with specialized techniques to generate feasible solutions, narrow the optimality gap, and reduce data-related computation. Our approach extends the benefits of cutting plane algorithms to discrete settings, allowing us to efficiently train optimized risk scores for many problems of interest.

\newpage
\paragraph{Contributions}

The main contributions of this paper are as follows.
\begin{itemize}

\item  We present a new machine learning approach to build risk scores. Our approach can train models that: (i) are fully optimized for feature selection and small integer coefficients; (ii) handle application-specific constraints without parameter tuning or post-processing; (iii) provide a certificate of optimality.

\item We develop a new cutting plane algorithm, called the \emph{lattice cutting plane algorithm} (\LCPA{}). \LCPA{} retains the benefits of cutting plane algorithms for convex empirical risk minimization problems, but does not stall on problems with non-convex regularizers or constraints. It can be easily implemented using a MIP solver (e.g., CPLEX, CBC), and can fit customized risk scores in a way that scales linearly with the number of samples in a dataset.

\item We design techniques that allow \LCPA{} to quickly recover a risk score with good performance and a small optimality gap. These include rounding and polishing heuristics, fast bound-tightening and initialization procedures, and strategies to reduce data-related computation. 

\item We present an extensive set of experiments comparing methods to learn risk scores on publicly available datasets. Our results show that our approach can consistently train risk scores with best-in-class performance in minutes. We highlight pitfalls of approaches that are often used in practice, and present new heuristic methods that address these issues to improve the common approaches.

\item We present results from a collaboration with the Massachusetts General Hospital where we built a customized risk score for ICU seizure prediction. Our results highlight the practical benefits of our approach when training models that obey real-world constraints, and illustrate the performance gains of certifiably optimal risk scores in such settings.

\item We provide a software package to build optimized risk scores in Python, available online at \hyperlink{http://github.com/ustunb/risk-slim}{http://github.com/ustunb/risk-slim}.

\end{itemize}

\paragraph{Organization}

In the remainder of Section \ref{Sec::Introduction}, we discuss related work.
In Section \ref{Sec::ProblemStatement}, we formally define the risk score problem.
In Section \ref{Sec::Methodology}, we present our cutting plane algorithm.
In Section \ref{Sec::AlgorithmicImprovements}, we describe techniques to improve it.
In Section \ref{Sec::Experiments}, we benchmark methods to build risk scores.
In Section \ref{Sec::SeizurePrediction}, we discuss an application to ICU seizure prediction.

The supplement to our paper contains:
proofs of all theorems (Appendix \ref{Appendix::Proofs});
a primer on how risk scores are developed in practice (Appendix \ref{Appendix::RiskScoreBackground});
additional algorithmic improvements (Appendix \ref{Appendix::AlgorithmicImprovements});
supporting material for the experiments in Sections \ref{Sec::Methodology} and \ref{Sec::AlgorithmicImprovements} (Appendix \ref{Appendix::SimulationDetails}); the performance benchmark in Section \ref{Sec::Experiments} (Appendix \ref{Appendix::Experiments}); and the seizure prediction application in Section \ref{Sec::SeizurePrediction} (Appendix \ref{Appendix::SeizurePrediction}).

\paragraph{Prior Work}

Our paper extends work that was first published in KDD \citep{ustun2016kdd}. Real-world applications of \RiskSLIM{} include building a screening tool for adult ADHD from a short self-reported questionnaire \citep{ustun2016adhd}, and building a risk score for ICU seizure prediction \citep{struck2017association}. Applications of this work have been discussed in a paper that was a finalist for the 2017 INFORMS Daniel H. Wagner Prize \citep{RudinUs18}, and the application to seizure prediction \citep{ustun2016adhd} was awarded the 2019 INFORMS Innovative Applications in Analytics Award.

\subsection{Related Work}
\label{Sec::RelatedWork}

\paragraph{Scoring Systems}

While several methods have been proposed to learn scoring systems for \emph{decision-making} \citep[see, e.g.,][]{ustun2016slim,carrizosa2013strongly,van2013risk,billiet2016interval,billiet2017interval,sokolovska2017fused,sokolovska2018provable}, this work aims to learn scoring systems for \emph{risk assessment} (i.e., risk scores). Risk scores represent the majority of scoring systems that are currently used in medicine and criminal justice. These models are primarily designed to output calibrated risk estimates~\citep[see e.g., Section \ref{Sec::SeizurePrediction} and][for a discussion on how miscalibrated risk estimates can lead to harmful decisions in medicine]{van2015calibration,alba2017discrimination}. As we will show in Section \ref{Sec::ExperimentalResults}, building risk scores that output calibrated risk estimates is challenging, and common heuristics used in risk score development (e.g., rounding, scaling) can undermine calibration in ways that are difficult to repair.

\RiskSLIM{} risk scores are the risk assessment counterpart to \SLIM{} scoring systems \citep{ustun2013supersparse,ustun2016slim}, which have been applied to problems such as sleep apnea screening \citep{ustun2016clinical}, Alzheimer's diagnosis \citep{Souillard15}, and recidivism prediction \citep{ustun2015recidivism,RudinWaCo19}. %The models from \SLIM{} are often as accurate as models from complex machine learning methods  \citep[see also][]{RudinWaCo19}. 
Both \RiskSLIM{} and \SLIM{} models are optimized for feature selection and small integer coefficients, and can be directly customized to obey application-specific constraints. \RiskSLIM{} models are designed for risk assessment and optimize the logistic loss. In contrast, \SLIM{} models are designed for decision-making and minimize the 0--1 loss. \SLIM{} models do not output probability estimates, and the scores will not necessarily have high AUC. However, they will perform better at the operating point on the ROC curve for which they were optimized. Optimizing the 0--1 loss is also NP-hard, so training \SLIM{} models with standard solvers may not scale to datasets with large sample sizes as is the case here. In practice, \RiskSLIM{} is better-suited for applications where models must output calibrated risk estimates and/or perform well at multiple operating points along the ROC curve.

A predecessor to \RiskSLIM{} is the work of \citet{ertekin2015bayesian} which uses a Bayesian approach. While this approach is not able to prove optimality of solutions, it uses bounds to limit the search space.

\paragraph{Machine Learning}

The cutting-plane algorithm  in this work can be adapted to empirical risk minimization problems with a convex loss function, a non-convex penalty, and non-convex constraints. Such problems can be solved to train a large class of machine learning models, including: scoring systems for decision-making  \citep{carrizosa2013strongly,van2013risk,billiet2016interval,billiet2017interval,sokolovska2017fused}; sparse rule-based models such as decision lists \citep[see e.g,][]{letham2013interpretable,angelino2017learning}, k-of-n rules \citep[see e.g.,][]{chevaleyre2013rounding} and other Boolean functions \citep[see e.g.,][]{malioutov2013exact,wang2015bayesian,lakkaraju2016interpretable}; and other $\ell_0$-regularized models \citep[][]{sato2015piecewise,sato2016feature,bertsimas2015best}. For each of these model types, our cutting-plane algorithm can train models that optimize the same objective function and obey the same constraints, but in a way that recovers a globally optimal solution, handles application-specific constraints, and scales linearly with the number of samples.

Our work highlights an alternative approach to build models that obey constraints related to, for example, \emph{interpretability} \citep[see e.g.][where interpretability is addressed through constraints on model form]{caruana2015intelligible,gupta2016monotonic,Rudin19,ChenEtAlFICO2018,LiEtAl18} \emph{safety} \citep{amodei2016concrete}, \emph{credibility} \citep{wang2017creating}, and \emph{parity} \citep{kamishima2011fairness,zafar2015fairness}. Such qualities depend on multiple model properties, which vary significantly across applications and present unknown performance trade-offs. Existing approaches often aim to address specific types of constraints for generic models by pre-processing or post-processing \cite[see e.g.,][]{goh2016satisfying, calmon2017optimized,wang2019repairing}. In contrast, our approach aims to address such constraints directly for a specific model class. When these models belong to a simple hypothesis class (e.g., risk scores), we can expect model performance on training data to generalize, and we can evaluate this empirically (e.g., using cross-validation). In this way, one can assess the impact of constraints on predictive performance and make informed choices between models.

Our work is part of a broader stream of research on integer programming and other discrete optimization methods in supervised learning~\citep[e.g.,][]{carrizosa2013strongly,liu2007variable,GoldbergEc2012,guan2009mixed,nguyen2012general,sato2015piecewise,sato2016feature,ErtekinRu18,bertsimas2015best,lakkaraju2016interpretable,angelino2017learning,ChenRu2018,ChangEtAl2012,verwer2019learning,HuRuSe18,RuWa18,goh2014box,ustun2019fairness}.
%
%A common thread throughout these works (and many others) suggest that certifiably optimal models not only perform better, but are useful for applications where models must satisfy constraints (see e.g., Section \ref{Sec::SeizurePrediction}).
%
A unique aspect of this work is that we recover models that are certifiably optimal or have small optimality gaps \citep[see also][]{ustun2016slim,angelino2017learning}. Our results suggest that certifiably optimal models not only perform better, but are useful for applications where models must satisfy constraints (see e.g., Section \ref{Sec::SeizurePrediction}).

\paragraph{Optimization}

We train risk scores by solving a MINLP with three main components: (i) a convex loss function;  (ii) a non-convex feasible region (i.e., small integer coefficients and application-specific constraints); (iii) a non-convex penalty function (i.e., the $\ell_0$-penalty).

In Section \ref{Sec::MethodComparison}, we show that this MINLP requires a specialized algorithm because off-the-shelf MINLP solvers fail to solve instances for small datasets. We propose solving the risk score problem with a cutting plane algorithm. Cutting planes have been extensively studied by the optimization community~\citep[see e.g.,][]{kelley1960cutting} and applied to solve \emph{convex} empirical risk minimization problems ~\citep{teo2007scalable,teo2009bundle,franc2008optimized,franc2009optimized,joachims2006training,joachims2009cutting}.

Our cutting plane algorithm (the Lattice Cutting Plane Algorithm -- \LCPA{}) builds a cutting plane approximation while performing branch-and-bound search. It can be easily implemented using a MIP solver with \emph{control callbacks}~\citep[see e.g.,][for similar uses of control callbacks]{Bai2009,NaoumSawaya:2010ky}. \LCPA{} retains the key benefits of existing cutting plane algorithms on empirical risk minimization problems, but does not stall on problems with non-convex regularizers or constraints. As we discuss in Section \ref{Sec::CPAStalling}, stalling affects many cutting plane algorithms, including variants that are not considered in machine learning \citep[see][for a list]{boyd2004convex}. \LCPA{} is similar to recent outer-approximation algorithms that been developed for convex MINLP problems \citep[see e.g.,][]{lubin2018polyhedral}, which have also been shown to outperform generic MINLP algorithms \citep{kronqvist2019review}.

\clearpage
\section{Risk Score Problem}
\label{Sec::ProblemStatement}

In what follows, we formalize the problem of learning a risk score, of the same form as the model in Figure \ref{Fig::ExampleRiskScore}. We start with a dataset of $\n$ i.i.d$.$ training examples $(\xb_i,y_i)_{i=1}^\n$ where $\xb_i \subseteq \R^{d+1}$ denotes a vector of features $[1, x_{i,1},\ldots,x_{i,d}]^\top$ and $y_i \in \{\pm 1\}$ denotes a class label. We represent the score as a linear function $s(\xb) = \dotprod{\lambdab}{\xb}$ where $\lambdab \subseteq \R^{d+1}$ is a vector of $d + 1$ coefficients $[\lambda_0, \lambda_1,\ldots,\lambda_d]^\top$, and $\lambda_0$ is an intercept. In this setup, coefficient $\lambda_j$ represents the points that feature $j$ contributes to the score. Given an example with features $\xb_i$, a user tallies the points to compute a score $s_i = \dotprod{\lambdab}{\xb_i}$, and then converts the score into an estimate of predicted risk. We estimate the \emph{predicted risk} that example $i$ is positive through the logistic link function\footnote{Other risk models can be used as well, so long as they produce a concave log-likelihood.} as:$$p_i = \prob{y_i = +1 ~|~ \xb_i} = \logprob{\lambdab}.$$

\subsubsection*{Model Desiderata}

Our goal is to train a risk score that is sparse, has small integer coefficients, and performs well in terms of the following measures:
\begin{enumerate}[leftmargin=*]

\item \textbf{Calibration}: A calibrated model outputs risk predictions that match their observed risks. We assess the calibration of a model using a \emph{reliability diagram}  \citep[see][]{degroot1983comparison}, which shows how the \emph{predicted risk} (x-axis) at each score matches the \emph{observed risk} (y-axis). We estimate the observed risk for a score of $s$ as $$\bar{p}_s = \frac{1}{|\{i: s_i = s\}|} \sum_{i: s_i = s}{\indic{y_i = +1}}.$$ We summarize the calibration of a model over the full reliability diagram using the \emph{expected calibration error}  \citep{naeini2014binary}: $$\textrm{CAL} = \frac{1}{\n{}} \sum_{s}\sum_{i: s_i = s} |p_i - \bar{p}_s|.$$ 

\item \textbf{Rank Accuracy}: A rank-accurate model outputs scores that can correctly rank examples according to their true risk. We assess the rank accuracy of a model using the \emph{area under the ROC curve}: $$\textrm{AUC} = \frac{1}{\nplus{}\nminus{}} \sum_{[i:y_i = +1]}\sum_{[k: y_k = -1]} \indic{s_i > s_k},$$ where $\nplus{} = |\{i: y_i = +1\}|$ and $\nminus{} = |\{i: y_i = -1\}|$. 

\end{enumerate}

As discussed in Section \ref{Sec::RelatedWork}, calibration is the primary performance objective when building a risk score. In principle, good calibration should ensure good rank accuracy. Nevertheless, we report AUC as an auxiliary performance metric because trivial risk scores (i.e., models that assign the same score to all examples) can have low CAL on datasets with class imbalance (see Section \ref{Sec::ExperimentalResults} for an example).

We determine the values of the coefficients by solving a mixed integer nonlinear program (MINLP), which we refer to as the \emph{risk score problem} or \RSMINLP{}.
\begin{definition}[Risk Score Problem, \RSMINLP{}]
\label{Def::RiskScoreProblem}
The risk score problem is a discrete optimization problem with the form:
\begin{align}
\label{Eq::RiskScoreOptimization}
\begin{split}
\min_{\lambdab} & \qquad \lossfun{\lambdab} + C_0 \zeronorm{\lambdab} \\ 
\st &  \qquad \lambdab \in \Lset,
\end{split}
\end{align}
where:
\begin{thmlist}
\item $\lossfun{\lambdab} = \logloss{\lambdab}$ is the normalized logistic loss function;
\item $\zeronorm{\lambdab} = \sum_{j=1}^d{\indic{\lambda_j\neq0}}$ is the $\ell_0$-seminorm;
\item $\Lset \subset \Z^{d+1}$ is a set of feasible coefficient vectors (user-provided);
\item $C_0 > 0$ is a trade-off parameter to balance fit and sparsity (user-provided).
\end{thmlist}
\end{definition}
\RSMINLP{} captures what we desire in a risk score. The objective minimizes the \emph{logistic loss} for calibration and AUC, and penalizes the $\ell_0$-seminorm (the count of non-zero coefficients) for sparsity. The trade-off parameter $C_0$ controls the balance between these competing objectives, and represents the maximum log-likelihood that is sacrificed to remove a feature from the optimal model. The constraints restrict coefficients to a set of small integers such as $\Lset = \{-5,\ldots,5\}^{d+1}$, and may be customized to encode other model requirements such as those in Table~\ref{Table::OperationalConstraints}.

\begin{table}[htbp]
\tableformat{}
\begin{tabular}{ll}
\toprule
\textbf{Model Requirement} & \textbf{Example} \\
\toprule Feature Selection & Choose between 5 to 10 total features \\
\midrule Group Sparsity & Include either $male$ or $female$ in the model but not both\\
\midrule Optimal Thresholding & Use at most 3 thresholds for a set of indicator variables: $\sum_{k=1}^{100} \indic{age \leq k} \leq 3 $\\
\midrule Logical Structure & If $male$ is in model, then include $hypertension$ or $bmi \geq 30$\\
\midrule Side Information & Predict $\prob{y = + 1 | \xb} \geq 0.90$ when $male = \text{TRUE}$ and $hypertension = \text{TRUE}$\\ 
\bottomrule
\end{tabular}
\caption{Model requirements that can be addressed by adding operational constraints to \RSMINLP.} 
\label{Table::OperationalConstraints}
\end{table}

A \emph{Risk-calibrated Supersparse Linear Integer Model} (\RiskSLIM{}) is a risk score that is an optimal solution to \eqref{Eq::RiskScoreOptimization}. By definition, the optimal solution to \RSMINLP{} attains the lowest value of the logistic loss among feasible models on the training data, provided that $C_0$ is small enough (see Appendix \ref{Appendix::SmallRegularizationParameter} for a proof). Thus, a \RiskSLIM{} risk score is a maximum likelihood logit model that satisfies all required constraints.

Our experiments in Section \ref{Sec::Experiments} show that models with lower loss typically attain better calibration and AUC on the training data \citep[see also][]{caruana2004data}, and that this generalizes to test data due to the simplicity of our hypothesis space. There are some theoretical results to explain why minimizing the logistic loss leads to good calibration and AUC. In particular, the logistic loss is a strictly proper loss \citep{reid2010composite,ertekin2011equivalence} which yields calibrated risk estimates under the parametric assumption that the true risk can be modeled using a logistic link function \citep[see][]{menon2012predicting}. Further, the work of~\citet{kotlowski2011bipartite} shows that a ``balanced" version of the logistic loss forms a lower bound on $1 -$AUC, so minimizing the logistic loss indirectly maximizes a surrogate of AUC.

\paragraph{Trade-off Parameter}

The trade-off parameter can be restricted to values between $C_0 \in [0, \lossfun{\bm{0}}]$. Setting $C_0 > \lossfun{\bm{0}}$ will produce a trivial model where $\lambdab^*= \bm{0}$. Using an exact formulation provides an alternative way to set the trade-off parameter $C_0$:

\begin{itemize}[leftmargin=*,itemsep=1pt]

\item If we are given a limit on model size (e.g., $\zeronorm{\lambdab} \leq R$), we can add it as a constraint in the formulation, and set $C_0$ to a small value such as $C_0 = 10^{-8}$. In this case, the optimal solution to \RSMINLP{} corresponds to the model minimizing the logistic loss that obeys the model size constraint, provided that $C_0$ is small enough (see Appendix~\ref{Appendix::SmallRegularizationParameter}).

\item If we wish to set the model size in a data-driven manner (e.g., to optimize a measure of cross-validated performance), we can solve several instances of \RSMINLP{} with a model size constraint $\zeronorm{\lambdab} \leq R$, where we fix $C_0$ to a small value and vary the model size limit from $R = 1$ to $R = d$. This approach produces the best models over the full $\ell_0$-regularization path after solving $d$ instances of \RSMINLP{}. In comparison, a standard approach (i.e., where we treat $C_0$ as a hyperparameter and solve an instance of \RSMINLP{} without a model size constraint for different values of $C_0$) requires solving at least $d$ instances, since we cannot determine (in advance) $d$ values of $C_0$ that produce the full range of risk scores.

\end{itemize}

\paragraph{Computational Complexity}

Optimizing \RSMINLP{} is a difficult computational task given that $\ell_0$-regularization, minimization over integers, and MINLP problems are all NP-hard~\citep{bonami2012algorithms}. These are worst-case complexity results that mean that finding an optimal solution to \RSMINLP{} may be intractable for high dimensional datasets. As we will show, however, \RSMINLP{} can be solved to optimality for many real-world datasets in minutes, and in a way that scales linearly in the sample size. 

\paragraph{Notation, Assumptions, and Terminology}

We let $\objvalfun{\lambdab} = \lossfun{\lambdab} + C_0 \zeronorm{\lambdab}$ denote the objective function of \RSMINLP{}, and let $\lambdab^\opt \in \argmin_{\lambdab \in \Lset} \objvalfun{\lambdab}$ denote an optimal solution. We bound the optimal values of the objective, loss, and $\ell_0$-seminorm as $\objvalfun{\lambdab^\opt} \in [\objvalmin, \objvalmax]$, $\lossfun{\lambdab^\opt} \in [\lossmin, \lossmax]$, $\zeronorm{\lambdab^\opt} \in [\zeronormmin, \zeronormmax]$, respectively. We denote the set of feasible coefficients for feature $j$ as $\Lset_j$, and let $\minlambda{j} = \min_{\lambda_j\in\Lset_j} \lambda_j$ and $\maxlambda{j} = \max_{\lambda_j\in\Lset_j} \lambda_j$. 

For clarity of exposition, we assume that: (i) the coefficient set contains the null vector, $\bf{0} \in \Lset$, which ensures that \RSMINLP{} is always feasible; (ii) the intercept is not regularized, which means that the more precise version of the \RSMINLP{} objective function is $\objval{\lambdab} = \lossfun{\lambdab} + C_0 \zeronorm{\lambdab_{[1,d]}}$ where $\lambdab = [\lambda_0, \lambdab_{[1,d]}]$.

We measure the optimality of a feasible solution $\lambdab' \in \Lset$ in terms of its \emph{optimality gap}, defined as $\frac{\objvalfun{\lambdab'} - \objvalmin}{\objvalfun{\lambdab'}}$. Given an algorithm to solve \RSMINLP{}, we denote the best feasible solution that the algorithm returns in a fixed time as $\lambdab^\textrm{best} \in \Lset.$ The optimality gap of $\lambdab^\textrm{best}$ is computed using an upper bound set as $\objvalmax{} = \objvalfun{\lambdab^\textrm{best}}$, and a lower bound $\objvalmin{}$ that is provided by the algorithm. We say that the algorithm has solved \RSMINLP{} to \emph{optimality} if $\lambdab^\textrm{best}$ has an optimality gap of $\varepsilon = 0.0\%$. This implies that it has found a best feasible solution to \RSMINLP{} and produced a lower bound $\objvalmin{} = \objval{\lambdab^\opt}$. 

\clearpage
\section{Methodology}
\label{Sec::Methodology}
\input{section_03_methodology}

\clearpage
\section{Algorithmic Improvements}
\label{Sec::AlgorithmicImprovements}
\input{section_04_algorithmic_improvements.tex}

\clearpage
\section{Experiments}
\label{Sec::Experiments}
\input{section_05_experiments.tex}

\clearpage
\section{ICU Seizure Prediction}
\label{Sec::SeizurePrediction}
\input{section_06_seizure_prediction.tex}

\clearpage
\section{Concluding Remarks}
\label{Sec::Discussion}
\input{section_07_discussion.tex}

\section*{Acknowledgments}

We gratefully acknowledge support from Siemens, Phillips, and Wistron. We would like to thank Paul Rubin for helpful discussions, and our collaborators Aaron Struck and Brandon Westover for their guidance on the seizure prediction application.

{\small
\bibliographystyle{plainnat}
\bibliography{risk_scores}
}

\clearpage

\pagenumbering{gobble}
\appendix
\section{Omitted Proofs}
\label{Appendix::Proofs}
\input{appendix_proofs.tex}
\clearpage
\section{Small Regularization Parameters do not Influence Accuracy}
\label{Appendix::SmallRegularizationParameter}
\input{appendix_tradeoff_parameter}

\clearpage
\section{Background on Risk Scores}
\label{Appendix::RiskScoreBackground}
\input{appendix_background_material}

\clearpage
\section{Details on Computational Experiments}
\label{Appendix::SimulationDetails}
\input{appendix_computational_experiments}

\clearpage
\section{Additional Details on Algorithmic Improvements}
\label{Appendix::AlgorithmicImprovements}
\input{appendix_algorithmic_improvements}

\clearpage
\section{Additional Experimental Results}
\label{Appendix::Experiments}
\input{appendix_extra_experiments}

\clearpage
\section{Supporting Material for Seizure Prediction}
\label{Appendix::SeizurePrediction}
\input{appendix_seizure_prediction}

\end{document}

%% file: preamble.tex
\hypersetup{%
  colorlinks=true,% hyperlinks will be black
  urlcolor=blue,%
  linkcolor=blue,%
  citecolor=blue,%
  linkbordercolor=blue,% hyperlink borders will be red
  pdfborderstyle={/S/U/W 1}% border style will be underline of width 1pt
}

\newcommand{\tableformat}[0]{\centering\footnotesize\renewcommand{\arraystretch}{1.25}}

\newcolumntype{f}[1]{>{\centering\arraybackslash}m{#1}}
\newcolumntype{d}[1]{>{\raggedright\arraybackslash}m{#1}}

\newcommand{\cell}[2]{\setlength{\tabcolsep}{0pt}\begin{tabular}{#1}#2 \end{tabular}}
\newcommand{\bfcell}[2]{\setlength{\tabcolsep}{0pt}\textbf{\begin{tabular}{#1}#2\end{tabular}}}

\setitemize{leftmargin=*,topsep=0.2em,parsep=2pt,label=\raisebox{0.25ex}{\tiny$\bullet$}}

\newlist{romanlist}{enumerate}{3}
\setlist[romanlist]{label=(\roman*),itemsep=0.1em, leftmargin=2em}

\newcommand{\textds}[1]{{\footnotesize\texttt{#1}}}
\newcommand{\textfn}[1]{{\textit{#1}}}
\newcommand{\textssm}[1]{#1}

\makeatletter\xpatchcmd{\algorithmic}{\itemsep\z@}{\itemsep=0.25ex plus10pt}{}{}\makeatother
\newcommand{\algrule}[2]{\par\vskip.5\baselineskip\noindent\hfil\rule{#1}{#2}\hfil\par\vskip.5\baselineskip}
\algnewcommand\algorithmicinput{\textbf{Input}}

\algnewcommand\algorithmicinitialize{\textbf{Initialize}}
\algnewcommand\algorithmicbigstep{\textbf{Step}}
\algnewcommand\INPUT{\item[\algorithmicinput]}
\algnewcommand\INITIALIZE{\item[\algorithmicinitialize]}
\algnewcommand{\STEP}[1]{\item[\algorithmicbigstep]{\textbf{#1}}}
\algnewcommand{\InputExplanation}[2][.6\linewidth]{\leavevmode\hfill\makebox[#1][r]{~{\footnotesize{#2}}}}
\algnewcommand{\InitializationExplanation}[2][.6\linewidth]{\leavevmode\hfill\makebox[#1][r]{~{\footnotesize{#2}}}}
\algnewcommand{\alginput}[2]{\Statex{#1}\InputExplanation{#2}}
\algnewcommand{\StateComment}[2]{\State{#1}\InputExplanation{#2}}
\algnewcommand{\alginitialize}[2]{\Statex{#1}\InitializationExplanation{#2}}
\algrenewcommand\algorithmiccomment[2][]{#1\hfill\textit{\scriptsize{#2}}}

\DeclarePairedDelimiter\ceil{\lceil}{\rceil}
\DeclarePairedDelimiter\floor{\lfloor}{\rfloor}
\newcommand{\round}[1]{\left\lceil{#1}\right\rfloor}
\renewcommand{\floor}[1]{\left\lfloor{#1}\right\rfloor}

\newcommand{\for}{\textnormal{ for }}
\newcommand{\txmax}[0]{\textnormal{max}}
\newcommand{\txmin}[0]{\textnormal{min}}

\newcommand{\txforall}{\textnormal{ for all }}

\newcommand{\xb}{\bm{x}}
\newcommand{\rhob}{\bm{\rho}}

\newcommand{\n}{n}
\newcommand{\m}{m}

\newcommand{\nplus}{\n^{+}}%\textnormal{\footnotesize \textsc{+}}}}
\newcommand{\nminus}{\n^{-}}%\textnormal{\footnotesize \textsc{--}}}}
\newcommand{\lambdab}{\bm{\lambda}}
\newcommand{\minlambda}[1]{\Lambda_{#1}^{\txmin{}}}
\newcommand{\maxlambda}[1]{\Lambda_{#1}^{\txmax{}}}

\newcommand{\data}{\mathcal{D}}
\newcommand{\Nmax}{\n^\text{max}}
\newcommand{\dmax}{d^\text{max}}
\newcommand{\Noriginal}{\n^\text{original}}

\newcommand{\X}{\mathcal{X}}

\newcommand{\J}{\mathcal{J}}
\newcommand{\Sset}{\mathcal{S}}

\newcommand{\R}{\mathbb{R}}

\newcommand{\Z}{\mathbb{Z}}

\newcommand{\Lset}{\mathcal{L}}

\newcommand{\st}{\textnormal{s.t.}}
\newcommand{\mipwhat}[1]{\textit{\scriptsize #1}}
\newcommand{\miprange}[3]{{#1}={#2}\textnormal{,...,}{#3}}

\DeclareMathOperator*{\argmin}{argmin}
\newcommand{\indic}[1]{\mathbbm{1}\left[#1\right]}
\newcommand{\prob}[1]{\textnormal{Pr}\left(#1\right)}
\newcommand{\conv}[1]{\textnormal{conv}\left(#1\right)}
\newcommand{\dotprod}[2]{\langle{#1},{#2} \rangle}
\newcommand{\vnorm}[1]{\left\|#1\right\|}
\newcommand{\zeronorm}[1]{\vnorm{#1}_0}

\newcommand{\logloss}[1]{\frac{1}{\n}\sum_{i=1}^\n \log(1+\exp(-\dotprod{#1}{y_i\xb_i}))}
\newcommand{\dlogloss}[1]{\frac{1}{\n}\sum_{i=1}^\n \frac{-y_i \xb_i}{1+\exp(-\dotprod{#1}{y_i\xb_i})}}
\newcommand{\logprob}[1]{\frac{1}{1+\exp(-\dotprod{#1}{\xb_i})}}

\newcommand{\objsym}[0]{V}
\newcommand{\objval}[1]{\objsym{}\left({#1}\right)}
\newcommand{\objvalfun}[1]{\objsym{}(#1)}

\newcommand{\losssym}{l}
\newcommand{\lossfun}[1]{\losssym{}({#1})}

\newcommand{\lossgrad}[1]{\nabla \lossfun{#1}}

\newcommand{\cutloss}[1]{\hat{l}^{#1}}
\newcommand{\cutlossfun}[2]{\cutloss{#1}({#2})}

\newcommand{\lossval}{L}

\newcommand{\samplelossfun}[2]{l_{#1}({#2})}
\newcommand{\sampleobjval}[2]{V_{#1}({#2})}

\newcommand{\opt}[0]{*}
\newcommand{\objvalmin}[0]{V^\textnormal{min}}
\newcommand{\objvalmax}[0]{V^\textnormal{max}}
\newcommand{\lossmin}[0]{L^\textnormal{min}}
\newcommand{\lossmax}[0]{L^\textnormal{max}}
\newcommand{\zeronormmin}[0]{R^\textnormal{min}}
\newcommand{\zeronormmax}[0]{R^\textnormal{max}}

\newcommand{\modelfont}{\renewcommand*\familydefault{\sfdefault}\normalfont}
\newcommand{\prow}[0]{\quad\mathrel{\raisebox{-0.75ex}{\dots}}}
\newcommand{\risklabel}[0]{{\color{black}\textbf{RISK}}}
\newcommand{\scorelabel}[0]{{\color{black}\textbf{SCORE}}}

\definecolor{predcolor}{gray}{0.95}
\definecolor{scorecolor}{gray}{0.95}
\definecolor{riskcolor}{gray}{0.95}

\newcommand{\instruction}[2]{{\color{white}\phantom{\textbf{add points from rows {#1} to {#2}}}}}
\newcommand{\scoringsystem}[0]{\scriptsize\centering\renewcommand{\arraystretch}{1.35}\modelfont}
\newcommand{\risktable}[0]{\par\vspace{0.5em}\scriptsize\centering\renewcommand{\arraystretch}{1.25}\modelfont}

\newcommand{\SLIM}{{\sc{SLIM}}}
\newcommand{\RiskSLIM}{{\sc{RiskSLIM}}}

\newcommand{\CPA}{\textsf{CPA}}
\newcommand{\LCPA}{\textsf{LCPA}}
\newcommand{\DCD}{\textsf{DCD}}
\newcommand{\SR}{\textsf{SequentialRounding}}
\newcommand{\ChainedUpdates}{\textsf{ChainedUpdates}}

\newcommand{\RSMINLP}{\textsc{RiskSlimMINLP}}
\newcommand{\RSMIP}[1]{\textsc{RiskSlimMIP}(#1)}
\newcommand{\RSLP}[2]{\textsc{RiskSlimLP}(#1,#2)}
\newcommand{\rsminlp}{\textsc{RiskSlimMINLP}}
\newcommand{\rsmip}{\textsc{RiskSlimMIP}}
\newcommand{\rslp}{\textsc{RiskSlimLP}}

\newcommand{\PLR}{{\sc{PLR}}}
\newcommand{\PLRCR}{{\sc{PooledRd}}}
\newcommand{\PLRLR}{{\sc{PooledRsRd}}}
\newcommand{\PLRSR}{{\sc{PooledSeqRd}}}
\newcommand{\PLRCRDCD}{{\sc{PooledRd*}}}
\newcommand{\PLRLRDCD}{{\sc{PooledRsRd*}}}
\newcommand{\PLRSRDCD}{{\sc{PooledSeqRd*}}}

\newcommand{\mto}{{\tiny$\gg$}}
\newcommand{\mPLR}{\textsc{PLR}}

\newcommand{\mRd}{\textsc{Rd}}
\newcommand{\mRsRd}{\textsc{RsRd}}
\newcommand{\mUnit}{\textsc{Unit}}

\newcommand{\mPLRUnit}{\mPLR{}\mto{}\mUnit{}}
\newcommand{\mPLRRd}{\mPLR{}\mto{}\mRd{}}
\newcommand{\mPLRRsRd}{\mPLR{}\mto{}\mRsRd{}}
\definecolor{best}{HTML}{BAFFCD}
\definecolor{issue}{HTML}{FFC8BA}
\definecolor{bestcal}{HTML}{6EB381}
\definecolor{bestauc}{HTML}{6EB381}
\definecolor{bad}{HTML}{FFC8BA}

\graphicspath{{figure/}}

\newcommand{\stallingplot}[1]{\includegraphics[trim=0cm 0cm 0cm 0cm, clip, width=0.45\textwidth]{#1}}

\newcommand{\includescalabilitylegend}[1]{}

\newcommand{\includenewheatlegend}[1]{\includegraphics[trim=3.0cm 2.0cm 4.0cm 3.75cm, clip, width=0.08\textwidth]{#1}}
\newcommand{\includenewheatmap}[1]{\includegraphics[trim=2.2cm 0.8cm 2.0cm 0.5cm, clip, width=0.225\textwidth]{#1}}
\newcommand{\includenewheatmapNOYAXIS}[1]{\quad \includegraphics[trim=7.15cm 0.8cm 2.0cm 0.5cm, clip, width=0.175\textwidth]{#1}}

\newcommand{\includeAlgPlot}[2]{\includegraphics[width=0.4\textwidth]{figure/breastcancer_N50e3_d30_LOOKUP_TIGHT_C0_0.01_#1_#2}}

\newlength\calplotwidth
\newlength\leftmostplotwidth

\newcommand{\addcalplot}[2]{%
\cell{c}{\includegraphics[trim=50mm 15mm 0mm 10mm,clip=true, width=0.1\textwidth, keepaspectratio]{#1_#2_max_coef_5_max_L0_5_calibration_plot.pdf}}}

\newcommand{\addleftcalplot}[2]{%
\cell{c}{\includegraphics[trim=0mm 15mm 0mm 10mm,clip=true, width=0.125\textwidth, keepaspectratio]{#1_#2_max_coef_5_max_L0_5_calibration_plot.pdf}}}

\newcommand{\demoplot}[3]{\cell{c}{\includegraphics[trim=0mm 15mm 2.5mm 15mm,clip=true, width=0.345\textwidth, keepaspectratio]{seizure2_paper_sign_L0_#1_#2_#3.pdf}}}

\newcommand{\metrics}[0]{\bfcell{l}{test cal\\test auc\\loss value\\model size\\opt. gap}}
\newcolumntype{C}[1]{>{\centering\let\newline\\\arraybackslash\hspace{0pt}}m{#1}}

\newcommand{\perfheader}[1]{{\small{#1}}}

%% file: section_03_methodology.tex
In this section, we present the cutting plane algorithm that we use to solve the risk score problem. In Section \ref{Sec::CPA}, we provide a brief introduction of cutting plane algorithms to discuss their practical benefits and to explain why existing algorithms stall on non-convex problems. In Section \ref{Sec::LatticeCPA}, we present a new cutting plane algorithm that does not stall. In Section \ref{Sec::MethodComparison}, we compare the performance of cutting plane algorithms to a commercial MINLP solver on instances of the risk score problem.

\subsection{Cutting Plane Algorithms}
\label{Sec::CPA}

In Algorithm \ref{Alg::CPA}, we present a simple cutting plane algorithm to solve \RSMINLP{} that we call \CPA{}.

\begin{algorithm}[htbp]
\caption{Cutting Plane Algorithm (\CPA{})}
\label{Alg::CPA}
\begin{algorithmic}[1]
\small\vspace{0.15em}
\INPUT
\alginput{$(\xb_i, y_i)_{i=1}^\n$}{training data}
\alginput{$\Lset$}{coefficient set}
\alginput{$C_0$}{$\ell_0$ penalty parameter}
\alginput{$\varepsilon^\textrm{stop} \in [0,1]$}{maximum optimality gap of acceptable solution}
\vspace{-0.75em}\algrule{0.75\textwidth}{0.1pt}\vspace{-0.75em}
\INITIALIZE
\alginitialize{$k \gets 0$}{iteration counter}
\alginitialize{$\cutlossfun{0}{\lambdab} \gets \left\{0\right\}$}{cutting plane approximation}
\alginitialize{$(\objvalmin, \objvalmax) \gets (0, \infty)$}{bounds on the optimal value of \RSMINLP{}}
\alginitialize{$\varepsilon \gets \infty$}{optimality gap}%
\vspace{0.5em}%
\While{$\varepsilon > \varepsilon^\text{stop}$}
\State $(\lossval{}^k, \lambdab^k) \gets$ provably optimal solution to $\RSMIP{\cutloss{k}}$%
\label{AlgStep::CPASolveMIP}
\State compute cut parameters $\lossfun{\lambdab^k}$ and $\lossgrad{\lambdab^k}$%
\label{AlgStep::CPACutParameters} 
%\Comment{see \eqref{Eq::CuttingPlaneParameters}}
%
\State $\cutlossfun{k+1}{\lambdab} \leftarrow \max \{\cutlossfun{k}{\lambdab}, \lossfun{\lambdab^k} + \dotprod{\lossgrad{\lambdab^k}}{\lambdab-\lambdab^k}\}$
\Comment{update approximate loss function $\cutloss{k}$}
\label{AlgStep::CPAAddCut}%
\State $\objvalmin \leftarrow \lossval{}^k + C_0\zeronorm{\lambdab^k}$%
\label{AlgStep::CPALowerBound}%
\Comment{optimal value of \rsmip{} is lower bound}
\If{$V(\lambdab^k) < \objvalmax{}$} 
\State $\objvalmax \gets V(\lambdab^k)$%
\Comment{update upper bound}
\State $\lambdab^{\textrm{best}} \leftarrow \lambdab^k$%
\Comment{update incumbent}
\EndIf
\State $\varepsilon \gets 1 - \objvalmin{}/\objvalmax{}$
\State $k \gets k + 1$
\EndWhile
\Ensure $\lambdab^{\textrm{best}}$\hfill{$\varepsilon$-optimal solution to \RSMINLP{}}
\end{algorithmic}
\vspace{-1.0em}\algrule{0.75\textwidth}{0.1pt}\vspace{-0.5em}
$\RSMIP{\cutloss{k}}$ is a surrogate problem for \RSMINLP{} that minimizes a cutting plane approximation $\cutloss{k}$ of the loss function $\losssym$:
\begin{align}
\label{Eq::RiskScoreMIP}
\begin{split}
\min_{\lossval{}, \lambdab} & \quad  \lossval{} + C_0 \zeronorm{\lambdab}\\ 
\st &  \quad  \lossval{} \geq \cutlossfun{k}{\lambdab}\\ 
& \quad  \lambdab \in \Lset. 
\end{split}
\end{align}
We present a MIP formulation for $\RSMIP{\cutloss{k}}$ in Appendix \ref{Appendix::PerformanceComparisonSetup}.
\end{algorithm}

\begin{figure}[htbp]
\centering
\includegraphics[width=0.425\textwidth,page=1]{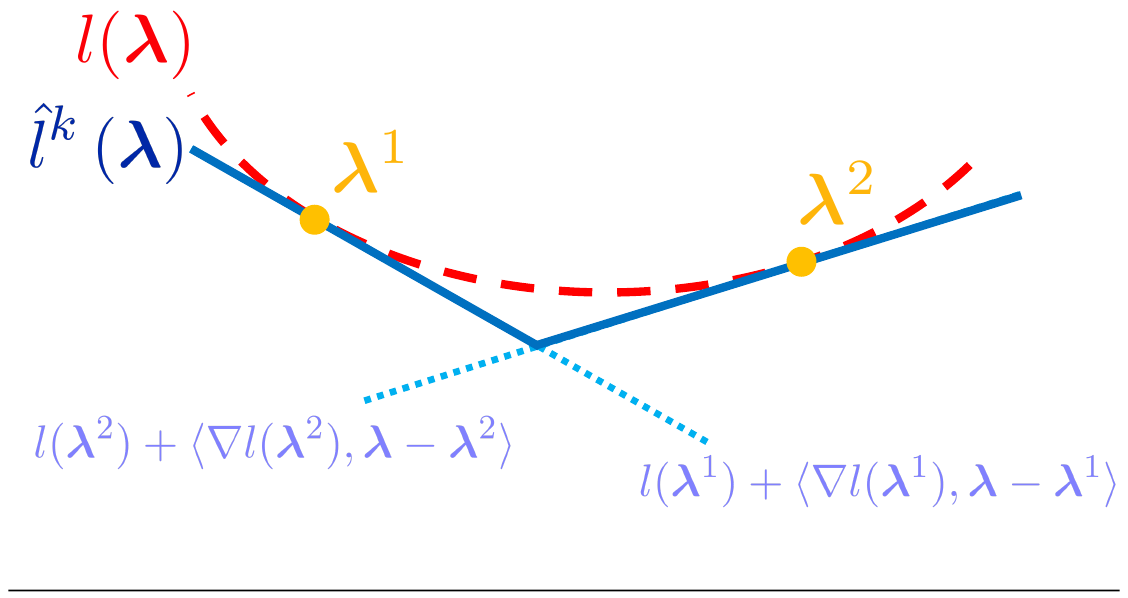} % left bottom right top
\caption{A convex loss function $\lossfun{\lambdab}$ and its cutting plane approximation $\cutlossfun{2}{\lambdab}$.} 
%= \max_{t = 1, 2} [\lossfun{\lambdab^t} + \dotprod{\lossgrad{\lambdab^t}}{\lambdab - \lambdab^t}]$.}
\label{Fig::CuttingPlaneIntuition}
\end{figure}

\CPA{} recovers the optimal solution to \RSMINLP{} by repeatedly solving a \emph{surrogate problem} that optimizes a linear approximation of the loss function $\lossfun{\lambdab}$. The approximation is built using \emph{cutting planes} or \emph{cuts}. Each cut is a supporting hyperplane to the loss function at a fixed point $\lambdab^t \in \Lset$:
$$\lossfun{\lambdab^t} + \dotprod{\lossgrad{\lambdab^t}}{\lambdab - \lambdab^t}.$$ 
Here, $\lossfun{\lambdab^t} \in \R_+$ and $\lossgrad{\lambdab^t} \in \R^{d+1}$ are \emph{cut parameters} that can be computed by evaluating the value and gradient of the loss at the point $\lambdab^t$:
\begin{align}
\label{Eq::CuttingPlaneParameters}
\lossfun{\lambdab^t} = \logloss{\lambdab^t}, \quad \lossgrad{\lambdab^t} &= \dlogloss{\lambdab^t}.
\end{align}
As shown in Figure \ref{Fig::CuttingPlaneIntuition}, we can construct a \emph{cutting plane approximation} of the loss function by taking the pointwise maximum of multiple cuts. In what follows, we denote the cutting plane approximation of the loss function built using $k$ cuts as:
\begin{align*}
\cutlossfun{k}{\lambdab} &= \max_{t=1\ldots k} \Big[ \lossfun{\lambdab^t} + \dotprod{\lossgrad{\lambdab^t}}{\lambdab - \lambdab^t} \Big].
\end{align*}

On iteration $k$, \CPA{} solves a surrogate mixed-integer program (MIP) that minimizes the cutting plane approximation $\cutloss{k}$, namely $\RSMIP{\cutloss{k}}$. \CPA{} uses the optimal solution to the surrogate MIP $(\lossval{}^k, \lambdab^k)$ in two ways: (i) it computes a new cut at $\lambdab^k$ to improve the cutting plane approximation; (ii) it computes bounds on optimal value of \RSMINLP{} to check for convergence. Here, the upper bound is set as the objective value of the best solution across all iterations:
\begin{align*}
\objvalmax &= \min_{t = 1 \ldots k} \; \Big[\lossfun{\lambdab^t} + C_0 \|\lambdab^t\|_0\Big]. \notag \\ 
\intertext{The lower bound is set as the optimal value of the surrogate problem at the current iteration:}
\objvalmin &= \cutlossfun{k}{\lambdab^k} + C_0\|\lambdab^k\|_0. %\label{Eq::CPALowerBound}
\end{align*}

\CPA{} converges to an optimal solution of \RSMINLP{} in a finite number of iterations \citep[see e.g.,][for a proof]{kelley1960cutting}. In particular, the cutting plane approximation of a convex loss function improves monotonically with each cut:
\begin{align*}
%\label{Eq::CuttingPlaneApproximationUnderestimates}
\cutlossfun{k}{\lambdab} \leq \cutlossfun{k+m}{\lambdab}\leq \lossfun{\lambdab} \text{ for all } \lambdab\in\Lset \text{ and } k, m \in \mathbb{N}.
\end{align*}
Since the cuts added at each iteration are not redundant, the lower bound improves monotonically with each iteration. Once the optimality gap $\varepsilon$ is less than a stopping threshold $\varepsilon^\text{stop}$, \CPA{} terminates and returns an $\varepsilon$-optimal solution $\lambdab^\text{best}$ to \RSMINLP{}.

\paragraph{Key Benefits of Cutting Plane Algorithms}
\label{Sec::CPAAdvantages}

\begin{figure}[htbp]
\centering
\includegraphics[width=0.45\textwidth]{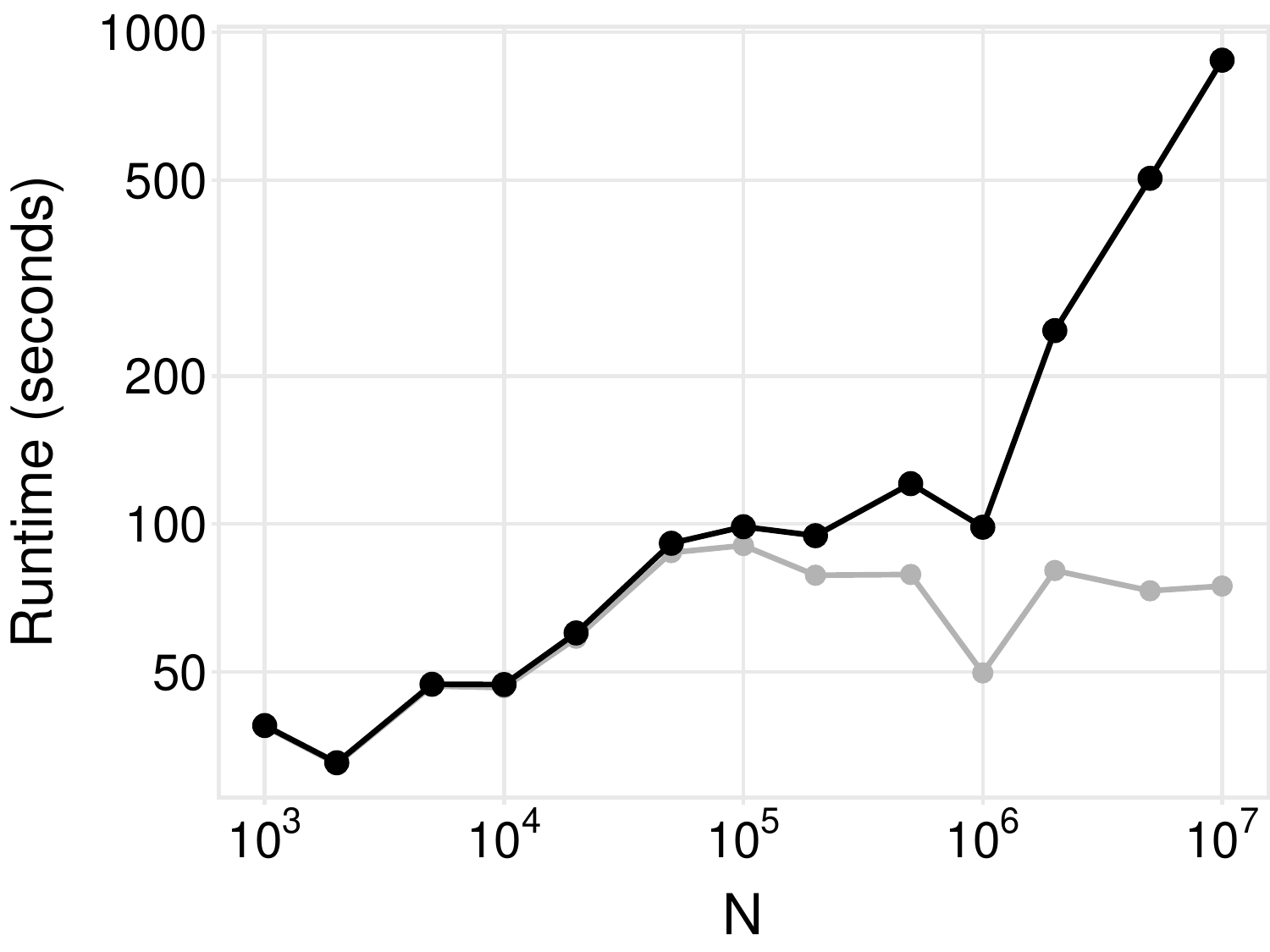}
\caption{Runtime of \CPA{} on synthetic datasets with $d = 10$ and $\n \in [10^3, 10^7]$ (see Appendix \ref{Appendix::SimulationDetails} for details). As $\n$ increases, the runtime for the solver (grey) remains roughly constant. The total runtime (black) scales at O$(\n)$, which reflects the scalability of matrix-vector operations used to compute cut parameters. }
\label{Fig::ConstantSolverTime}
\end{figure}

\CPA{} has three important properties that motivate why we want to use a cutting plane algorithm to solve the risk score problem:
\begin{enumerate}[label=(\roman*), itemsep=0.1em,leftmargin=2em]

\item \emph{Scalability in the Sample Size}: Cutting plane algorithms use the training data only when computing cut parameters, and not while solving \rsmip{}. Since the parameters in \eqref{Eq::CuttingPlaneParameters} can be computed using elementary matrix-vector operations in O($\n{}d$) time at each iteration, running time scales \emph{linearly} in $\n{}$ for fixed $d$ (see Figure \ref{Fig::ConstantSolverTime}).

\item \emph{Control over Data-related Computation}: Cutting plane algorithms compute cut parameters in a single isolated step (e.g., Step \ref{AlgStep::CPACutParameters} in Algorithm \ref{Alg::CPA}). Users can reduce data-related computation by customizing their implementation to compute these cut parameters efficiently (e.g., via distributed computing, or techniques that exploit structural properties of a specific model class as in Section \ref{Sec::DataRelatedComputation}).

\item \emph{Ability to use a MIP Solver}: Cutting plane algorithms have a special benefit in our setting since the surrogate problem can be solved with a MIP solver (rather than a MINLP solver). MIP solvers provide a fast implementation of branch-and-bound search and other features to speed up the search process (e.g., built-in heuristics, preprocessing and cut generation procedures, lazy evaluation of cut constraints, and control callbacks that let us customize the search with specialized techniques). As we show in Figure \ref{Fig::MethodComparison}, using a MIP solver can substantially improve our ability to solve \RSMINLP{}, despite the fact that one may have to solve multiple MIPs.

\end{enumerate} 

\paragraph{Stalling in Non-Convex Settings}
\label{Sec::CPAStalling}

Cutting plane algorithms for empirical risk minimization \citep[][]{joachims2006training, franc2008optimized,teo2009bundle} are similar to \CPA{} in that they solve a surrogate optimization problem at each iteration (e.g., Step \ref{AlgStep::CPASolveMIP} of Algorithm \ref{Alg::CPA}). When these algorithms are used to solve convex risk minimization problems, the surrogate is convex and therefore tractable. When the algorithms are used to solve risk minimization problems with non-convex regularizers or constraints, however, the surrogate is non-convex. In these settings, cutting plane algorithms will typically \emph{stall} as they eventually reach an iteration where the surrogate problem cannot be solved to optimality within a fixed time limit.

\begin{figure}[h]
\centering
\begin{tabular}{@{}rl@{}}
\stallingplot{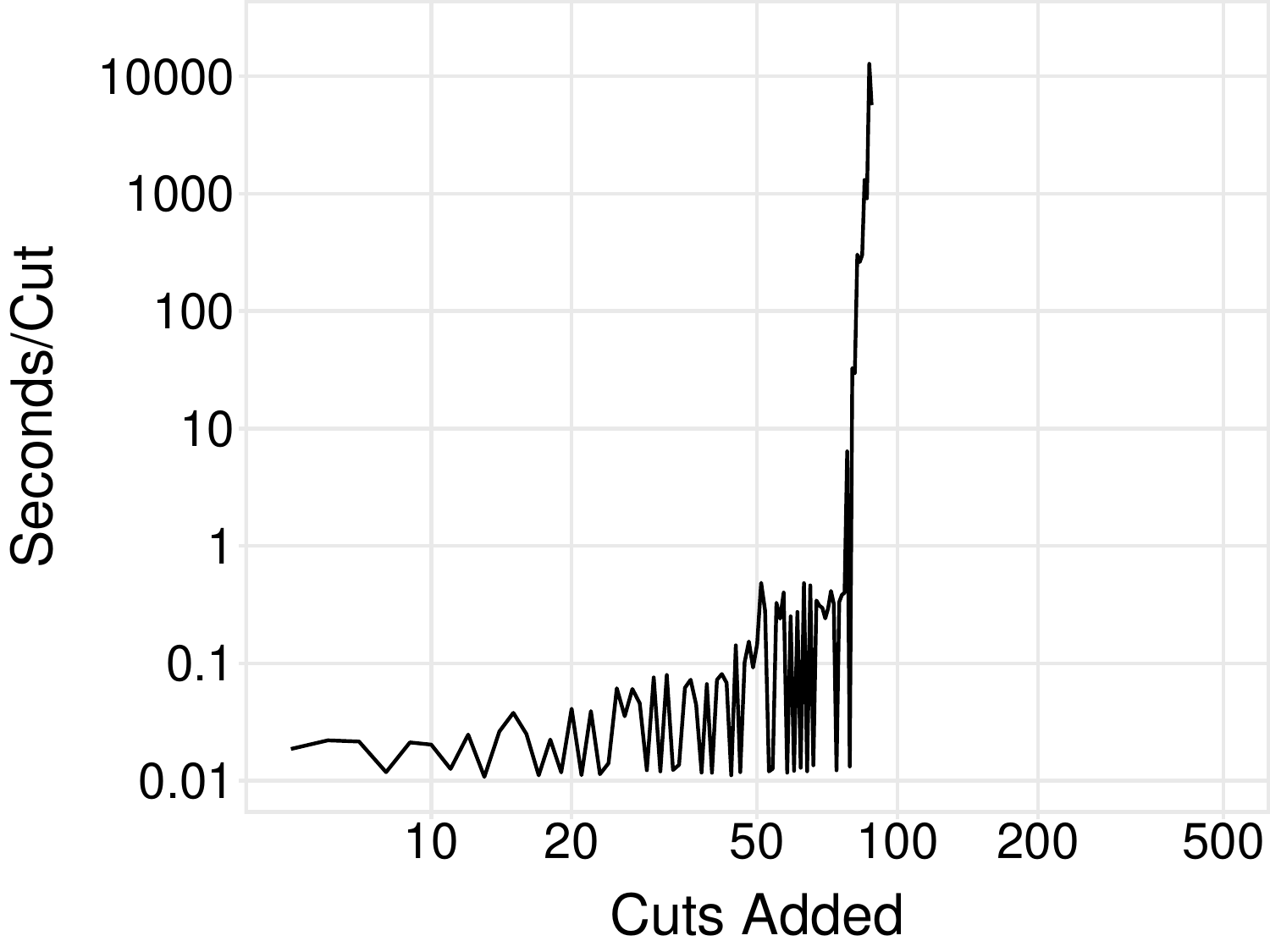} & \stallingplot{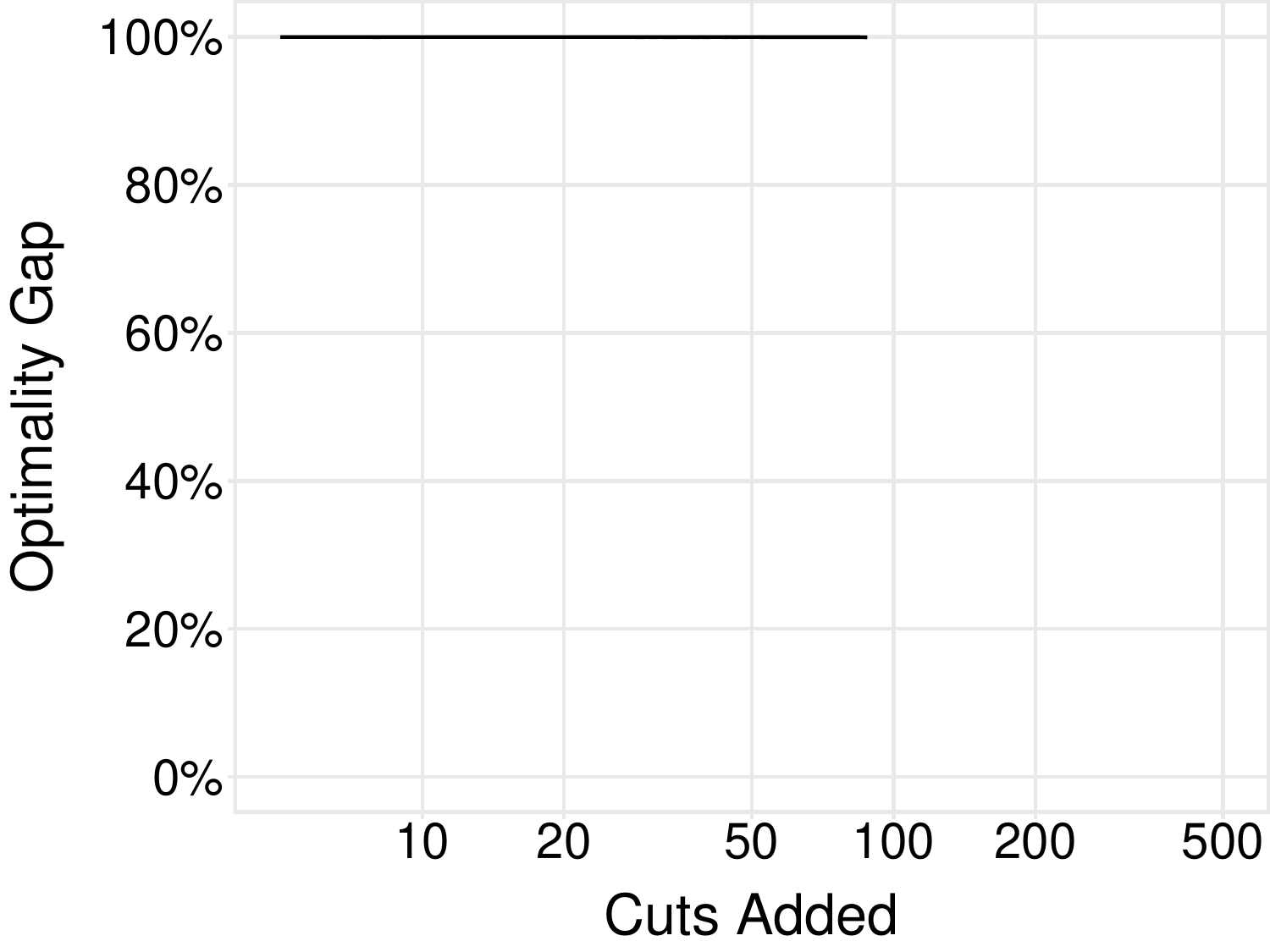} 
\end{tabular}
\caption{Performance profile of \CPA{} on \RSMINLP{} for a synthetic dataset with $\n = $ 50,000 and $d = 20$ (see Appendix \ref{Appendix::SimulationDetails} for details). We plot the time per iteration (left, in log-scale) and optimality gap (right) for each iteration over 6 hours. \CPA{} stalls on iteration 86, at which point the time to solve \rsmip{} to optimality increases exponentially. The best solution obtained after 6 hours corresponds to a risk score with poor performance.}
\label{Fig::StallingInCPA}
\end{figure}

In Figure \ref{Fig::StallingInCPA}, we illustrate the stalling behavior of \CPA{} on a difficult instance of \RSMINLP{} for a synthetic dataset where $d = 20$ (see also Figure \ref{Fig::MethodComparison}). As shown, the first iterations terminate quickly as the surrogate problem $\rsmip{}$ contains a trivial approximation of the loss. Since the surrogate becomes increasingly difficult to optimize with each iteration, however, the time to solve $\rsmip{}$ increases exponentially, leading \CPA{} to stall at iteration $k = 86$. In this case, the solution returned by \CPA{} after 6 hours has a large optimality gap and a highly suboptimal loss. This is unsurprising, as the solution was obtained by optimizing a low-fidelity approximation of the loss (i.e., an 85-cut approximation of a 20-dimensional function). Since the value of the loss is tied to the performance of the model (see Section \ref{Sec::Experiments}), the solution corresponds to a risk score with poor performance.

There is no simple fix to prevent standard cutting plane algorithms such as \CPA{} from stalling on non-convex problems. This is because they need a globally optimal solution to a surrogate optimization problem at each iteration to compute a valid lower bound. In non-convex risk minimization problems, this requires finding the optimal solution of a non-convex surrogate problem, and \emph{certifying} that there does not exist a better solution to the surrogate problem. If, for example, \CPA{} only solved the surrogate until it found a feasible solution with a non-zero optimality gap, then it could produce a cutting plane that discards the true optimal solution. In this case, the lower bound computed in Step \ref{AlgStep::CPALowerBound} would exceed the true optimal value, leading the algorithm to terminate prematurely and return a suboptimal solution with invalid bounds. 

%Seeing how the stalling behavior of \CPA{} is related to the mechanism used to check convergence, a tempting (but flawed) solution is to use an algorithm that constructs a cutting plane approximation by computing cuts at central points of $\rsmip{}$, such as the center of gravity algorithm of \citet{levin1965algorithm}, or the analytic center algorithm of \citet{atkinson1995cutting}. Such algorithms are guaranteed to converge in a fixed number of iterations and do not require computing a lower bound. In this case, however, they would still stall on high-dimensional problems as they would need to compute central points by solving a non-convex optimization problem to optimality at each iteration.

\subsection{The Lattice Cutting Plane Algorithm}
\label{Sec::LatticeCPA}

\begin{figure}[htbp]
\centering
\begin{tabular}{@{}rl@{}}
\stallingplot{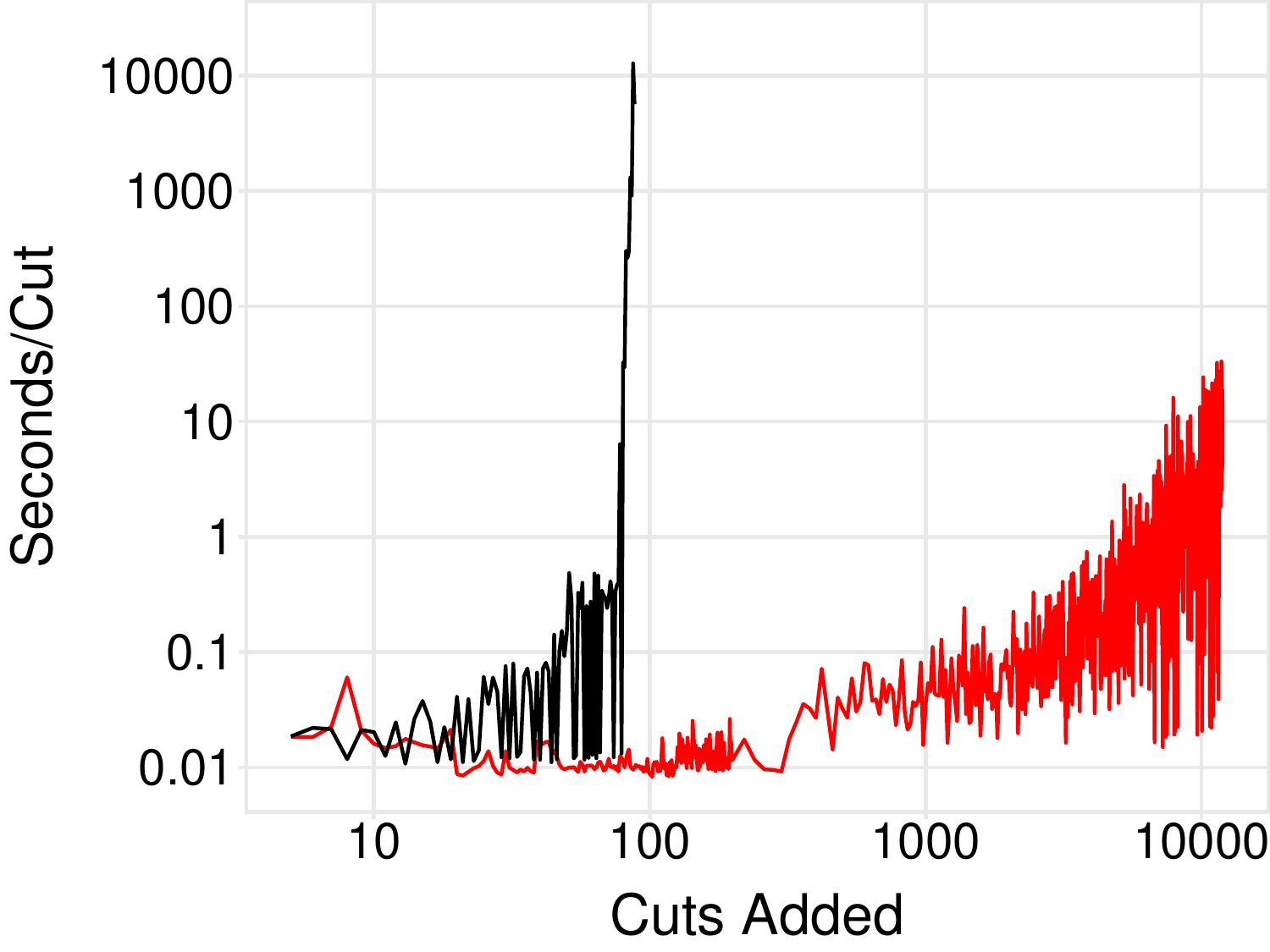} & \stallingplot{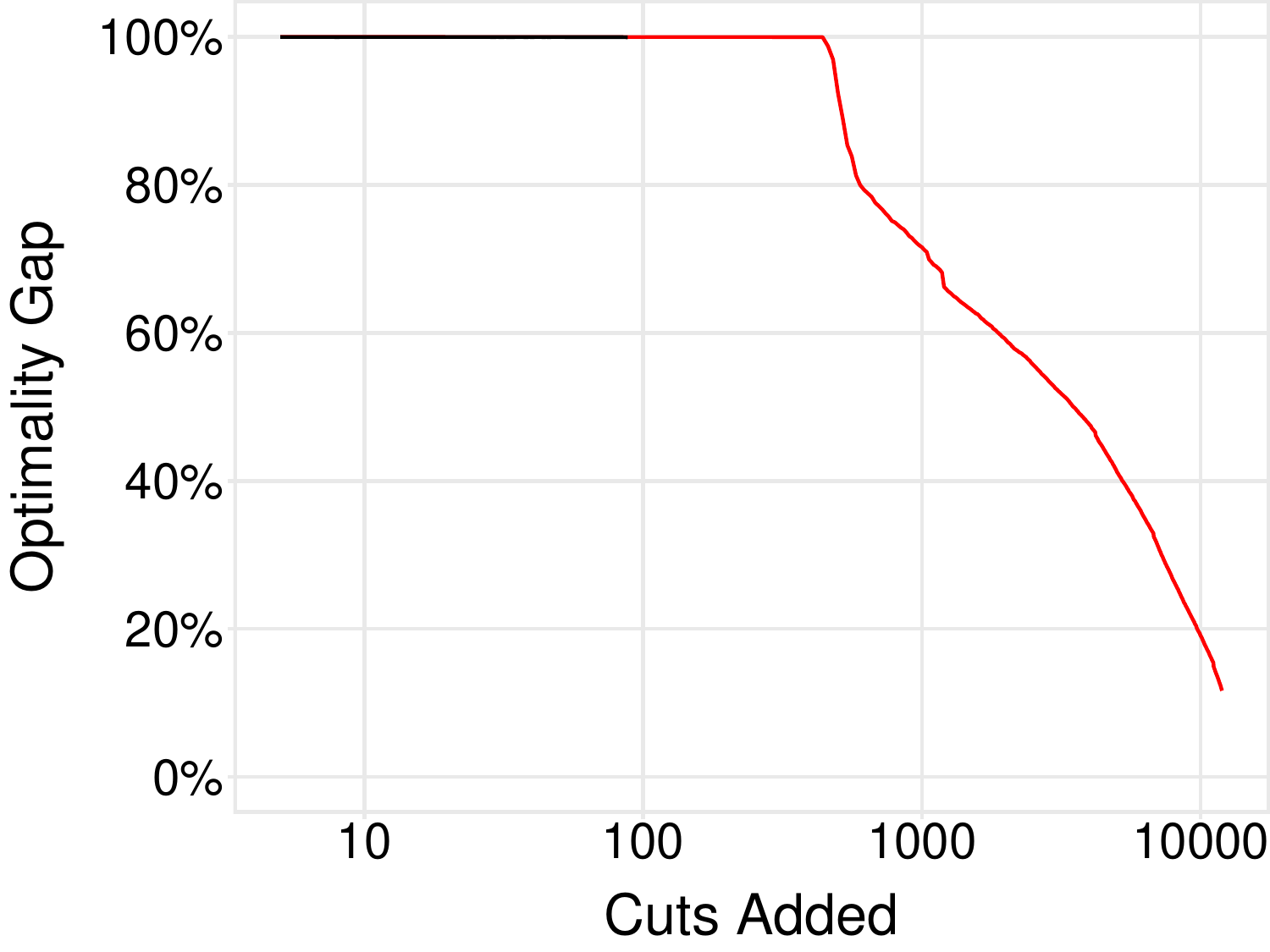} 
\end{tabular}
\caption{Performance profile of \LCPA{} (red) and \CPA{} (black) on the \RSMINLP{} instance in Figure \ref{Fig::StallingInCPA}. Unlike \CPA{}, \LCPA{} does not stall. \LCPA{} recovers a high-quality risk score (i.e., whose objective value is $\leq10\%$ of the optimal value) in 9 minutes after adding 4,655 cuts. The remaining time is used to reduce the optimality gap.}
\label{Fig::NoStallingInLCPA}
\end{figure}

To avoid stalling in non-convex settings, we solve the risk score problem using the \emph{lattice cutting plane algorithm} (\LCPA{}) shown in Algorithm \ref{Alg::LCPA}. \LCPA{} has the same benefits as other cutting plane algorithms for the risk score problem, such as scalability in the sample size, control over data-related computation, and the ability to use a MIP solver. As shown in Figure \ref{Fig::NoStallingInLCPA}, however, \LCPA{} does not stall. This is because it can add cuts and compute a lower bound without having to optimize a non-convex surrogate.

\LCPA{} recovers the optimal solution to \RSMINLP{} via \emph{branch-and-bound} (B\&B) search. The search process recursively splits the feasible region of \RSMINLP{}, discarding parts that are infeasible or provably suboptimal. \LCPA{} solves a \emph{surrogate linear program} (LP) over each region. It updates the cutting plane approximation when the surrogate LP yields an integer feasible solution. At that point, it sets the lower bound for the risk score problem as the smallest lower bound of the surrogate LP over unexplored regions. 

\newcommand{\bbsplit}{\textsf{SplitRegion}}
\newcommand{\bbselection}{\textsf{RemoveNode}}
\newcommand{\bbpool}{\mathcal{N}}
\newcommand{\bbval}[1]{v^{#1}}
\newcommand{\bbpart}[1]{\mathcal{R}^{#1}}
\newcommand{\LP}[0]{t}%{\tiny\text{t}}}
\begin{algorithm}[htbp]
\algrenewcommand{\InputExplanation}[2][.6\linewidth]{\leavevmode\hfill\makebox[#1][r]{~{\scriptsize{#2}}}}
\algrenewcommand{\InitializationExplanation}[2][.6\linewidth]{\leavevmode\hfill\makebox[#1][r]{~{\scriptsize{#2}}}}

\caption{Lattice Cutting Plane Algorithm (\LCPA{})}
\label{Alg::LCPA}
\footnotesize
\begin{algorithmic}[1]
\vspace{0.15em}

\INPUT
\alginput{$(\xb_i, y_i)_{i=1}^\n$}{training data}
\alginput{$\Lset$}{coefficient set}
\alginput{$C_0$}{$\ell_0$ penalty parameter}
\alginput{$\varepsilon^\textrm{stop} \in [0,1]$}{optimality gap of acceptable solution}
\alginput{\bbselection{}}{procedure to remove node from a node set (provided by MIP solver)}
\alginput{\bbsplit{}}{procedure to split region into disjoint subsets (provided by MIP solver)}%
\alginput{$\RSLP{\cutloss{}}{\bbpart{}}$}{LP relaxation of $\RSMIP{\cutloss{}}$ over the region $\bbpart{} \subseteq \conv{\Lset}$ (see Definition \ref{Def::RSLP})}
\vspace{-0.5em}\algrule{0.5\textwidth}{0.1pt}\vspace{-0.5em}
\INITIALIZE
\alginitialize{$k \gets 0$}{number of cuts}
\alginitialize{$\cutlossfun{k}{\lambdab} \gets \left\{0\right\}$}{cutting plane approximation}
\alginitialize{$(\objvalmin, \objvalmax) \gets (0, \infty)$}{bounds on the optimal value of \RSMINLP{}}
\alginitialize{$\bbpart{0} \gets \conv{\Lset}$}{initial region is convex hull of coefficient set}
\alginitialize{$\bbval{0} \gets 0$}{lower bound of the objective value of the surrogate LP at $\bbpart{0}$}
\alginitialize{$\bbpool \gets \{(\bbpart{0}, \bbval{0})\}$}{node set}
\alginitialize{$\varepsilon \gets \infty$}{optimality gap}\vspace{0.75em}

\While{$\varepsilon > \varepsilon^\textrm{stop}$}

\State $(\bbpart{t}, \bbval{t}) \gets \bbselection{}\left(\bbpool{}\right)$ \Comment{$t$ is index of removed node} \label{AlgStep::LCPAChooseNode}
\State solve $\RSLP{\cutloss{k}}{\bbpart{t}}$ \label{AlgStep::LCPASolveLP} 
\State $\lambdab^\LP{} \gets$ coefficients from optimal solution to  $\RSLP{\cutloss{k}}{\bbpart{t}}$
\State $V^\LP{}\gets$ optimal value of $\RSLP{\cutloss{k}}{\bbpart{t}}$
\If{optimal solution is integer feasible} 
\label{AlgStep::LCPAIntegerFeasible}

\State compute cut parameters $\lossfun{\lambdab^\LP{}}$ and $\lossgrad{\lambdab^\LP{}}$ \label{AlgStep::LCPAComputeCut}

\State $\cutlossfun{k+1}{\lambdab} \leftarrow \max \{\cutlossfun{k}{\lambdab}, \lossfun{\lambdab^\LP{}} + \dotprod{\lossgrad{\lambdab^k}}{\lambdab-\lambdab^\LP{}}\}$ \label{AlgStep::LCPAAddCut} 
\Comment{update approximate loss function $\cutloss{k}$}

\If{$V^\LP{} < \objvalmax{}$} \label{AlgStep::LCPACheckUB} 
\State $\objvalmax \gets V^\LP{}$ 
\Comment{update lower bound}
\label{AlgStep::LCPAUpdateUB}

\State $\lambdab^{\textrm{best}} \leftarrow \lambdab^\LP{}$  
\Comment{update best solution}
\label{AlgStep::LCPAUpdateIncumbent}

\State $\bbpool{} \leftarrow \bbpool{} \setminus \{ (\bbpart{s}, \bbval{s}) ~ | ~ \bbval{s} \geq \objvalmax\}$ 
\Comment{prune suboptimal nodes}
\label{AlgStep::LCPAPruneNodes}

\EndIf 
\label{AlgStep::LCPAIntegerFeasibleEnd}

\State $k \gets k + 1$

\ElsIf{optimal solution is not integer feasible} 
\label{AlgStep::LCPANonIntegerSolution} 
		\State $(\bbpart{}$$'$, $\bbpart{}$$'') \gets \bbsplit(\bbpart{t}, \lambdab^\LP{})$ \Comment{$\bbpart{}$$'$, $\bbpart{}$$''$ are disjoint subsets of $\bbpart{t}$} \label{AlgStep::LCPABranch}
%		\State $\bbval{} \gets $ \Comment{$V^\LP{}$ is lower bound of \rslp{} over $\bbpart{}$$', \bbpart{}$$''$}
		\State $\bbpool{} \leftarrow \bbpool{} \cup \{ (\bbpart{}$$'$, $V^\LP{}), (\bbpart{}$$''$$, V^\LP{})\}$ \label{AlgStep::LCPAAddChildNodes} \Comment{$V^\LP{}$ is lower bound of \rslp{} for child regions $\bbpart{}$$', \bbpart{}$$''$}
%\If{$\bbval{t} > \objvalmin$} 
%\State optional: improve bounds via chained updates (Algorithm \ref{Alg::ChainedUpdates}) 
%\EndIf
\EndIf
\State $\objvalmin \gets \min_{s = 1\ldots|\bbpool{}|} \bbval{s}$ \Comment{$\objvalmin$ is smallest lower bound among nodes in $\bbpool{}$} \label{AlgStep::LCPAUpdateLB}
\State $\varepsilon \gets 1 - \objvalmin/\objvalmax$ \Comment{update optimality gap}
\EndWhile
\Ensure \vspace{0.25em}$\lambdab^\text{best}$ \hfill {$~\varepsilon$-optimal solution to \RSMINLP{}}
%
%\vspace{-0.5em}\algrule{0.5\textwidth}{0.1pt}\vspace{-0.5em}
\end{algorithmic}
%$\RSLP{\cutloss{k}}{\bbpart{}}$ is an LP relaxation of $\RSMIP{\cutloss{k}}$ over the convex region $\bbpart{} \subseteq \conv{\Lset}$.
%\vspace{-0.5em}
%%
%\begin{align}
%\label{Eq::RiskScoreLP}
%\begin{split}
%\min_{\lossval{},\lambdab,\bm{\alpha}}  &\quad \lossval{} + C_0 \sum_{j=1}^d \alpha_j \\
%\st & \quad \lambdab  \in  \bbpart{} \\ 
%       & \quad \lossval{}   \geq \cutlossfun{}{\lambdab}\\
%        & \quad \alpha_j  = \max(\lambda_j, 0)/\maxlambda{j} + \min(\lambda_j, 0)/\minlambda{j} \qquad\text{for}~ j = 1,\ldots,d.
%\end{split}
%\end{align}
%We present an LP formulation for $\RSLP{\cutloss{k}}{\bbpart{}}$ in Definition \ref{Def::RSLP}.
\end{algorithm}

\FloatBarrier
\begin{definition}[\rslp{}]
\label{Def::RSLP}
Given a bounded convex region $\bbpart{} \subseteq \conv{\Lset}$, trade-off parameter $C_0 > 0$, cutting plane approximation $\cutloss{k}: \R^{d+1} \to \R_+$ with cut parameters $\{\lossfun{\lambdab^t},\lossgrad{\lambdab^t}\}_{t=1}^k$, %
and bounds $\objvalmin$, $\objvalmax{}$, $\lossmin{}$, $\lossmax{}$, $\zeronormmin$, $\zeronormmax$,
the surrogate optimization problem $\RSLP{\cutloss{k}}{\bbpart{}}$ can be formulated as the linear program:
{%
\small%\setstretch{1.25}
\begin{subequations}
\label{Formulation::RSLP}
\begin{equationarray}{@{}crcl>{\qquad}l>{\qquad}r@{}}
\min_{\lossval{},\lambdab,\bf{\alpha}} & V  \notag \\ %& =  & \lossval{} + C_0 R    & & \\ 
\st 
& V & = &  \lossval{} + C_0 R    & & \mipwhat{objective value} \label{Con::RSLPObjVar} \notag\\
& R & = & \sum_{j=1}^d \alpha_j & & \mipwhat{relaxed $\ell_0$-norm} \label{Con::RSLPL0Var} \notag\\
& \lossval{} & \geq & \lossfun{\lambdab^t} + \dotprod{\lossgrad{\lambdab^t}}{\lambdab{} - \lambdab^t} & \miprange{t}{1}{k} & \mipwhat{cut constraints} \label{Con::RSLPCuts} \notag\\
&  \lambda_j  & \leq &  \maxlambda{j} \alpha_j  & \miprange{j}{1}{d} & \mipwhat{$\ell_0$-indicator constraints} \label{Con::RSLPL0UB}  \notag\\
&  \lambda_j  & \geq & - \minlambda{j} \alpha_j  & \miprange{j}{1}{d} & \mipwhat{$\ell_0$-indicator constraints} \label{Con::RSLPL0LB}  \notag\\[1.5em]
& \lambdab & \in & \bbpart{} & & \mipwhat{feasible region} \label{Con::RSLPRegion}  \notag\\
& V  & \in & [\objvalmin{}, \objvalmax{}] & & \mipwhat{objective bounds} \label{Con::RSLPObjBds} \notag\\
& \lossval{} & \in & [\lossmin{}, \lossmax{}] &  & \mipwhat{loss bounds} \label{Con::RSLPLossBds} \notag\\
& R & \in & [\zeronormmin, \zeronormmax] & & \mipwhat{relaxed $\ell_0$-bounds} \label{Con::RSLPL0Bds}  \notag\\
%
%&  \lambda_j  & \in & [\minlambda{j}, \maxlambda{j}] & \miprange{j}{1}{d} & \mipwhat{relaxed coefficient bounds} \label{Con::RSLPCoefBds} \notag\\
%
& \alpha_j  & \in & [0,1]  & \miprange{j}{1}{d} & \mipwhat{relaxed $\ell_0$-indicators} \label{Con::RSLPL0VarBds} \notag
\end{equationarray}
\end{subequations}
}
\end{definition}

\paragraph{Branch-and-Bound Search} 

In Algorithm \ref{Alg::LCPA}, we represent the state of the B\&B search process using a B\&B tree. We refer to each leaf of the tree as a \emph{node}, and denote the set of all nodes as $\bbpool{}$. Each \emph{node} $(\bbpart{t}, \bbval{t}) \in \bbpool$ consists of: a \emph{region} of the convex hull of the coefficient set $\bbpart{t} \subseteq \conv{\Lset}$; and a lower bound on the objective value of the surrogate LP over this region $\bbval{t}$.

Each iteration of \LCPA{} removes a node from the node set $(\bbpart{t}, \bbval{t}) \in \bbpool{}$, then solves the surrogate LP for the corresponding region, that is, $\RSLP{\cutloss{k}}{\bbpart{t}}$. Subsequent steps of the algorithm are determined by the solution status of the surrogate LP:
\begin{itemize}

\item If $\RSLP{\cutloss{k}}{\bbpart{t}}$ has an integer solution, \LCPA{} updates the cutting plane approximation $\cutloss{k}$ with a new cut at $\lambdab^\LP{}$ in Step \ref{AlgStep::LCPAAddCut}.

\item If $\RSLP{\cutloss{k}}{\bbpart{t}}$ has a real-valued solution, \LCPA{} adds two child nodes $(\bbpart{}$$', v^\LP{})$ and $(\bbpart{}$$'',v^\LP{})$ to the node set $\bbpool{}$ in Step \ref{AlgStep::LCPAAddChildNodes}. The child nodes are produced by applying a splitting rule, which splits $\bbpart{t}$ into disjoint regions $\bbpart{}$$'$ and $\bbpart{}$$''$. The lower bound for each child node is set as the optimal value of the surrogate LP $v^\LP{}$.

\item If $\RSLP{\cutloss{k}}{\bbpart{t}}$ is infeasible, then \LCPA{} discards the node from the node set. 
\end{itemize}
%
%The number of nodes will increase whenever the surrogate LP has a real-valued solution, and decrease whenever the surrogate LP is infeasible or has an integer feasible solution.
%
The B\&B search is governed by two procedures that are implemented in a MIP solver: 
\begin{itemize}
\item $\bbselection{}$, which removes a node ($\bbpart{t}, \bbval{t})$ from the node set $\bbpool{}$ (e.g., the node with the smallest lower bound $\bbval{t}$).

\item $\bbsplit{}$, which splits $\bbpart{t}$ into disjoint subsets of $\bbpart{t}$ (e.g., split on a fractional component of $\lambdab^\LP{}$, which returns $\bbpart{}$$' = \{ \lambdab \in \bbpart{\LP{}} \,|\, \lambda_j^\LP{} \geq \lceil\lambda_j^\LP{}\rceil\}$ and $\bbpart{}$$'' = \{\lambdab \in \bbpart{\LP{}} \,|\, \lambda_j^\LP{} \leq \lfloor \lambda_j^\LP{} \rfloor \}$). The output conditions for $\bbsplit{}$ must ensure that the regions at each node remain disjoint, the total number of nodes remains finite, and the total search region shrinks even when the surrogate LP has a real-valued solution.

\end{itemize}

%\paragraph{Convergence} 

\LCPA{} evaluates the optimality of solutions to the risk score problem by using bounds on the objective value of \RSMINLP{}. The upper bound $\objvalmax{}$ is set as the objective value of the best integer feasible solution in Step \ref{AlgStep::LCPAUpdateUB}. The lower bound $\objvalmin{}$ is set as the smallest objective value among all nodes in Step \ref{AlgStep::LCPAUpdateLB}. The value of $\objvalmin$ can be viewed as a lower bound on the objective value of the surrogate LP over the \emph{remaining search region} $\bigcup_{t} \bbpart{t}$ (i.e., $\objvalmin$ is a lower bound on the objective value of $\RSLP{\cutloss{k}}{\bigcup_{t} \bbpart{t}}$). Thus, $\objvalmin{}$ will increase when we reduce the remaining search region or add cuts.%
%\footnote{The \LCPA{} lower bound is typically weaker than the \CPA{} lower bound. Specifically, the \CPA{} lower bound corresponds to an integer feasible solution while the \LCPA{} lower bound corresponds to a continuous solution. As such, on instances where it does not stall, \CPA{} may converge faster than the \LCPA{} since it produces a stronger lower bound at each iteration.}.

Each iteration of \LCPA{} reduces the remaining search region by either finding an integer feasible solution, identifying an infeasible region, or splitting a region into disjoint subsets. Thus, $\objvalmin{}$ increases monotonically as the search region becomes smaller, and cuts are added at integer feasible solutions. Likewise, $\objvalmax{}$ decreases monotonically as it is set as the objective value of the best integer feasible solution. Since there are a finite number of nodes, even in the worst-case, \LCPA{} terminates after a finite number of iterations, returning an optimal solution to the risk score problem.

\begin{remark}[Worst-Case Data-Related Computation for \LCPA{}]
\label{Rem::LCPAMaxCuts}
Given any training dataset $(\xb_i, y_i)_{i=1}^\n$, any trade-off parameter $C_0 > 0$, and any finite coefficient set $\Lset \subset \Z^{d+1}$, {\textnormal\LCPA{}} returns an optimal solution to the risk score problem after computing at most $|\Lset|$ cutting planes, and processing at most $2^{|\Lset|} - 1$ nodes.
\end{remark}

\paragraph{Implementation with a MIP Solver with Lazy Cut Evaluation} 

\LCPA{} can easily be implemented using a MIP solver (e.g., CPLEX, Gurobi, GLPK) with \emph{control callbacks}. In this approach, the solver handles the B\&B related steps of Algorithm \ref{Alg::LCPA}, and one needs only to write a few lines of code to update the cutting plane approximation when the algorithm finds an integer feasible solution. In a basic implementation, the solver would call the control callback when it finds an integer feasible solution (i.e., Step \ref{AlgStep::LCPAIntegerFeasible}). The code would retrieve the integer feasible solution, compute the cut parameters, and add a cut to the surrogate LP, handing control back to the solver at Step \ref{AlgStep::LCPACheckUB}.

A key benefit of using a MIP solver is the ability to add cuts as \emph{lazy constraints}. In practice, if we were to add cuts as generic constraints to the surrogate LP, the time to solve the surrogate LP would increase with each cut, which would progressively slow down \LCPA{}. When we add cuts as lazy constraints, the solver branches using a surrogate LP that contains a subset of relevant cuts, and only evaluates the complete set of cuts when \LCPA{} finds an integer feasible solution. In this case, \LCPA{} still returns the optimal solution. However, computation is significantly reduced as the surrogate LP is much faster to solve for the vast majority of cases where it is infeasible or yields a real-valued solution. From a design perspective, lazy cut evaluation reduces the marginal computational cost of adding cuts, which allows us to add cuts liberally (i.e., without having to worry about slowing down \LCPA{} by adding too many cuts).

%The downside of this approach, however, is that by omitting the cuts from the surrogate LP, we are also removing information related to the loss function from the surrogate LP, and thereby making the search process oblivious to the value of the loss function. We are able to partially mitigate this issue by explicitly include an auxiliary variable for the value of the loss in the surrogate LP, and updating bounds on this variable over the course of \LCPA{} (see Section \ref{Sec::ChainedUpdates}).
 
\subsection{Performance Comparison with MINLP Algorithms}
\label{Sec::MethodComparison}

In what follows, we benchmark \CPA{} and \LCPA{} against three MINLP algorithms as implemented in a commercial MINLP solver \citep[Artelsys Knitro 9.0, which is an updated version of the solver in][]{byrd2006knitro}.%
% \footnote{Knitro is one of several MINLP solvers that can solve the risk score problem directly \citep[see][for others]{bussieck2010minlp}. We chose Knitro because it let us control factors that would otherwise lead an MINLP solver to perform poorly in benchmarks, namely: (i) Knitro provides the ability to solve LP subproblems with a third-party LP solver, which let us ensure that all algorithms used the same LP solver \citep[i.e., CPLEX 12.6.3,][]{cplex}; (ii) Knitro provides the ability to compute the objective function of the MINLP with a user-defined function handle, which let us monitor and control data-related computation by using the same functions to evaluate the objective, its gradient and Hessian.}

\begin{figure}[h]
\centering
\resizebox{\textwidth}{!}{
\setlength{\tabcolsep}{0pt}
\renewcommand{\arraystretch}{0.0}
\begin{tabular}{l>{\quad}c>{\quad}c@{}c@{}c@{}}
\toprule
& & 
\hspace{2.5em}\cell{c}{\footnotesize\textsf{LCPA}}  & 
\hspace{0.75em}\cell{c}{\footnotesize\textsf{CPA}}  & 
\cell{c}{\footnotesize\textsf{ActiveSetMINLP}} \\
\midrule
{\renewcommand{\arraystretch}{1.05}%
\itshape\scriptsize
\begin{tabularx}{0.5\textwidth}{X}
{\bf{Time to Train a Good Risk Score}}\\[0.25em]
i.e., the time for an algorithm to find a solution whose loss is $\leq 10\%$ of the optimal loss. This reflects the time to obtain a risk score with good calibration without a proof of optimality.
\end{tabularx}} & 
\cell{c}{\includenewheatlegend{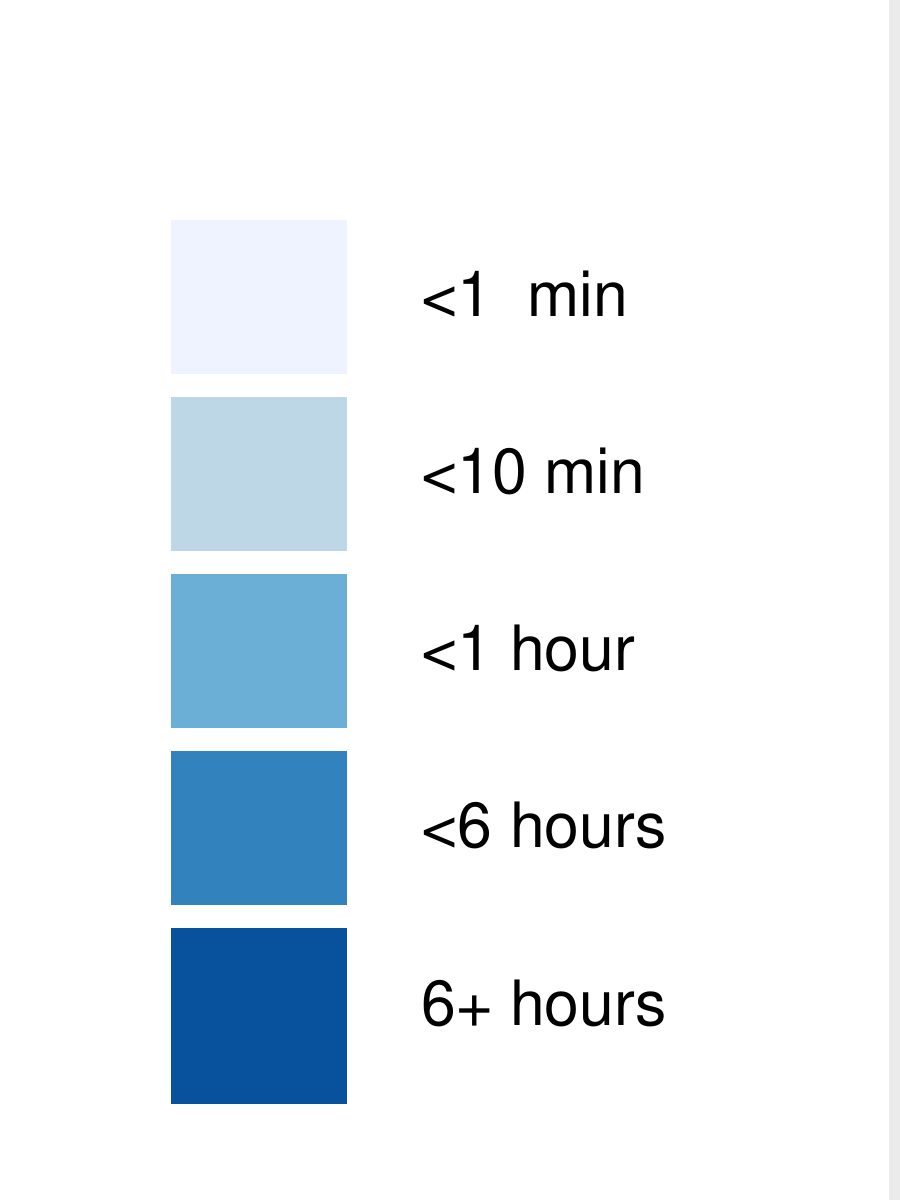}} &
\cell{c}{\includenewheatmap{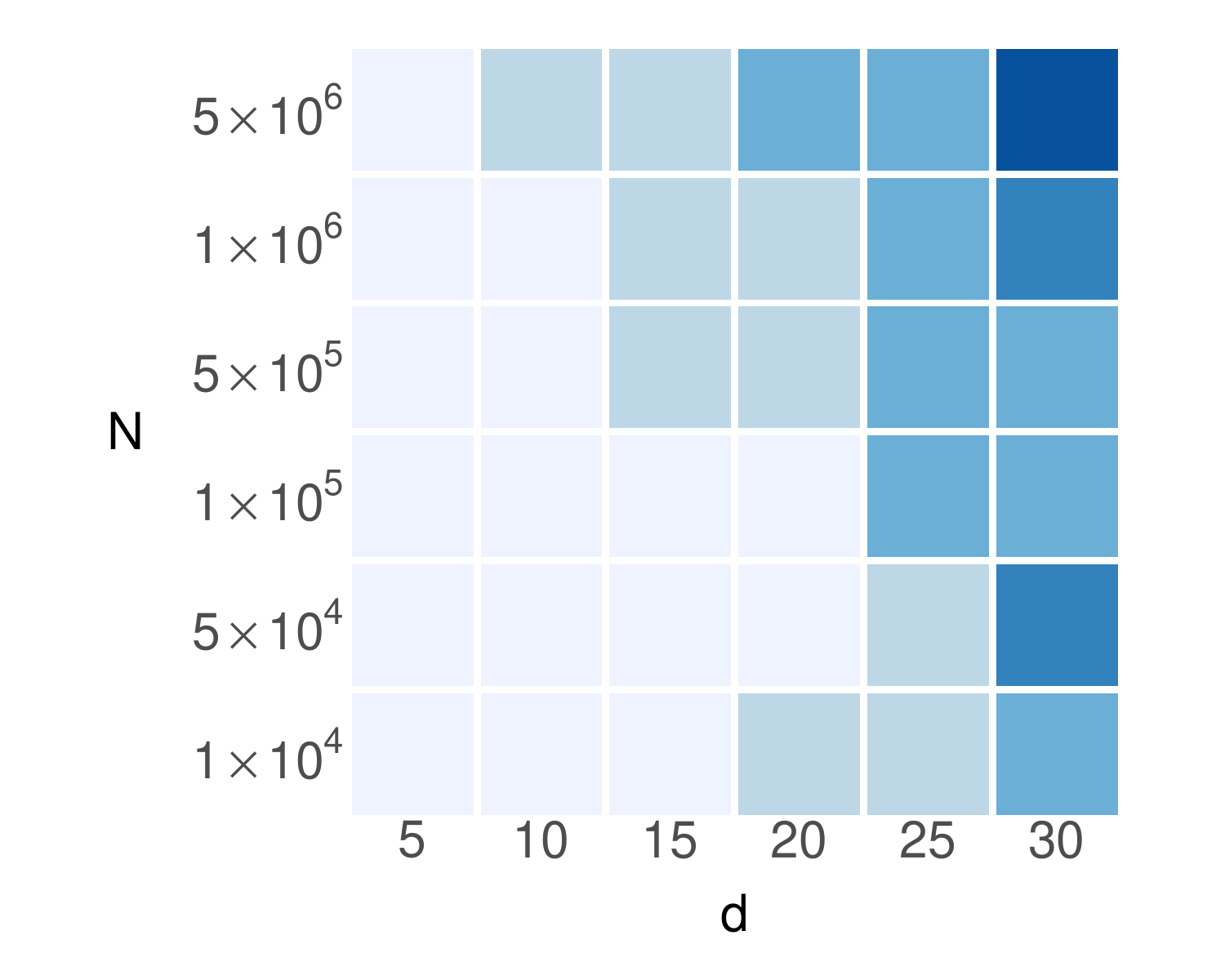}} & 
\cell{c}{\includenewheatmapNOYAXIS{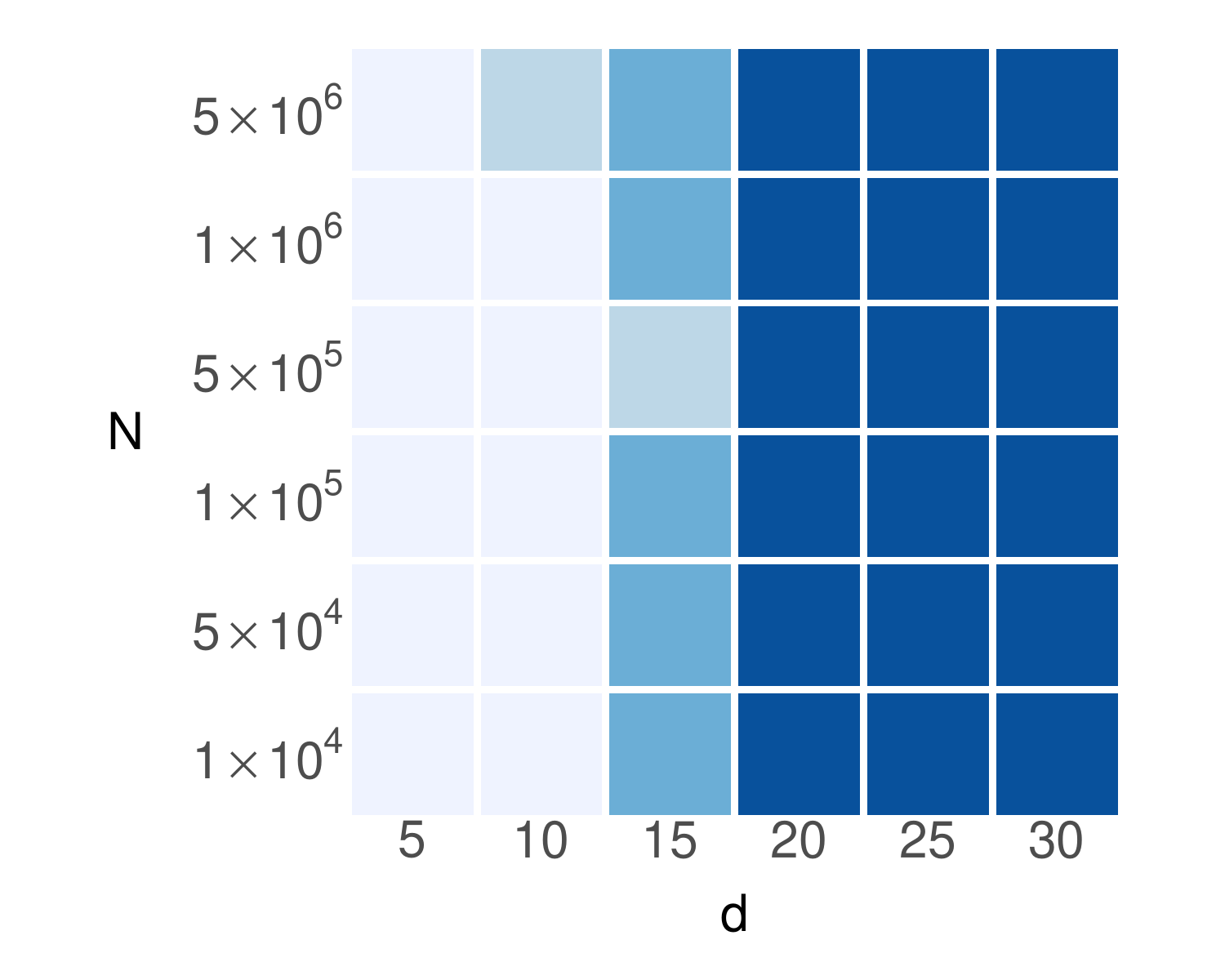}} & 
\cell{c}{\includenewheatmapNOYAXIS{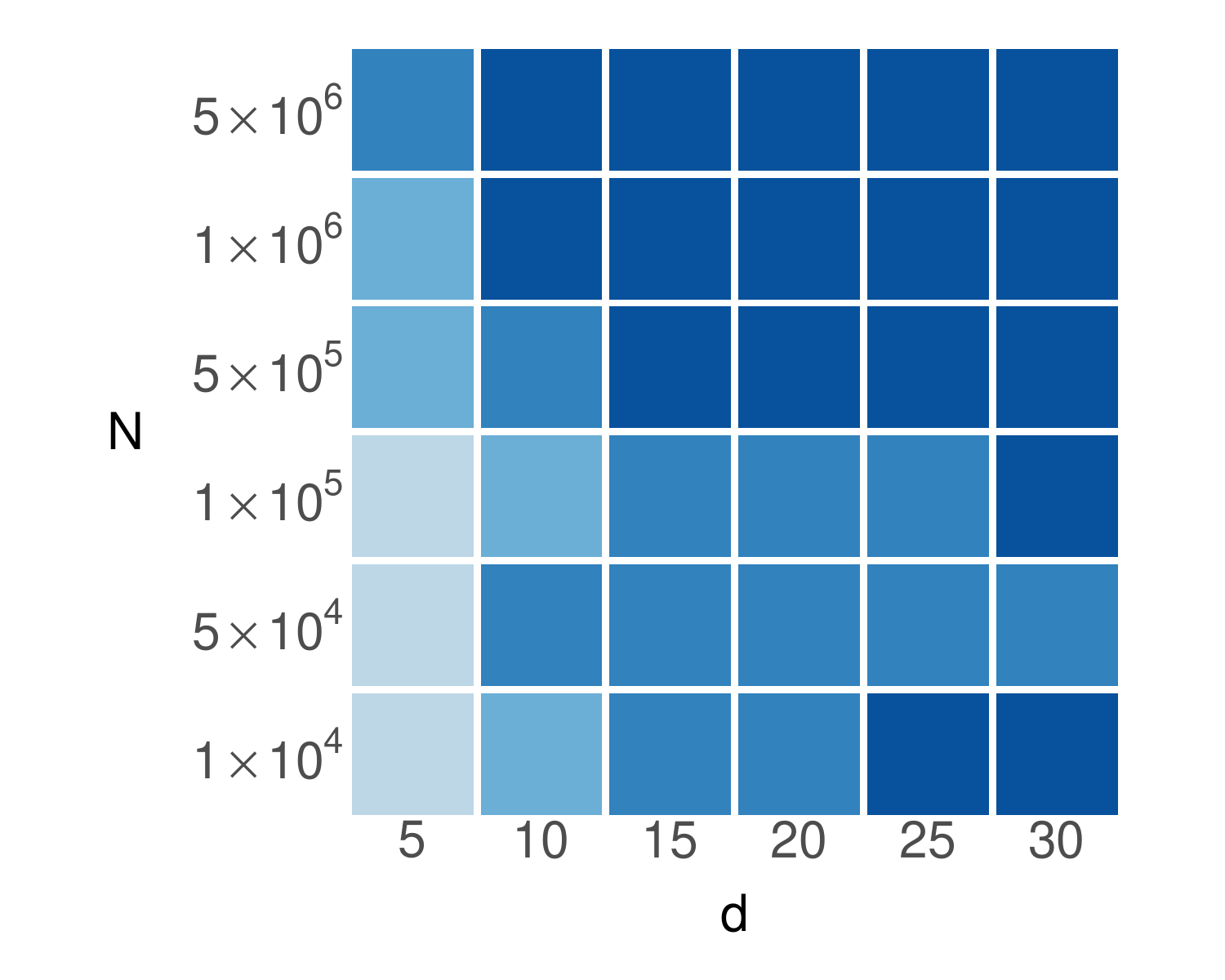}} \\ 
\midrule
{\renewcommand{\arraystretch}{1.05}%
\itshape\scriptsize
\begin{tabularx}{0.5\textwidth}{X}
\bf{Optimality Gap of Best Solution at Termination}\\[0.25em]
i.e., $(\objvalmax{}-\objvalmin{})/\objvalmax{}$, where $\objvalmax{}$ is the objective value of the best solution found at termination. A gap of 0.0\% means an algorithm has found the optimal solution and provided a proof of optimality within 6 hours.
\end{tabularx}} & 
\cell{c}{\includenewheatlegend{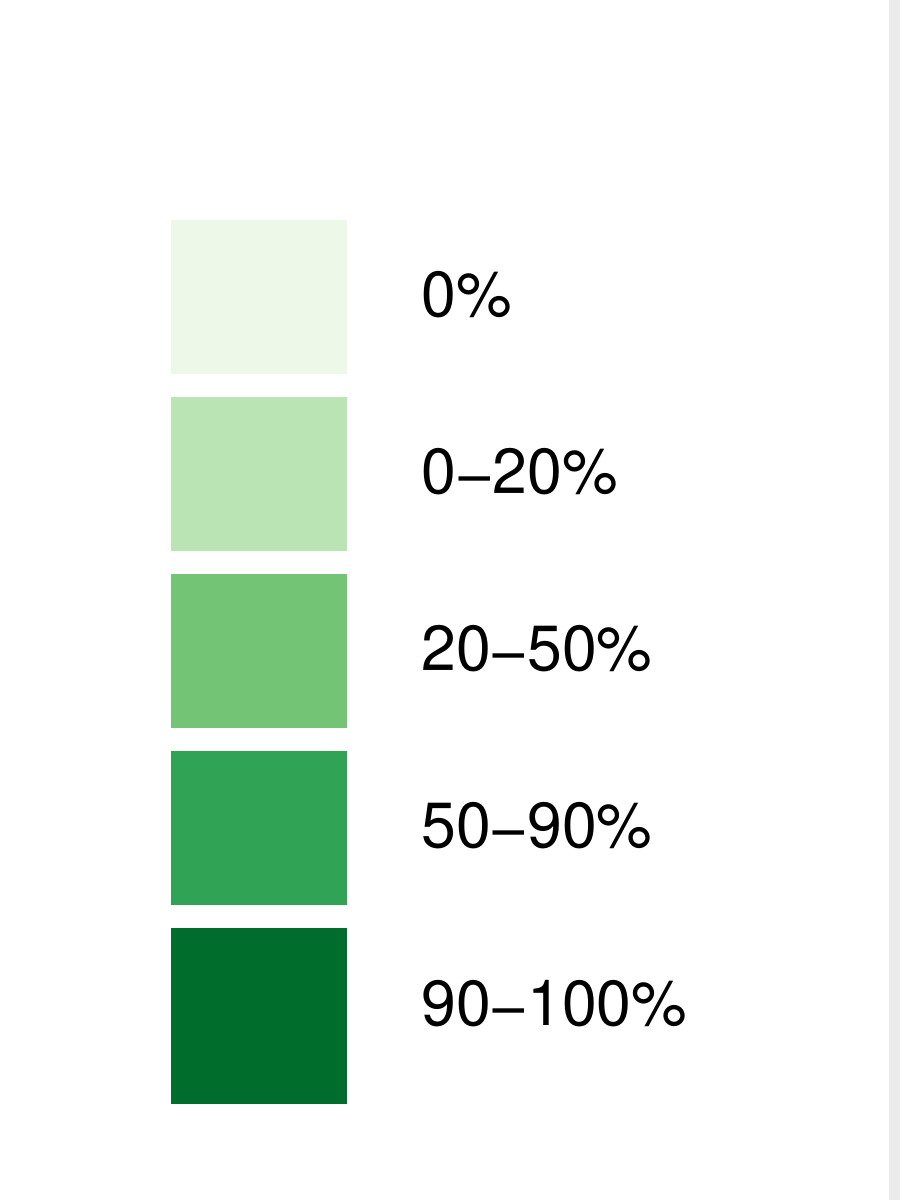}} &
\cell{c}{\includenewheatmap{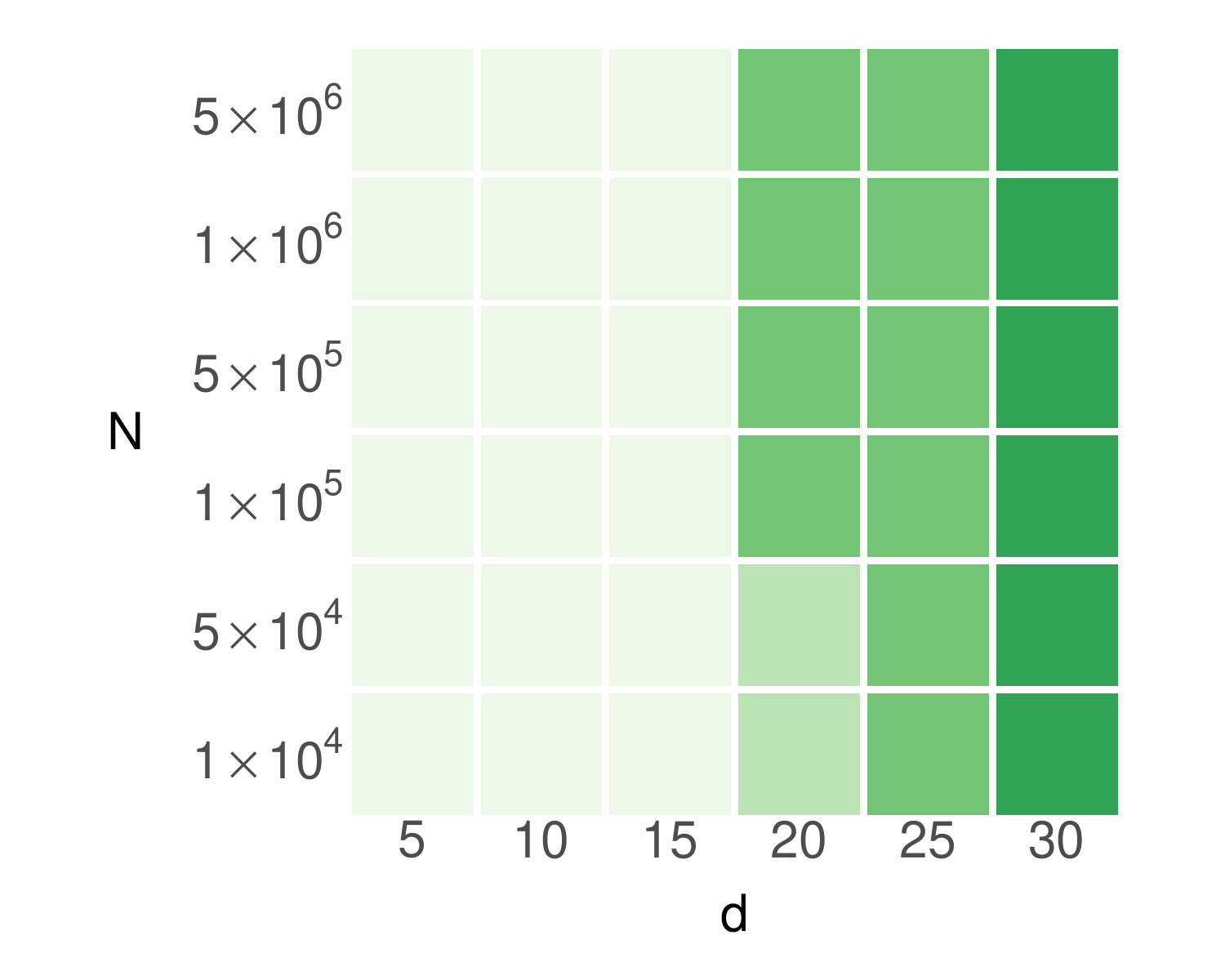}} & 
\cell{c}{\includenewheatmapNOYAXIS{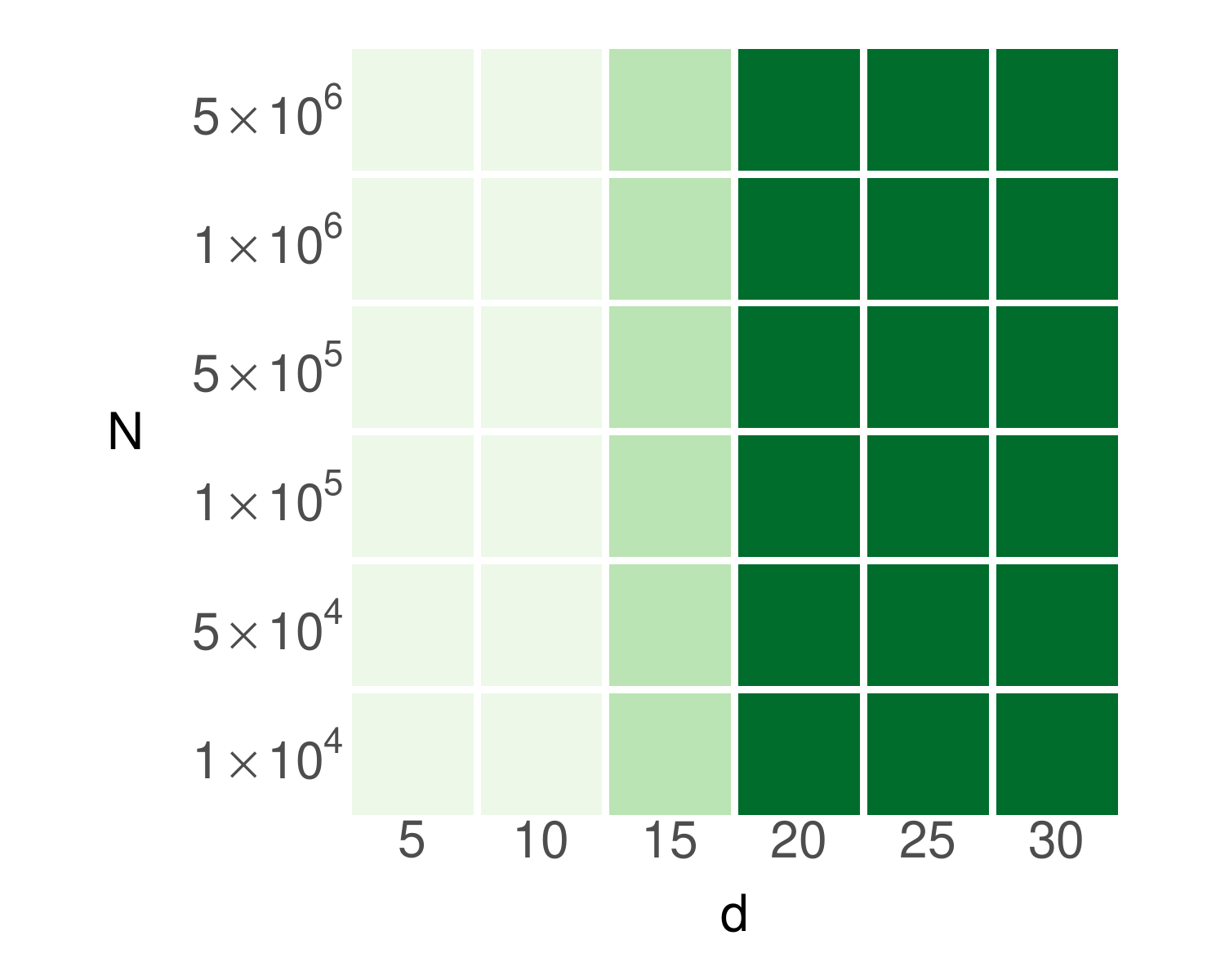}} & 
\cell{c}{\includenewheatmapNOYAXIS{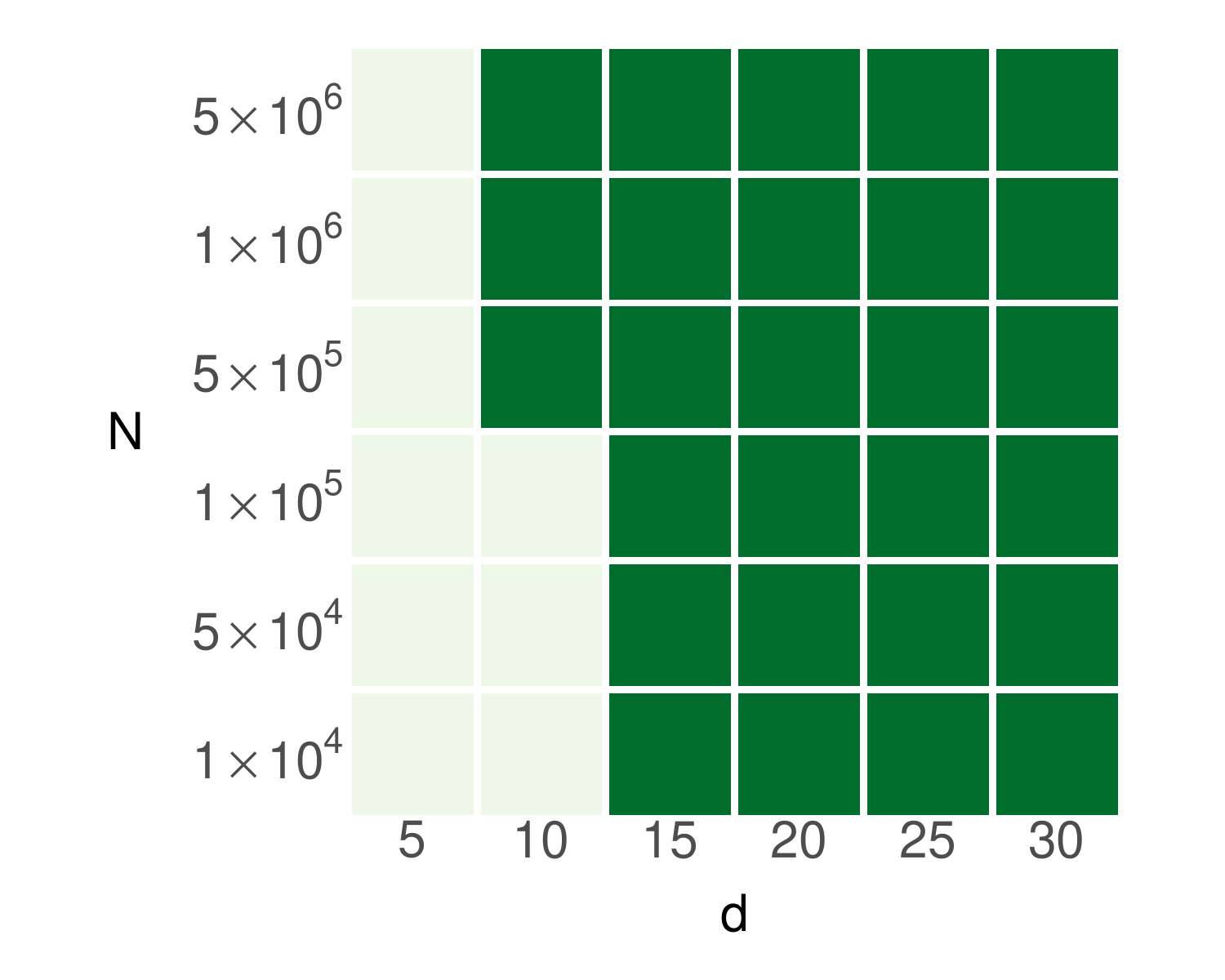}} \\ 
\midrule
{\renewcommand{\arraystretch}{1.05}%
\itshape\scriptsize
\begin{tabularx}{0.5\textwidth}{X}
\bf{\% Time Spent on Data-Related Computation}\\[0.25em]
i.e., the proportion of total runtime that an algorithm spends computing the value, gradient, or Hessian of the loss function.
\end{tabularx}} & 
\cell{c}{\includenewheatlegend{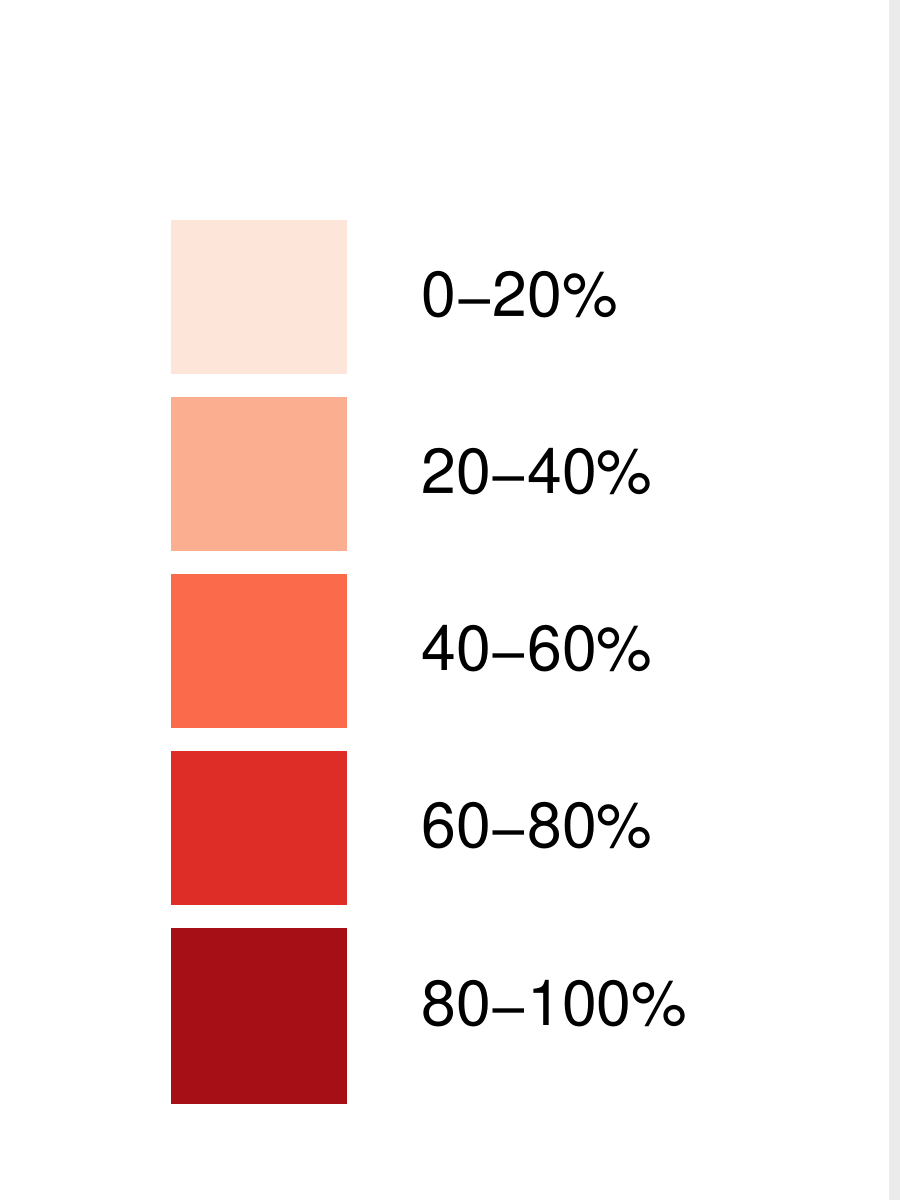}} &
\cell{c}{\includenewheatmap{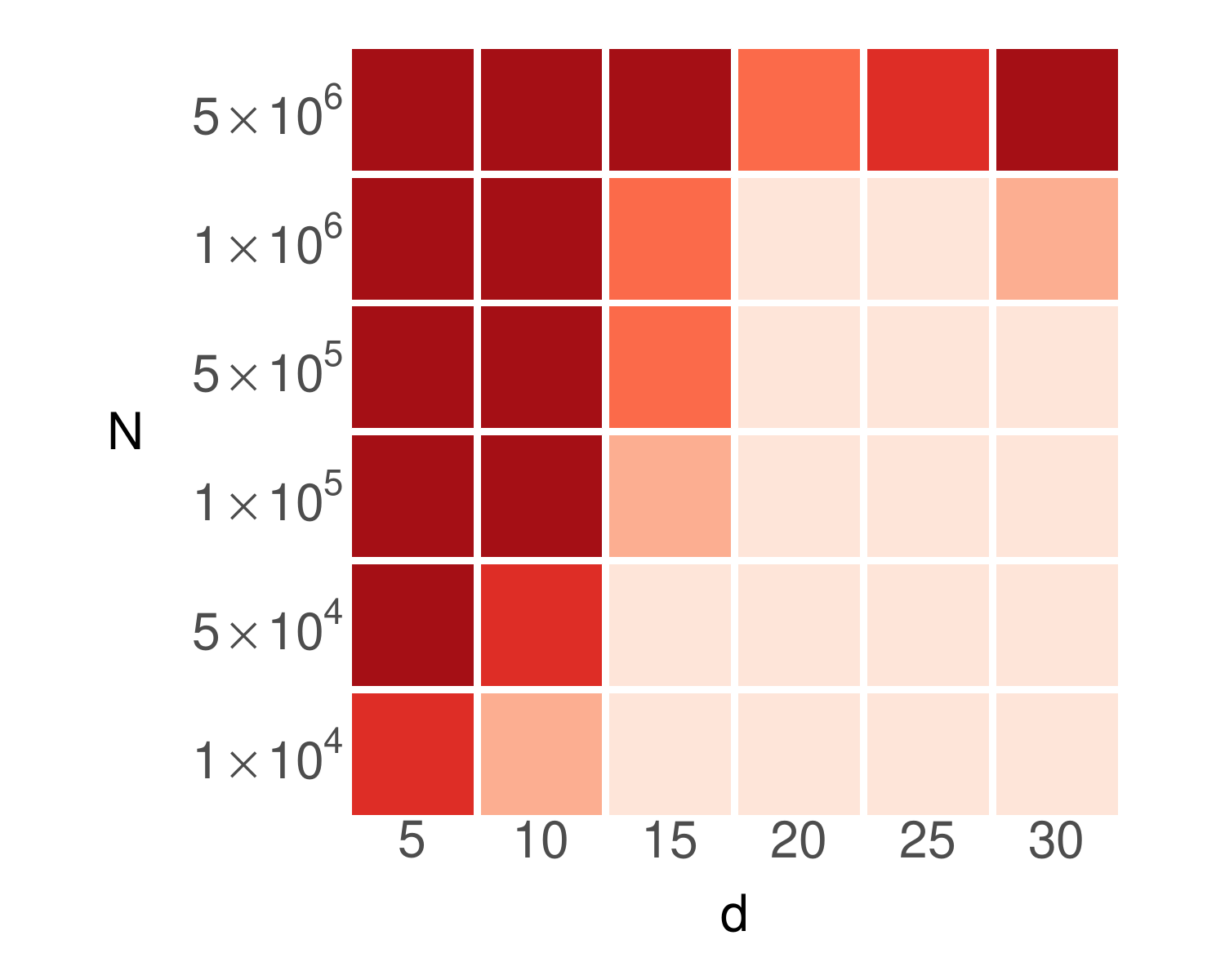}} & 
\cell{c}{\includenewheatmapNOYAXIS{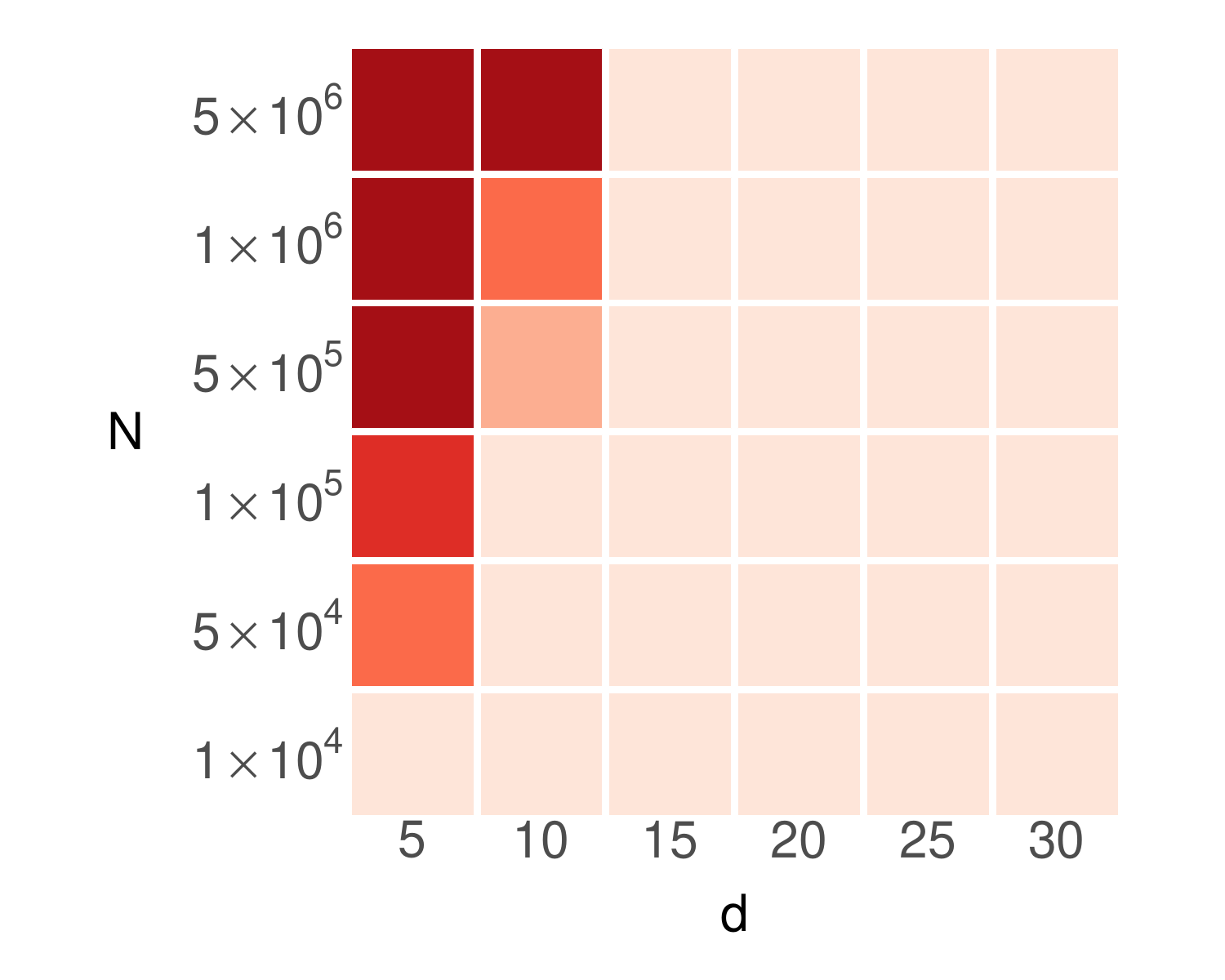}} & 
\cell{c}{\includenewheatmapNOYAXIS{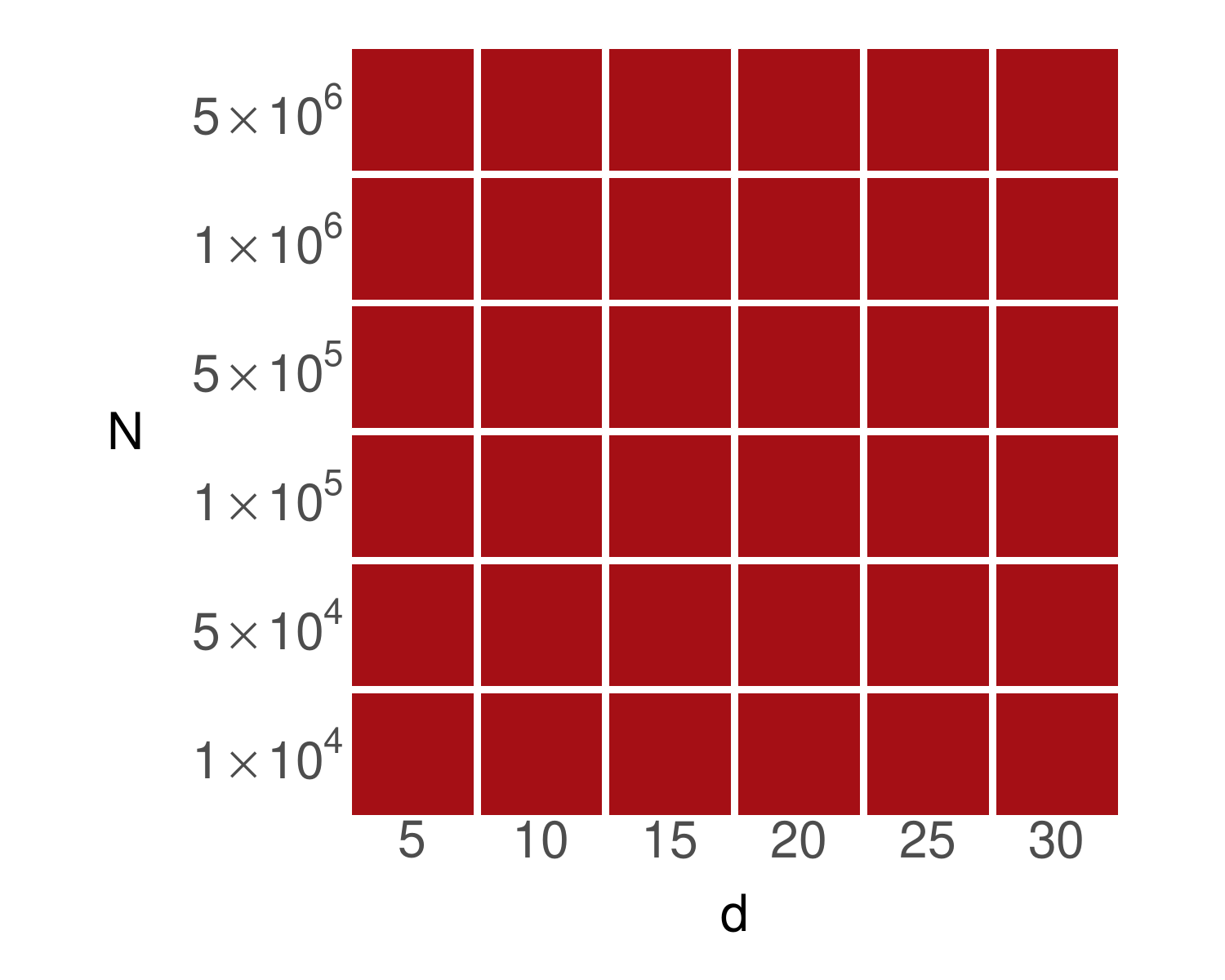}} \\ 

\bottomrule
\end{tabular}
}
\caption{Performance of \LCPA{}, \CPA{}, and a commercial MINLP solver on difficult instances of \RSMINLP{} for synthetic datasets with $d$ dimensions and $n$ samples (see Appendix \ref{Appendix::SimulationDetails} for details). 
\textsf{ActiveSetMINLP} fails to produce good risk scores on instances with large $n$ or $d$ as it struggles with data-related computation.
\CPA{} and \LCPA{} scale linearly in $\n{}$ when $d$ is fixed: if they solve an instance for a given $d$, then they can solve instances for larger $\n{}$ in $O(\n)$ additional time. \CPA{} stalls when $d \geq 15$ and returns a low-quality risk score when $d \geq 20$. In contrast, \LCPA{} consistently recovers a good model without stalling.
Results reflect the performance for a basic \LCPA{} implementation without the improvements in Section \ref{Sec::AlgorithmicImprovements}. We show results for two other MINLP algorithms in Appendix \ref{Appendix::SimulationDetails}.}
\label{Fig::MethodComparison}
\end{figure}

In Figure \ref{Fig::MethodComparison}, we show the performance of algorithms on difficult instances of the risk score problem for synthetic datasets with $d$ dimensions and $n$ samples (see Appendix \ref{Appendix::SimulationDetails} for details). We consider the following performance metrics: (i) the time to find a near-optimal solution; (ii) the optimality gap of the best solution at termination; and (iii) the proportion of time spent on data-related computation. Since all three MINLP algorithms behave similarly, we only show the best one in Figure \ref{Fig::MethodComparison} (i.e., \textsf{ActiveSetMINLP}), and include results for the others in Appendix \ref{Appendix::PerformanceComparison}.

As shown, \LCPA{} finds an optimal or near-optimal solution for almost all instances of the risk score problem, and pairs the solution with a small optimality gap. \CPA{} performs similarly to \LCPA{} on low-dimensional instances. On instances with $d \geq 15$, however, \CPA{} stalls after a few iterations and returns a highly suboptimal solution (i.e., a risk score with poor performance). In comparison, the MINLP algorithms can only handle instances with small $n$ or $d$. On larger instances, the solver is slowed down by operations that involve data-related computation, fails to converge within the 6-hour time limit and fails to recover a high-quality solution. Seeing how MINLP solvers are designed to solve a diverse set of optimization problems, we do not believe that they can identify and exploit the structure of the risk score problem in the same way as a cutting plane algorithm.

%% file: section_04_algorithmic_improvements.tex
In this section, we describe specialized techniques to improve the performance of the lattice cutting plane algorithm (\LCPA{}) on the risk score problem. They include:
\begin{itemize}

\item \emph{Polishing Heuristic}. We present a technique called discrete coordinate descent (\DCD{}; Section \ref{Sec::DCD}), which we use to polish integer solutions found by \LCPA{} (solutions satisfying the condition in Step~\ref{AlgStep::LCPAIntegerFeasible}). \DCD{} aims to improve the objective value of all integer solutions, which produces stronger upper bounds over the course of \LCPA{}, and reduces the time to recover a good risk score.

\item \emph{Rounding Heuristic}. We present a rounding technique called \SR{} (Section \ref{Sec::SequentialRounding}) to generate integer solutions. We use \SR{} to round real-valued solutions of the surrogate LP 
(which are solutions that satisfy the condition in Step \ref{AlgStep::LCPANonIntegerSolution}) and then polish the rounded solution with \DCD{}. Rounded solutions may improve the best solution found by \LCPA{}, producing stronger upper bounds and reducing the time to recover a good risk score.

\item \emph{Bound Tightening Procedure}. We design a fast procedure to strengthen bounds on the optimal values of the objective, loss, and number of non-zero coefficients called \ChainedUpdates{} (Section \ref{Sec::ChainedUpdates}). We call \ChainedUpdates{} whenever the solver updates the upper bound in Step \ref{AlgStep::LCPAUpdateUB} or the lower bound in Step \ref{AlgStep::LCPAUpdateLB}. \ChainedUpdates{} improves the lower bound, and reduces the optimality gap of the final risk score.

\end{itemize}

We present additional techniques to improve \LCPA{} in Appendix \ref{Appendix::AlgorithmicImprovements} such as an initialization procedure and techniques to reduce data-related computation.

\subsection{Discrete Coordinate Descent}
\label{Sec::DCD}

\emph{Discrete coordinate descent} (\DCD{}) is a technique to polish an integer solution (Algorithm~\ref{Alg::DCD}). It takes as input an integer solution $\lambdab = [\lambda_0, \ldots, \lambda_d]^\top \in \Lset$ and iteratively changes a single coordinate $j$ to attain an integer solution with a better objective value. The coordinate at each iteration is set to minimize the objective value, i.e., $j \in \argmin_{j'} \objval{\lambdab + \delta_{j'} e_{j'}}$.  

\DCD{} terminates once it can no longer strictly improve the objective value along any coordinate. This eliminates the potential of cycling, and thereby guarantees that the procedure will terminate in a finite number of iterations. The polished solution returned by \DCD{} satisfies a type of local optimality guarantee for discrete optimization problems. Formally, it is \emph{1-opt} with respect to the objective value, meaning that one cannot improve the objective value by changing any single coefficient \citep[see e.g.,][for a technique to find a 1-opt point for a different optimization problem]{park2015semidefinite}. 

In practice, the most expensive computation in \DCD{} is finding a step-size $\delta_j \in \Delta_j$ that minimizes the objective along coordinate $j$ (Step \ref{AlgStep::FindBestMove} of Algorithm~\ref{Alg::DCD}). We can significantly reduce this computation by using golden section search.  This approach requires $\n{}d\log_2|\Lset_j|$ flops per iteration compared to $\n{}d|\Lset_j|$ flops per iteration required by a brute force approach (i.e., which evaluates the loss for all $\lambda_j \in \Lset_j$).

\begin{algorithm}[t]
\caption{Discrete Coordinate Descent (\DCD{})}
\label{Alg::DCD}
\begin{algorithmic}[1]
\small\vspace{0.15em}
\INPUT 
\alginput{$(\xb_i, y_i)_{i=1}^\n$}{training data}
\alginput{$\Lset$}{coefficient set}
\alginput{$C_0$}{$\ell_0$ penalty parameter}
\alginput{$\lambdab \in \Lset$}{integer solution to \RSMINLP{}}
\alginput{$\J \subseteq \{0,\ldots,d\}$}{valid descent directions}
\vspace{-0.5em}\algrule{0.35\textwidth}{0.1pt}\vspace{-0.5em}
%\INITIALIZE
\Repeat{}
\State{$V \gets \objval{\lambdab}$}\Comment{objective value at current solution}
\For{$j \in \J$}
\State $\Delta_j \gets  \left\{ \delta \in \Z ~|~ \lambdab + \delta e_j  \in \Lset \right \}$ %
\Comment{list feasible moves along dim $j$}
\State $\delta_j \gets \argmin_{\delta \in \Delta_j} \objval{\lambdab + \delta}$
\Comment{find best move in dim $j$}\label{AlgStep::FindBestMove}
\State $v_j \gets V(\lambdab + \delta_j e_j)$%
\Comment{store objective value for best move in dim $j$}
\EndFor
\State $m \gets  \argmin_{j\in \J} v_j$ \Comment{descend along dim that minimizes objective}\label{AlgStep::StoreValueBestMove}
%\If{$v_m < V$}
\State $\lambdab \gets \lambdab  + \delta_m e_m$
%\State $\J \gets \{0,\ldots,d\} \setminus \{m\}$ \Comment{ignore dim $m$ on next iteration}
%\EndIf
\Until{$v_m \geq V$}
\Ensure $\lambdab$ \hfill solution that is 1-opt with respect to the objective of \RSMINLP{}
\end{algorithmic}
\end{algorithm}

In Figure \ref{Fig::PerformanceProfile1OPT}, we show how \DCD{} improves the performance of \LCPA{} when we use it to polish feasible solutions found by the MIP solver (i.e., the polishing is placed just after Step \ref{AlgStep::LCPAIntegerFeasible} of Algorithm \ref{Alg::LCPA}).

\begin{figure}[ht]
\centering
%\begin{tabular}{c}
%\setlength{\tabcolsep}{0pt}
%\includeAlgLegend{1OPT}{legend.pdf} \\ 
\begin{tabular}{@{}lr@{}}
%\includeAlgPlot{1OPT}{ub_vs_time.pdf} & \includeAlgPlot{1OPT}{gap_vs_time.pdf}
\includeAlgPlot{1OPT}{ub_vs_nodes.pdf} & \includeAlgPlot{1OPT}{gap_vs_nodes.pdf} 
\end{tabular}
%\end{tabular}
\caption{Performance profile of \LCPA{} in a basic implementation (black) and with \DCD{} (red). We use \DCD{} to polish every integer solution found by the MIP solver whose objective value is within 10\% of the current upper bound. We plot the number of total nodes processed of \LCPA{} (x-axis) against the upper bound (y-axis; left) and the optimality gap (y-axis; right). We mark iterations where \LCPA{} updates the incumbent solution. Results reflect performance on \RSMINLP{} for a synthetic dataset with $d = 30$ and $\n =$ 50,000 (see Appendix \ref{Appendix::SimulationDetails} for details).}
\label{Fig::PerformanceProfile1OPT}
\end{figure}

\subsection{Sequential Rounding}
\label{Sec::SequentialRounding}
\newcommand{\real}{\textrm{real}}

\SR{} (Algorithm \ref{Alg::SequentialRounding}) is a rounding heuristic to generate integer solutions for the risk score problem. In comparison to na\"ive rounding, which returns the closest rounding from a set of $2^{d+1}$ possible roundings, \SR{} returns a rounding that iteratively finds a local optimizer of the risk score problem. 

Given a real-valued solution $\lambdab^\text{real} \in \conv{\Lset}$, the procedure iteratively rounds one component (up or down) in a way that reduces the objective of \RSMINLP{}. On iteration $k$, it has already rounded $k$ components of $\lambdab^\text{real}$, and must round one of the remaining $d + 1 -k$ components to $\lceil \lambda_j^\text{real}\rceil$ or $\lfloor \lambda_j^\text{real}\rfloor$. To this end, it computes the objective of all feasible (component, direction)-pairs and chooses the best one. Formally, the minimization on iteration $k$ requires $2 \cdot (d + 1 - k)$ evaluations of the loss function. Thus, given that there are $d +1$ iterations, \SR{} terminates after $2\cdot\sum_{k=1}^{d} k = d(d + 1)$ evaluations of the loss function.

In Figure \ref{Fig::PerformanceProfileSR}, we show the impact of using \SR{} in \LCPA{}. Here, we apply \SR{} to the non-integer solution of \rslp{} when the lower bound changes (i.e., just after Step \ref{AlgStep::LCPASolveLP} of Algorithm \ref{Alg::LCPA}), then polish the rounded solution using \DCD{}. As shown, this strategy can reduce the time required for \LCPA{} to find a high-quality risk score, and attain a lower optimality gap. 

\begin{algorithm}[t]
\caption{\SR{}}
\label{Alg::SequentialRounding}
\begin{algorithmic}[1]
\small\vspace{0.15em}
\INPUT
\alginput{$(\xb_i, y_i)_{i=1}^\n$}{training data}
\alginput{$\Lset$}{coefficient set}
\alginput{$C_0$}{$\ell_0$ penalty parameter}
\alginput{$\lambdab \in \conv{\Lset}$}{non-integer infeasible solution from $\rslp{}$}
\vspace{-1.0em}\algrule{0.35\textwidth}{0.1pt}\vspace{-0.5em}
\State $\J^\real{} \gets \{j: \lambda_j \neq \round{\lambda_j} \}$ 
\Comment{index set of non-integer coefficients}
\Repeat{}

\State $\lambdab^{j, \text{up}} \gets (\lambda_1,\ldots,\ceil{\lambda_j},\ldots,\lambda_d)$ for all $j \in \J^\text{real}$

\State $\lambdab^{j, \text{down}} \gets (\lambda_1,\ldots,\floor{\lambda_j},\ldots,\lambda_d)$ for all $j \in \J^\text{real}$

\State $v^\text{up} \gets \min_{j\in\J^\text{real}} V(\lambdab^{j, \text{up}})$

\State $v^\text{down} \gets \min_{j\in\J^\text{real}} V(\lambdab^{j, \text{down}})$ 
\If{$v^\text{up} < v^\text{down}$}

\State $k \gets \argmin_{j \in \J^\text{real}} V(\lambdab^{j,\text{up}})$ and $\lambda_k \gets \ceil{\lambda_k}$ 

\Else

\State $k \gets \argmin_{j \in \J^\text{real}} V(\lambdab^{j,\text{down}})$ and $\lambda_k \gets \floor{\lambda_k}$ 

\EndIf

\State $\J^\text{real} \gets \J^\text{real} \setminus \{k\}$

\Until{$\J^\text{real} = \emptyset$}

\Ensure $\lambdab \in \Lset$ \hfill{integer solution}

\end{algorithmic}
\end{algorithm}

\begin{figure}[ht]
\centering
%\begin{tabular}{c}
%\setlength{\tabcolsep}{0pt}
%\includeAlgLegend{SR}{legend.pdf} \\ 
%\begin{tabular}{lr}\setlength{\tabcolsep}{0pt} \includeAlgPlot{SR}{ub_vs_time.pdf} & \includeAlgPlot{SR}{gap_vs_time.pdf} \end{tabular}
\begin{tabular}{@{}lr@{}}
%\includeAlgPlot{SRP}{ub_vs_time.pdf} & \includeAlgPlot{SRP}{gap_vs_time.pdf} 
\includeAlgPlot{SRP}{ub_vs_nodes.pdf} & \includeAlgPlot{SRP}{gap_vs_nodes.pdf}
\end{tabular}
\caption{Performance profile of \LCPA{} in a basic implementation (black) and with \SR{} and \DCD{} polishing (red). We call \SR{} to round non-integer solutions to \rslp{} in Step \ref{AlgStep::LCPANonIntegerSolution}, and then polish the integer solution with \DCD{}. We plot large points to show when \LCPA{} updates the incumbent solution. Results reflect performance on \RSMINLP{} for a synthetic dataset with $d = 30$ and $\n =$ 50,000 (see Appendix \ref{Appendix::SimulationDetails} for details). Here, \SR{} and \DCD{} reduce the upper bound and optimality gap of \LCPA{} compared to a basic implementation.}
\label{Fig::PerformanceProfileSR}
\end{figure}

%\SR{} can be viewed as a special case of \DCD{}. In particular, instead of conducting a line search over all feasible values at each iteration, \SR{} rounds each real-valued component to a nearby integer value. Likewise, instead of terminating at a 1-OPT point, \SR{} terminates in $d+1$ iterations after all components are rounded. In that case, one can see why \DCD{} might perform better than \SR{} since it considers a larger set of integer coefficients and is not restricted to only $d+1$ iterations. In comparison to \DCD{}, \SR{} terminates after a fixed number of steps and can be accelerated with a subsampling technique presented in Appendix \ref{Sec::Subsampling}. 

\subsection{Chained Updates}
\label{Sec::ChainedUpdates}

We describe a fast bound tightening technique called \ChainedUpdates{} (Algorithm \ref{Alg::ChainedUpdates}). This technique iteratively bounds the optimal values of the objective, loss, and $\ell_0$-norm by iteratively setting the values of $\objvalmin$, $\objvalmax$, $\lossmin$, $\lossmax$, and $\zeronormmax$ in \rslp{}. Bounding these quantities over the course of B\&B restricts the search region without discarding the optimal solution, thereby improving the lower bound and reducing the optimality gap. 

\paragraph{Initial Bounds on Objective Terms}

We initialize \ChainedUpdates{} with values of $\objvalmin$, $\objvalmax$, $\lossmin$, $\lossmax$, and $\zeronormmax$ that can be computed using only the training data $(\xb_i, y_i)_{i=1}^\n$ and the coefficient set $\Lset$. We start with Proposition \ref{Prop::LossBounds}, which provides initial values for $\lossmin$ and $\lossmax$ using the fact that the coefficient set $\Lset$ is bounded. %(see Appendix \ref{Appendix::Proofs} for a proof).

\begin{proposition}[Bounds on Logistic Loss over a Bounded Coefficient Set]
\label{Prop::LossBounds}
Given a training dataset $(\xb_i, y_i)_{i=1}^\n$ where $\xb_i \in \R^d$ and $y_i \in \{\pm 1\}$ for $i = 1,\ldots,\n$, consider the normalized logistic loss of a linear classifier with coefficients $\lambdab$:
$$\lossfun{\lambdab} = \logloss{\lambdab}.$$
If the coefficients belong to a bounded set $\Lset$, then the value of the normalized logistic loss must obey $\lossfun{\lambdab} \in [\lossmin{}, \lossmax{}] \txforall{} \lambdab \in \Lset,$ where:
\begin{align*}
\lossmin{} &= \frac{1}{\n} \sum_{i: y_i = +1} \log{(1+ \exp(-s_i^{\max}))} + \frac{1}{\n} \sum_{i: y_i = -1} \log{(1+ \exp(s_i^{\min}))},\\
\lossmax{} &=  \frac{1}{\n} \sum_{i: y_i = +1} \log{(1+ \exp(-s_i^{\min}))} + \frac{1}{\n} \sum_{i: y_i = -1} \log{(1+ \exp(s_i^{\max}))},\\
%\intertext{and,}
s_i^{\min} &= \min_{\lambdab \in \Lset} \dotprod{\lambdab}{\xb_i}  \for i = 1,\ldots, \n, \\
s_i^{\max} &= \max_{\lambdab \in \Lset} \dotprod{\lambdab}{\xb_i}  \for i = 1,\ldots, \n.
\end{align*}
\end{proposition}
%\begin{proof}See Appendix \ref{Appendix::Proofs}.\end{proof}
%
The value of $\lossmin$ in Proposition \ref{Prop::LossBounds} represents the ``best-case" loss in a separable setting where we assign each positive example its maximal score $s_i^{\max}$, and each negative example its minimal score $s_i^{\min}$. Conversely, $\lossmax$ represents the ``worst-case" loss when we assign each positive example its minimal score $s_i^{\min}$ and each negative example its maximal score $s_i^{\max}$.
% Both $\lossmin$ and $\lossmax$ can be computed in O($\n$) flops using only the training data and the coefficient set by evaluating $s_i^{\min}$ and $s_i^{\max}$ as follows:
% %
% \begin{align}
% %
% s_i^{\min} &= \min_{\lambdab \in \Lset} \dotprod{\lambdab}{\xb_i} = \sum_{j=0}^d \indic{x_{ij} > 0} x_{ij} \minlambda{j} + \indic{x_{ij} < 0} x_{ij} \maxlambda{j}, \label{Eq::MaxScoreComputation}\\
% %
% s_i^{\max} &= \max_{\lambdab \in \Lset} \dotprod{\lambdab}{\xb_i}  = \sum_{j=0}^d \indic{x_{ij}>0} x_{ij} \maxlambda{j} + \indic{x_{ij} < 0} x_{ij} \minlambda{j}. \label{Eq::MinScoreComputation}
% %
% \end{align}
%In practice, the $\lossmin$ bound is useful, whereas the $\lossmax$ bound is too loose and likely to be discarded. For instance, if we have $\bf{0} \in \Lset$,  then $\lossfun{\bf{0}}$ yields a much tighter upper bound than $\lossmax$. 
%
We initialize the bounds on the number of non-zero coefficients $R$ to $\in \{0,\ldots,d\}$, trivially. In some cases, these bounds may be stronger due to operational constraints (e.g., we can set $R \in \{0,\ldots, 5\}$ if models are required to use $\leq5$ features). %If so, then the values of $\lossmin$ and $\lossmax$ can be further reduced since $s_i^{\min}$ and $s_i^{\max}$ are bounded by the number of non-zero coefficients. Once again, $s_i^{\min}$ (or $s_i^{\max}$) can still be computed efficiently in O($\n$) flops by choosing the $\zeronormmax$ smallest (or largest) terms in the right-hand side of Equation \eqref{Eq::MaxScoreComputation} (or \ref{Eq::MinScoreComputation}). 
Having initialized $\lossmin$, $\lossmax$, $\zeronormmin$ and $\zeronormmax$, we set the bounds on the optimal objective value as $\objvalmin = \lossmin + C_0\zeronormmin$ and $\objvalmax = \lossmax + C_0\zeronormmax$, respectively.

\paragraph{Dynamic Bounds on Objective Terms}

In Propositions \ref{Prop::ZeroNormUB} to \ref{Prop::LossLB}, we present bounds that can use information from the solver in \LCPA{} to strengthen the values of $\lossmin$, $\lossmax$, $\zeronormmax$, $\objvalmin$ and $\objvalmax$ (see Appendix \ref{Appendix::Proofs} for proofs).
\begin{proposition}[Upper Bound on Optimal Number of Non-Zero Coefficients]
\label{Prop::ZeroNormUB}

Given an upper bound on the optimal value $\objvalmax \geq \objvalfun{\lambdab^\opt}$, and a lower bound on the optimal loss $\lossmin \leq \lossfun{\lambdab^\opt}$, the optimal number of non-zero coefficients is at most 
%$\zeronormmax \geq \zeronorm{\lambdab^\opt}$ where 
$$\zeronormmax = \floor{\frac{\objvalmax - \lossmin}{C_0}}.$$

\end{proposition}
\begin{proposition}[Upper Bound on Optimal Loss]
\label{Prop::LossUB}

Given an upper bound on the optimal value $\objvalmax \geq \objvalfun{\lambdab^\opt}$, and a lower bound on the optimal number of non-zero coefficients $\zeronormmin \leq \zeronorm{\lambdab^\opt}$, the optimal loss is at most 
%$\lossmax \geq \lossfun{\lambdab^\opt}$ where 
$$\lossmax = \objvalmax - C_0 \zeronormmin.$$
\end{proposition}
\begin{proposition}[Lower Bound on Optimal Loss]
\label{Prop::LossLB}
Given a lower bound on the optimal value $\objvalmin \leq \objvalfun{\lambdab^\opt}$, and an upper bound on the optimal number of non-zero coefficients $\zeronormmax \geq \zeronorm{\lambdab^\opt}$, the optimal loss is at least %$\lossmin \leq \lossfun{\lambdab^\opt}$ where 
$$\lossmin = \objvalmin - C_0 \zeronormmax.$$
\end{proposition}

\paragraph{Implementation}

In Algorithm \ref{Alg::ChainedUpdates}, we present a bound-tightening procedure that uses the results of Propositions \ref{Prop::ZeroNormUB} to \ref{Prop::LossLB} to strengthen the values of $\objvalmin$, $\objvalmax$, $\lossmin$, $\lossmax$, and $\zeronormmax$ in \rslp{}.

\begin{algorithm}[t]
\caption{\ChainedUpdates{}}
\label{Alg::ChainedUpdates}
\begin{algorithmic}[1]\small
\vspace{0.15em}
\INPUT
\alginput{$C_0$}{$\ell_0$ penalty parameter}
\alginitialize{$\objvalmin$, $\objvalmax$, $\lossmin$, $\lossmax$, $\zeronormmin$, $\zeronormmax$}{initial bounds on $\objvalfun{\lambdab^\opt}$, $\lossfun{\lambdab^\opt}$ and $\zeronorm{\lambdab^\opt}$}
\vspace{-0.5em}\algrule{0.5\textwidth}{0.5pt}\vspace{-0.5em}
\Repeat{}
\State $\objvalmin \gets  \max\left(\objvalmin, ~ {\lossmin + C_0 \zeronormmin}\right)$ 
\Comment{update lower bound on $\objvalfun{\lambdab^\opt}$} 
\label{AlgStep::UpdateObjvalMin}

\State $\objvalmax \gets \min\left(\objvalmax, ~ {\lossmax + C_0 \zeronormmax}\right)$ 
\Comment{update upper bound on $\objvalfun{\lambdab^\opt}$} 
\label{AlgStep::UpdateObjvalMax}

\State $\lossmin \gets \max\left(\lossmin, ~\objvalmin - C_0 \zeronormmax \right)$ 
\Comment{update lower bound on $\lossfun{\lambdab^\opt}$} 
\label{AlgStep::UpdateLossMin}

\State $\lossmax \gets \min\left(\lossmax, ~\objvalmax - C_0 \zeronormmin \right)$ 
\Comment{update upper bound on $\lossfun{\lambdab^\opt}$} 
\label{AlgStep::UpdateLossMax}

\State $\zeronormmax  \gets \min\left(\zeronormmax, ~ \floor{\frac{\objvalmax - \lossmin}{C_0}}\right)$  
\Comment{update upper bound on $\zeronorm{\lambdab^\opt}$} 
\label{AlgStep::ZeroNormMax}

\Until{there are no more bound updates due to Steps \ref{AlgStep::UpdateObjvalMin} to \ref{AlgStep::ZeroNormMax}.}
\vspace{0.25em}
\Ensure $\objvalmin$, $\objvalmax$, $\lossmin$, $\lossmax$, $\zeronormmin$, $\zeronormmax$
\vspace{0.25em}
\end{algorithmic}
\end{algorithm}

\begin{figure}[htbp]
\centering
\includegraphics[trim=0.25in 1.25in 0.25in 1.0in, clip, width=0.6\textwidth]{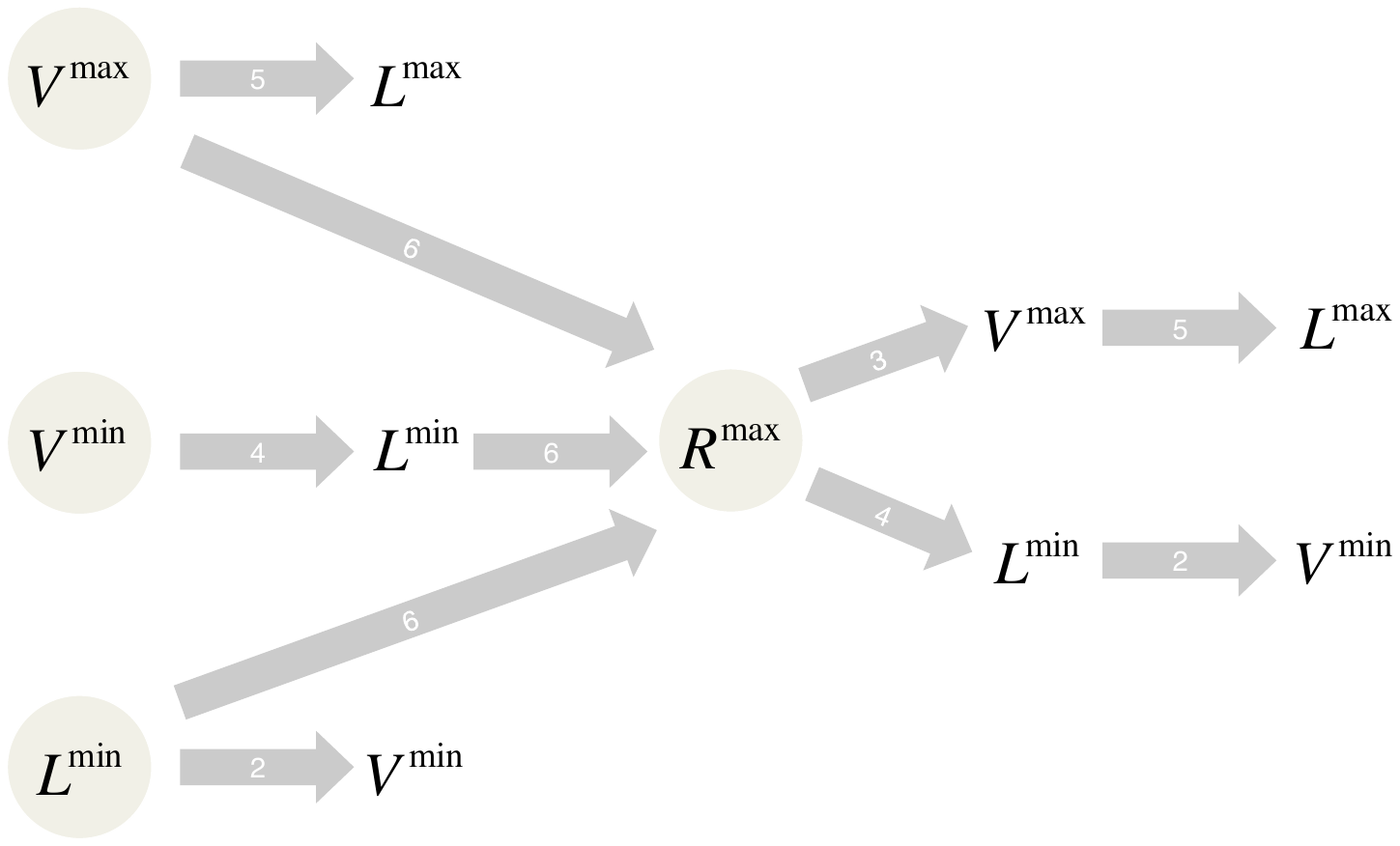}
\caption{All possible ``chains" of updates in \ChainedUpdates{}. Circles represent ``source" terms that can be updated by \LCPA{} to trigger \ChainedUpdates{}. The path from each source term shows all bounds that can be updated by the procedure. The number in each arrow references the update step in Algorithm \ref{Alg::ChainedUpdates}.}
\label{Fig::AllPossibleUpdateChains}
\end{figure}

Propositions \ref{Prop::ZeroNormUB} to \ref{Prop::LossLB} impose dependencies between $\objvalmin$, $\objvalmax$, $\lossmin$, $\lossmax$, $\zeronormmin$ and $\zeronormmax$ that may produce a complex ``chain" of updates. As shown in Figure \ref{Fig::AllPossibleUpdateChains}, \ChainedUpdates{} can update multiple terms, and may update the same term more than once. Consider a case where we call \ChainedUpdates{} after \LCPA{} improves $\objvalmin$. Say the procedure updates $\lossmin$ in Step \ref{AlgStep::UpdateLossMin}. If \ChainedUpdates{} updates $\zeronormmax$ in Step \ref{AlgStep::ZeroNormMax}, then it will also update $\objvalmax$, $\lossmin$, $\lossmax$, and $\objvalmin$. However, if \ChainedUpdates{} does not update $\zeronormmax$ in Step \ref{AlgStep::ZeroNormMax}, then it will not update $\objvalmax$, $\lossmin$, $\lossmax$, $\objvalmin$ and terminate. 

Considering these dependencies, Algorithm \ref{Alg::ChainedUpdates} applies Propositions \ref{Prop::ZeroNormUB} to \ref{Prop::LossLB} until it can no longer improve $\objvalmin$, $\objvalmax$, $\lossmin$, $\lossmax$ or $\zeronormmax$. This ensures that \ChainedUpdates{} will return its strongest possible bounds regardless of the term that was first updated.

In our implementation, we call \ChainedUpdates{} whenever \LCPA{} improves $\objvalmax$ or $\objvalmin$ (i.e., Step \ref{AlgStep::LCPAUpdateUB} or Step \ref{AlgStep::LCPAUpdateLB} of Algorithm \ref{Alg::LCPA}). If \ChainedUpdates{} improves any bounds, we pass this information back to the solver by updating the bounds on the auxiliary variables in the \rslp{} (Definition \ref{Def::RSLP}). As shown in Figure \ref{Fig::PerformanceChainUpdates}, this technique can considerably improve the lower bound and optimality gap over the course of \LCPA{}. 

\begin{figure}[b]
\centering
\begin{tabular}{@{}lr@{}}
\includeAlgPlot{BOUNDS}{lb_vs_nodes.pdf} & \includeAlgPlot{BOUNDS}{gap_vs_nodes.pdf}
\end{tabular}
\caption{Performance profile of \LCPA{} in a basic implementation (black) and with \ChainedUpdates{} (red). Results reflect performance on a \RSMINLP{} instance for a synthetic dataset with $d = 30$ and $\n =$ 50,000 (see Appendix \ref{Appendix::SimulationDetails}).}
\label{Fig::PerformanceChainUpdates}
\end{figure}

%% file: section_05_experiments.tex
In this section, we compare the performance of methods to create risk scores. We have three goals: (i) to benchmark the performance and computation of our approach on real-world datasets; (ii) to highlight pitfalls of traditional approaches used in practice; and (iii) to present new approaches that address the pitfalls of traditional approaches.

\subsection{Setup}
\label{Sec::ExperimentalSetup}

We considered 6 publicly available datasets shown in Table \ref{Table::ExperimentalDatasets}. We chose these datasets to see how methods are affected by factors such as class imbalance, the number of features, and feature encoding. For each dataset, we fit risk scores using \RiskSLIM{} and 6 baseline methods that post-processed the coefficients of the best logistic regression model built using Lasso, Ridge or Elastic Net. We used each method to fit a risk score with small integer coefficients $\lambda_j \in \{-5,\ldots,5\}$ that obeys the model size constraint $\zeronorm{\lambdab} \leq \zeronormmax{}$. We benchmarked each method for target model sizes $\zeronormmax{} \in \{2,\ldots, 10\}.$
\begin{table}[htbp]
%\tableformat{}
\resizebox{\textwidth}{!}{
\begin{tabular}{@{\;}l@{}r@{\;\;\;}r@{\;\;\;}r@{\;\;\;}l@{\;\;}l@{\;}}
\toprule

\textbf{Dataset}  & 
$\n$ & 
$d$ & 
$\textrm{Pr}(y_i=1)$ & 
\textbf{Conditions for $y_i=1$} & 
\textbf{Reference}

\\ \toprule 

\textds{income} & 
32,561 & 
36 & 
24.1\% &
person in 1994 US census earns over \$50,000 & 
\citet{kohavi1996scaling}

\\ \midrule

\textds{mammo} & 
961 & 
14 &
46.3\% &
person has breast cancer &
\citet{elter2007prediction}

\\ \midrule

\textds{mushroom} & 
8,124 & 
113 & 
48.2\% &
mushroom is poisonous  & 
\citet{schlimmer1987concept} 

\\ \midrule 

\textds{rearrest} & 
22,530 & 
48 & 
59.0\% &
person is arrested after release from prison & 
\citet{ustun2015recidivism}

\\ \midrule

\textds{spambase} & 
4,601 & 
57 & 
39.4\%&
e-mail is spam & 
\citet{cranor1998spam}

\\ \midrule 

\textds{telemarketing}\;\; & 
41,188 & 
57 & 
11.3\% &
person opens bank account after marketing call & 
\citet{moro2014data} 

\\ \bottomrule
\end{tabular}
}
\caption{Datasets used in Section \ref{Sec::Experiments}. All datasets are available on the UCI repository \citep{bache2013uci}, other than  {\scriptsize\texttt{rearrest}} which must be requested from ICPSR. We processed each dataset by dropping examples with missing values, and by binarizing categorical variables and some real-valued variables. We provide processed datasets and the code to process {\scriptsize\texttt{rearrest}} at \href{http://github.com/ustunb/risk-slim}{http://github.com/ustunb/risk-slim}.
}
\label{Table::ExperimentalDatasets}
\end{table}

\paragraph{RiskSLIM}

We formulated an instance of \RSMINLP{} with the constraints: $\lambda_0 \in \{-100,\ldots,100\}$, $\lambda_j \in \{-5, \ldots, 5\}$, and $\zeronorm{\lambdab} \leq \zeronormmax{}$. We set the trade-off parameter to a small value $C_0 = 10^{-6}$ to recover the sparsest model among equally accurate models (see Appendix \ref{Appendix::SmallRegularizationParameter}). We solved each instance for at most 20 minutes on a 3.33 GHz CPU with 16 GB RAM using CPLEX 12.6.3 \citep{cplex}. %Our \LCPA{} implementation includes the improvements in Section \ref{Sec::AlgorithmicImprovements} and is available online at \href{https://github.com/ustunb/risk-slim}{http://github.com/ustunb/risk-slim}.

\paragraph{Penalized Logistic Regression}

\PLR{} is the best logistic regression model produced over the full regularization path using a weighted combination of the $\ell_1$ and $\ell_2$ penalties (i.e., the best model produced by Lasso, Ridge or Elastic Net). We train \PLR{} models using the \textsf{glmnet} package of \cite{friedman2010glmnet}. The coefficients of each model are the solution to the optimization problem:
{\small
\begin{align*}
    \min_{\lambdab \in \R^{d+1}} \quad \frac{1}{2\n{}} \sum_{i=1}^\n{} \log(1+\exp(-\dotprod{\lambdab}{y_i\xb_i})) +  \gamma \cdot \left ( \alpha \vnorm{\lambdab}_1 +  (1 - \alpha){\vnorm{\lambdab}^2_2}\right)
\end{align*}%
}%
where $\alpha \in [0, 1]$ is the elastic-net mixing parameter and $\gamma \geq 0$ is a regularization penalty. We trained 1,100 \PLR{} models by choosing 1,100 combinations of ($\alpha, \gamma$): 11 values of $\alpha \in \{0.0, 0.1,\ldots, 0.9, 1.0\}$ $\times$ 100 values of $\gamma$ (chosen automatically by \textsf{glmnet} for each $\alpha$). This free parameter grid produces 1,100 \PLR{} models that include models obtained by: (i) Lasso ($\ell_1$-penalty), which corresponds to \PLR{} when $\alpha = 1.0$; (ii) Ridge ($\ell_2$-penalty), which corresponds to \PLR{} when $\alpha = 0.0$; (iii) standard logistic regression, which corresponds to \PLR{} when $\alpha = 0.0$ and $\gamma$ is small.

\paragraph{Traditional Approaches}
While there is considerable variation in how risk scores are developed in practice, many researchers follow a two-step approach: (i) fit a sparse logistic regression model with real-valued coefficients; (ii) convert it into a risk score with integer coefficients. We consider three methods that adopt this approach. Each method first trains a \PLR{} model (i.e., the one that maximizes the 5-CV AUC and obeys the model size constraint), and then converts it into a risk score by applying a common rounding heuristic:
\begin{itemize}

\item \mPLRRd{} (Rounding): We round each coefficient to the nearest integer in $\{-5\ldots5\}$ by setting $\lambda_j \leftarrow \round{\min(\max(\lambda_j, - 5), 5)}$, and round the intercept as $\lambda_0 \leftarrow \round{\lambda_0}$.

\item \mPLRUnit{} (Unit Weighting): We round each coefficient to $\pm 1$ as $\lambda_j \gets \text{sign}(\lambda_j)\indic{\lambda_j \neq 0}$. Unit weighting is a common heuristic in medicine and criminal justice \citep[see e.g.,][]{antman2000timi,kessler2005world,usdoj2005riskclassification,duwe2016sacrificing}, and sometimes called the \emph{Burgess method} \citep[as it was first proposed by][]{burgess1928factors}.

\item \mPLRRsRd{} (Rescaled Rounding) We first rescale coefficients so that the largest coefficient is $\pm 5$, then round each coefficient to the nearest integer (i.e., $\lambda_j  \rightarrow \round{\gamma \lambda_j}$ where $\gamma = 5 / \max_{j}|\lambda_j|$).  Rescaling is often used to avoid rounding small coefficients to zero, which happens when $|\lambda_j|<0.5$ \citep[see e.g.,][]{le1993new}.
%,goel2015precinct}.

\end{itemize}
%
%These methods do not reflect significant human input used in risk score development ( e.g., domain experts will perform preliminary feature selection, round coefficients, or choose a scaling factor before rounding, manually and without validation).

\paragraph{Pooled Approaches}
%1'100 PLR Models -> [Rounding + Post-Processing] -> 1'100 Risk Scores -> [Choose Instance that Maximizes 5-CV AUC] -> Risk Score
We also propose three new methods that use a \emph{pooling} strategy and the loss-minimizing heuristics from Section \ref{Sec::AlgorithmicImprovements}. Each method generates a pool of \PLR{} models with real-valued coefficients, applies the same post-processing procedure to each model in the pool, then selects the best risk score among feasible risk scores. The methods include:
\begin{itemize}[]%[label={},leftmargin=0em]

\item \PLRCR{} (Pooled \PLR{} + Rounding): We fit a pool of 1,100 models using \PLR{}. For each model in the pool, we round each coefficient to the nearest integer in $\{-5,\ldots, 5\}$ by setting $\lambda_j \leftarrow \round{\min(\max(\lambda_j, - 5), 5)}$, and round the intercept as $\lambda_0 \leftarrow \round{\lambda_0}$.

\item \PLRCRDCD{} (Pooled \PLR{} + Rounding + Polishing): We fit a pool of 1,100 models using \PLRCR{}. For each model in the pool, we polish the rounded coefficients using \DCD{}.

\item \PLRSRDCD{} (Pooled \PLR{} + Sequential Rounding + Polishing): We fit a pool of 1,100 models using \PLR{}. For each model in the pool, we round the coefficients using \SR{} and then polish the rounded coefficients using \DCD{}.

\end{itemize}
To ensure that the polishing step in \PLRCRDCD and \PLRSRDCD{} does not increase the number of non-zero coefficients (which would violate the model size constraint), we run \DCD{} only on the set $\{j ~|~ \lambda_j \neq 0 \}$ (i.e., by fixing the set of  zeros coefficients).
 
%While other pooled methods can be designed by combining rounding and polishing methods, we omit certain variations for the sake of clarity, namely: (i) methods that use rescaled rounding and unit weighting (as they perform worse than na\"ive rounding when we use pooling); (ii) methods that use \SR{} without \DCD{} polishing (as it worse than \PLRSRDCD{}). We report results for other pooled methods that make use of rounding and polishing of these variations on a different real-world dataset in Section \ref{Sec::SeizurePrediction} (i.e., \PLRLR{}, \PLRLRDCD{}, \PLRSR{}). 

% \subsubsection*{Notes on the Choice of Methods}  

% We also considered training risk scores by directly solving \RSMINLP{} with a commercial MINLP solver, but could not recover models that performed well on any dataset due to computational issues (discussed also in Section \ref{Sec::MethodComparison}). We did not consider methods to fit scoring systems for decision-making (see Section \ref{Sec::RelatedWork} for a list) since these models do not output risk predictions. \citet{ustun2016slim} and \citet{ustun2015recidivism} present experiments comparing the performance of decision-making ability of scoring systems and other popular classifiers on these datasets.

\paragraph{Performance Evaluation}

%\paragraph{Reliability Diagrams}

We evaluate the calibration of each risk score by plotting a \emph{reliability diagram}, which shows how the predicted risk (x-axis) matches the observed risk (y-axis) for each distinct score \citep[][]{degroot1983comparison}. The \emph{observed} risk at a score of $s$ is defined as $$\bar{p}_s = \frac{1}{|\{i: s_i = s\}|} \sum_{i: s_i = s}{\indic{y_i = +1}}.$$ If a model has over 30 distinct scores, we group them into 10 bins before plotting the reliability diagram. A model with perfect calibration should output predictions that are perfectly aligned with observed risk, as shown by a reliability diagram where all points lie on the $x = y$ line.

We report the following summary statistics for each model:
\begin{itemize}

\item \emph{Calibration Error}, computed as $\textrm{CAL} = \frac{1}{\n{}} \sum_{s}\sum_{i: s_i = s} |p_i - \bar{p}_s|$ where $p_i$ is the \emph{predicted risk} of example $i$, and $\bar{p}_s$ is the observed risk for all examples with a score of $s$. CAL is the expected calibration error over the reliability diagram~\citep[see, e.g.,  ][]{naeini2014binary}.

\item \emph{Area under the ROC curve}, computed as $\textrm{AUC} = \frac{1}{\nplus{}\nminus{}} \sum_{i:y_i = +1}\sum_{k: y_k = -1} \indic{s_i > s_k},$ where $\nplus{} = |\{i: y_i = +1\}|$, $\nminus{} = |\{i: y_i = -1\}|$. Note that trivial models (i.e., models that predict one class) achieve the best possible CAL (0.0\%) but poor AUC (0.5).

\item \emph{Logistic Loss}, computed as $\textrm{Loss} = \frac{1}{\n{}} \sum_{i=1}^\n{} \log(1+\exp(-y_is_i))$. The loss reflects the objective values of the risk score problem when $C_0$ is small. We report the loss to see if minimizing the objective value of the risk score problem improves CAL and AUC.

\item \emph{Model Size}: the number of non-zero coefficients excluding the intercept $\sum_{j=1}^d \indic{\lambda_j \neq 0}$.

\end{itemize}

\paragraph{Parameter Tuning}

We use \emph{nested} 5-fold cross-validation (5-CV) to choose the free parameters of a final risk score \citep[see][]{cawley2010over}. The final risk score is fit using the entire dataset for an instance of the free parameters that satisfies the model size constraint and maximizes the 5-CV mean test AUC.  %\footnote{It is well-known that when 5-CV statistics are used to set the free parameters for the final model, then these statistics produce an overly optimistic estimate of test performance \cite[see][]{cawley2010over}. To avoid this bias, the 5 fold-based models trained to construct 5-CV estimates must have their parameters set in the same way as the final model. Explicitly, for each the 5 fold-based models, the free parameters must be set by running an inner 5-CV, and choosing a free parameter instance that satisfies the model size constraint and maximizes the inner 5-CV mean test AUC.} 
%\footnote{Nested-CV is only needed for \PLRCR{}, \PLRCRDCD{}, \PLRSRDCD{} as these methods tune the parameters of a final risk score using 5-CV statistics. \RiskSLIM{} can produce an unbiased performance estimate using standard 5-CV because it does not require parameter tuning. This is also true for traditional methods such as \mPLRRd{}, \mPLRRsRd{}, \mPLRUnit{} as the parameter tuning step occurs before post-processing.}

\subsection{Discussion}
\label{Sec::ExperimentalResults}

\paragraph{On the Performance of Risk Scores}

We compare the performance of \RiskSLIM{} to traditional approaches in Figure \ref{Fig::RegPlotsTraditional}, and to pooled approaches in Figure~\ref{Fig::RegPlotsAdvanced}. These results show that \RiskSLIM{} models consistently attain better calibration and AUC than alternatives. We present these results in greater detail for risk scores with a target model size of $\zeronormmax = 5$ in Table~\ref{Table::ExperimentalResults}. Here, \RiskSLIM{} has the best 5-CV mean test CAL on 5/6 datasets, the best 5-CV mean test AUC on 5/6 datasets, and no method has better test CAL \emph{and} test AUC than \RiskSLIM{}. 

\newcommand{\regheader}[1]{\text{#1}}
\newcommand{\addregrow}[3]{\texttt{\scriptsize{#1}} & \includeregplot{#2}{MXE}{#3} & \includeregplot{#2}{CAL}{#3} & \includeregplot{#2}{AUC}{#3}}

\newcommand{\includereglegend}[1]{\cell{c}{\includegraphics[width=0.75\textwidth, trim=0mm 5mm 0mm 5mm, clip=true]{reg_plot_#1_legend.pdf}}}

\newcommand{\includeregplot}[3]{\cell{c}{\includegraphics[trim=16mm 6mm 0mm 0mm, clip=true, width=0.25\textwidth, keepaspectratio]{reg_plot_#1_#2_#3.pdf}}}

\begin{figure}[htbp]
\centering
\scriptsize
\includereglegend{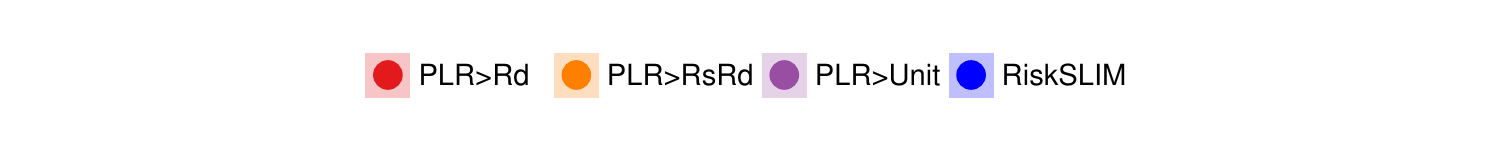}
\resizebox{0.82\textwidth}{!}{
\begin{tabular}{m{2cm}>{\hspace{0.4cm}}c>{\hspace{0.4cm}}c>{\hspace{0.4cm}}c}
\toprule & 
\textsc{Optimization Metric} & 
\textsc{Performance Metric} & 
\textsc{Selection Metric}\\
\cmidrule(lr){2-2} 
\cmidrule(lr){3-3} 
\cmidrule(lr){4-4}
\cell{l}{\textsc{Dataset}} & 
\regheader{Training Loss} & 
\regheader{5-CV Mean Test CAL} & 
\regheader{5-CV Mean Test AUC}\\[0.1em]
\toprule
\addregrow{income}{traditional}{adult}\\ 
\midrule
\addregrow{mammo}{traditional}{mammo} \\
\midrule
\addregrow{mushroom}{traditional}{mushroom} \\ 
\midrule
\addregrow{rearrest}{traditional}{recidivism_v01_rearrest} \\ 
\midrule
\addregrow{spambase}{traditional}{spambase} \\ 
\midrule
\addregrow{telemarketing}{traditional}{bank_large}\\ 
\bottomrule
\end{tabular}
}
\caption{Summary statistics for risk scores built using \RiskSLIM{} and traditional approaches. Each point represents the best risk score with integer coefficients $\lambda_j \in \{-5,\ldots,5\}$ and model size $\zeronorm{\lambdab} \leq \zeronormmax$ for $\zeronormmax \in\{2,\ldots,10\}$. We show the variation in 5-CV mean test CAL and AUC for each method by shading the range between the 5-CV minimum and maximum. The black line in each plot is a baseline, which shows the performance of a single \PLR{} model with real-valued coefficients and no model size constraint.}
\label{Fig::RegPlotsTraditional}
\end{figure}

\begin{figure}[htbp]
\centering
\scriptsize
\includereglegend{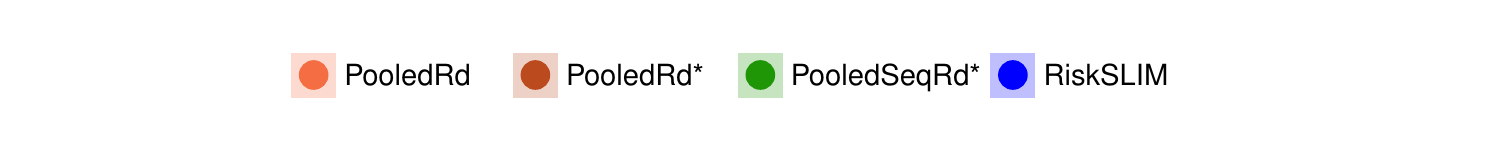}
\resizebox{0.82\textwidth}{!}{
\begin{tabular}{m{2cm}>{\hspace{0.4cm}}c>{\hspace{0.4cm}}c>{\hspace{0.4cm}}c}
\toprule &
\textsc{Optimization Metric} & 
\textsc{Performance Metric} & 
\textsc{Selection Metric}\\
\cmidrule(lr){2-2} 
\cmidrule(lr){3-3} 
\cmidrule(lr){4-4}
\cell{l}{\textsc{Dataset}} & 
\regheader{Training Loss} & 
\regheader{5-CV Mean Test CAL} & 
\regheader{5-CV Mean Test AUC}\\[0.1em]
\midrule
\addregrow{income}{advanced}{adult}\\ 
 \midrule
\addregrow{mammo}{advanced}{mammo} \\
\midrule
\addregrow{mushroom}{advanced}{mushroom} \\ 
\midrule
\addregrow{rearrest}{advanced}{recidivism_v01_rearrest} \\ 
\midrule
\addregrow{spambase}{advanced}{spambase} \\ 
\midrule
\addregrow{telemarketing}{advanced}{bank_large} \\ 
\bottomrule
\end{tabular}
}
\caption{Summary statistics for risk scores built using \RiskSLIM{} and pooled approaches. Each point represents the best risk score with integer coefficients $\lambda_j \in \{-5,\ldots,5\}$ and model size $\zeronorm{\lambdab} \leq \zeronormmax$ for $\zeronormmax \in\{2,\ldots,10\}$. We show the variation in 5-CV mean test CAL and AUC for each method by shading the range between the 5-CV minimum and maximum. The black line in each plot is a baseline, which shows the performance of a single \PLR{} model with real-valued coefficients and no model size constraint.}
\label{Fig::RegPlotsAdvanced}
\end{figure}

We make two observations to explain the empirical performance of risk scores.
\begin{romanlist}

\item Models that attain low values of the logistic loss have good calibration~\citep[see Figures \ref{Fig::RegPlotsTraditional} and \ref{Fig::RegPlotsAdvanced} and the empirical results of e.g.,][]{caruana2004data,caruana2006empirical}. 

\item Since we are fitting from a simple class of models, risk scores tend to generalize (see the test CAL/AUC and training CAL/AUC of \RiskSLIM{} models in Figures~\ref{Fig::model_arrest_RiskSLIM} to~\ref{Fig::model_spambase_RiskSLIM}, and other risk scores in Table~\ref{Table::GeneralizationResults} in Appendix~\ref{Appendix::Experiments}).

\end{romanlist} 
Since \RiskSLIM{} models optimize the loss over exact constraints on model form, they attain minimal or near-minimal values of the loss. Thus, they perform well in terms of training CAL/AUC as per (i) and test CAL/AUC as per (ii). These observations also explain why methods that use loss-minimizing heuristics produce risk scores with better CAL and AUC than those that do not (e.g., \PLRCRDCD{} has better test CAL/AUC than \PLRCR{} since \DCD{} polishing can only reduce the loss).

\begin{table}[t]
\centering
\renewcommand{\textds}[1]{{\texttt{\small{#1}}}}
\resizebox{\textwidth}{!} {
%%%%%
\begin{tabular}{ll*{5}{C{2.0cm}}C{2.7cm}C{1.8cm}}
\toprule
& 
& \multicolumn{3}{c}{\textsc{Traditional Approaches}}
& \multicolumn{3}{c}{\textsc{Pooled Approaches}}
& \\
%\cmidrule(lr){3-3} 
\cmidrule(lr){3-5} 
\cmidrule(lr){6-8}

\textbf{Dataset} & 
\textbf{Metric}
& \perfheader{\mPLRRd{}}
& \perfheader{\mPLRRsRd{}}
& \perfheader{\mPLRUnit{}}
& \perfheader{\PLRCR{}}
& \perfheader{\PLRCRDCD{}}
& \perfheader{\PLRSRDCD{}}
& \perfheader{\RiskSLIM{}}  \\ 

\toprule

\cell{l}{\textds{income}\\$n$~=~32561\\$d$~=~36}
 & \metrics{}
 & \cell{c}{10.5$\%$\\0.787\\0.465\\2\\-}
 & \cell{c}{19.5$\%$\\0.813\\0.777\\3\\-}
 & \cell{c}{25.4$\%$\\0.814\\0.599\\5\\-}
 & \cell{c}{3.0$\%$\\0.845\\0.392\\5\\-}
 & \cell{c}{3.1$\%$\\\cellcolor{best}0.854\\\cellcolor{best}0.383\\5\\-}
 & \cell{c}{4.2$\%$\\0.832\\0.417\\4\\-}
 & \cell{c}{\cellcolor{best}2.6$\%$\\\cellcolor{best}0.854\\0.385\\5\\9.7$\%$}

\\ \midrule

\cell{l}{\textds{mammo}\\$n$~=~961\\$d$~=~14}
 & \metrics{}
 & \cell{c}{10.5$\%$\\0.832\\0.526\\3\\-}
 & \cell{c}{16.2$\%$\\\cellcolor{best}0.846\\0.745\\5\\-}
 & \cell{c}{8.5$\%$\\0.842\\0.484\\5\\-}
 & \cell{c}{10.9$\%$\\0.845\\0.503\\3\\-}
 & \cell{c}{7.1$\%$\\0.841\\0.480\\3\\-}
 & \cell{c}{7.4$\%$\\0.845\\0.480\\3\\-}
 & \cell{c}{\cellcolor{best}5.0$\%$\\0.843\\\cellcolor{best}0.469\\5\\0.0$\%$}

 \\ \midrule

\cell{l}{\textds{mushroom}\\$n$~=~8124\\$d$~=~113}
 & \metrics{}
 & \cell{c}{22.1$\%$\\0.890\\0.543\\\cellcolor{issue}1\\-}
 & \cell{c}{8.0$\%$\\0.951\\0.293\\2\\-}
 & \cell{c}{19.9$\%$\\0.969\\0.314\\5\\-}
 & \cell{c}{12.6$\%$\\0.984\\0.211\\4\\-}
 & \cell{c}{4.6$\%$\\0.986\\0.130\\4\\-}
 & \cell{c}{5.4$\%$\\0.978\\0.144\\5\\-}
 & \cell{c}{\cellcolor{best}1.8$\%$\\\cellcolor{best}0.989\\\cellcolor{best}0.069\\5\\0.0$\%$}
 \\ 
   
\midrule

\cell{l}{\textds{rearrest}\\$n$~=~22530\\$d$~=~48}
 & \metrics{}
 & \cell{c}{7.3$\%$\\0.555\\0.643\\\cellcolor{issue}1\\-}
 & \cell{c}{24.2$\%$\\0.692\\1.437\\5\\-}
 & \cell{c}{21.8$\%$\\0.698\\0.703\\5\\-}
 & \cell{c}{5.2$\%$\\0.676\\0.618\\4\\-}
 & \cell{c}{\cellcolor{best}1.4$\%$\\0.676\\0.618\\4\\-}
 & \cell{c}{3.8$\%$\\0.677\\0.624\\4\\-}
 & \cell{c}{2.4$\%$\\\cellcolor{best}0.699\\\cellcolor{best}0.609\\5\\3.9$\%$}
 \\ 
   
\midrule

\cell{l}{\textds{spambase}\\$n$~=~4601\\$d$~=~57}
 & \metrics{}
 & \cell{c}{15.0$\%$\\0.620\\0.666\\\cellcolor{issue}1\\-}
 & \cell{c}{29.5$\%$\\0.875\\1.090\\4\\-}
 & \cell{c}{33.4$\%$\\0.861\\0.515\\5\\-}
 & \cell{c}{26.5$\%$\\0.910\\0.624\\5\\-}
 & \cell{c}{16.3$\%$\\0.913\\0.381\\5\\-}
 & \cell{c}{17.9$\%$\\0.908\\0.402\\5\\-}
 & \cell{c}{\cellcolor{best}11.7$\%$\\\cellcolor{best}0.928\\\cellcolor{best}0.349\\5\\27.8$\%$}

 \\ \midrule
 
 \cell{l}{\textds{telemarketing}\\$n$~=~41188\\$d$~=~57}
 & \bfcell{l}{test cal\\test auc\\loss value\\model size\\opt. gap}
 & \cell{c}{2.6$\%$\\0.574\\0.352\\\cellcolor{issue}0\\-}
 & \cell{c}{11.2$\%$\\0.700\\11.923\\3\\-}
 & \cell{c}{6.2$\%$\\0.715\\0.312\\3\\-}
 & \cell{c}{1.9$\%$\\0.759\\0.292\\4\\-}
 & \cell{c}{\cellcolor{best}1.3$\%$\\\cellcolor{best}0.760\\\cellcolor{best}0.289\\5\\-}
 & \cell{c}{\cellcolor{best}1.3$\%$\\\cellcolor{best}0.760\\\cellcolor{best}0.289\\5\\-}
 & \cell{c}{\cellcolor{best}1.3$\%$\\\cellcolor{best}0.760\\\cellcolor{best}0.289\\5\\3.5$\%$}

 \\ \bottomrule
\end{tabular}
}
\caption{Summary statistics for risk scores with integer coefficients $\lambda_j \in \{-5,\ldots,5\}$ for a model size constraint $\zeronorm{\lambdab} \leq 5$. Here: \emph{test cal} is the 5-CV mean test CAL; \emph{test auc} is the 5-CV mean test AUC; \emph{model size} and \emph{loss value} pertain to a final model trained using the entire dataset. For each dataset, we highlight the method that attains the best test cal, auc, and loss value in green. We also highlight methods that produce trivial models in red.}
\label{Table::ExperimentalResults}
\end{table}

\paragraph{On the Caveats of CAL}

The \mPLRRd{} risk score for \textds{telemarketing} in Table \ref{Table::ExperimentalResults} highlights a key shortcoming of CAL that illustrates why we report AUC: trivial and near-trivial models can have misleadingly low CAL. Here, \mPLRRd{} rounds all coefficients other than the intercept to zero, producing a model that trivially assigns a constant score to all examples $s_i = \dotprod{\lambdab}{\xb_i} = \lambda_0 = 2$. Since there is only one score, the predicted risk for all points is $p_i = 11.9\%$,and the observed risk is the proportion of positive examples $\bar{p} = \prob{y_i = +1} = 11.3\%$. Thus, a trivial model has a training CAL of 0.7\%, the lowest among all methods, which (misleadingly) suggests that it has the best performance on training data (see Table \ref{Table::GeneralizationResults} in Appendix~\ref{Appendix::Experiments} for values of training CAL). In this case, one could instead determine that the model is trivial by its training AUC, which is 0.500. This result also shows why we choose free parameters that maximize the 5-CV mean test AUC rather than CAL: choosing free parameters to minimize the 5-CV mean test CAL can result in a trivial model. 

\setlength{\calplotwidth}{0.15\textwidth}
\setlength{\leftmostplotwidth}{0.187\textwidth}

\newcommand{\addahrscalrow}[2]{%
\texttt{\texttt{#1}}&%
\addleftcalplot{#2}{ahrs_enet_rnd_logit}&%
\addcalplot{#2}{ahrs_enet_srnd_logit}&%
\addcalplot{#2}{ahrs_enet_unit_logit}&%
\addcalplot{#2}{RSLP_CR}&\addcalplot{#2}{RSLP_CRDCD}&%
\addcalplot{#2}{RSLP_SRDCD}&%
\addcalplot{#2}{risk_slim}%
}

\begin{figure}[ht]
\scriptsize\centering
\resizebox{\textwidth}{!} {
\setlength{\tabcolsep}{2.5pt}
\begin{tabular}{m{1.25cm}*{3}{c@{ }}|*{3}{c@{ }}|c}
\toprule
& \multicolumn{3}{c}{\textsc{Traditional Approaches}} & \multicolumn{3}{c}{\textsc{Pooled Approaches}} & \\  
\cmidrule(lr){2-4} \cmidrule(lr){5-7}
& \perfheader{\mPLRRd{}} & \perfheader{\mPLRRsRd{}} & \perfheader{\mPLRUnit{}}
& \perfheader{\PLRCR{}} & \perfheader{\PLRCRDCD{}} & \perfheader{\PLRSRDCD{}}
& \RiskSLIM{}  \\  \toprule
\addahrscalrow{income}{adult} \\  \midrule
\addahrscalrow{mammo}{mammo} \\  \midrule
\addahrscalrow{mushroom}{mushroom} \\  \midrule
\addahrscalrow{rearrest}{recidivism_v01_rearrest} \\  \midrule
\addahrscalrow{spambase}{spambase} \\ \midrule
\addahrscalrow{telemark}{bank_large} \\  \bottomrule
\end{tabular}
}
\caption{Reliability diagrams for risk scores with integer coefficients $\lambda_j \in \{-5,\ldots,5\}$ for a model size constraint $\zeronorm{\lambdab} \leq 5$. We plot results for models from each fold on test data in grey, and for the final model on training data in black.}
\label{Fig::CalibrationPlots}
\end{figure}

\paragraph{On Calibration Issues of Risk Scores}

The reliability diagrams in Figure \ref{Fig::CalibrationPlots} highlight two issues with respect to the calibration of risk scores that are difficult to capture using a summary statistic:
\begin{itemize}

\item Monotonicity violations in observed risk. For example, the reliability diagrams for \mPLRRd{} on \textds{spambase}, or \PLRCR{} and \PLRSRDCD{} on \textds{mushroom} show that the observed risk does not increase monotonically with predicted risk. There is not a way to calibrate the risk scores to remove this, and it is non-intuitive to have an increase in risk score correspond to a decrease in actual risk.

\item Irregular spacing and coverage of predicted risk. For example, the \mPLRRd{} risk score for \textds{income} outputs risk predictions that range between only 20\% to 60\%, and the \mPLRRsRd{} risk score for \textds{rearrest} produces risk predictions that are clustered at end points.

\end{itemize}

The results in Figure \ref{Fig::CalibrationPlots} suggest that such issues can be mitigated by optimizing the logistic loss (see, e.g., the calibration of risk scores built using \RiskSLIM{}, \PLRCRDCD{}, \PLRSRDCD{} where integer coefficients are determined by directly optimizing the logistic loss). In contrast, these issues are difficult to address by post-processing. Consider, for example, using Platt scaling \citep{platt1999probabilistic} to improve the calibration of the \mPLRRd{} and \mPLRRsRd{} risk scores in Figure \ref{Fig::CalibrationPlots}. As shown in Figure \ref{Fig::PlattScalingIssues}, Platt scaling improves calibration by centering and spreading risk estimates over the reliability diagram. However, it does not resolve issues that were introduced by earlier heuristics, such as monotonicity violations, a lack of coverage in risk predictions, or low AUC. 

\renewcommand{\addahrscalrow}[2]{{\texttt{\tiny#1}}&%
\addleftcalplot{#2}{ahrs_enet_rnd_logit}&%
\addcalplot{#2}{ahrs_enet_rnd_platt}&%
\addcalplot{#2}{ahrs_enet_srnd_logit}&%
\addcalplot{#2}{ahrs_enet_srnd_platt}&%
\addcalplot{#2}{risk_slim}%
}

\begin{figure}[htbp]
\centering
\resizebox{\textwidth}{!} {
\tiny
\setlength{\tabcolsep}{2.5pt}
\begin{tabular}{m{1.25cm}*{2}{c@{ }}|*{2}{c@{ }}|c}
& \multicolumn{2}{c}{\textsc{Rounding}} & 
\multicolumn{2}{c}{\textsc{Rescaled Rounding}} & \\ 
 \cmidrule(lr){2-3} \cmidrule(lr){4-5}

& \textsc{Raw}
& \textsc{+ Platt Scaling}
& \textsc{Raw}
& \textsc{+ Platt Scaling}
& \RiskSLIM{}

\\[-0.1em] \toprule
\addahrscalrow{income}{adult} \\ 
\midrule
\addahrscalrow{rearrest}{recidivism_v01_rearrest}
%\midrule
%\addahrscalrow{bank}{bank_large} \\ 
%\midrule
%\addahrscalrow{mammo}{mammo} \\ 
%\midrule
%\addahrscalrow{mushroom}{mushroom} \\ 
%\midrule
%\addahrscalrow{spambase}{spambase} \\
\end{tabular}
}
\caption{Reliability diagrams for \mPLRRd{} and \mPLRRsRd{} risk scores with and without Platt scaling, and for \RiskSLIM{}. Platt scaling improves calibration by centering and spreading risk predictions. However, it cannot overcome calibration issues introduced by rounding heuristics such a lack of decision points (e.g. \mPLRRd{} on \textds{income} and \textds{rearrest}, left) or monotonicity violations (e.g, \mPLRRsRd{} on \textds{income}, top middle).}
\label{Fig::PlattScalingIssues}
\end{figure}

\paragraph{On the Pitfalls of Traditional Approaches}

Our results in Figure \ref{Fig::RegPlotsTraditional} and Table \ref{Table::ExperimentalResults} show that risk scores built using traditional approaches perform poorly in terms of CAL and AUC. In particular:
\begin{itemize}

\item Rounding (\mPLRRd{}) produces risk scores with low AUC when it eliminates features from the model by rounding small coefficients $\lambda_j < |0.5|$ to zero. 

\item Rescaled rounding (\mPLRRsRd{}) hurts calibration since the logistic loss is not scale-invariant (see e.g., the reliability diagram for \textds{rearrest} \mPLRRsRd{} in Figure \ref{Fig::CalibrationPlots}). 

\item  Unit weighting (\mPLRUnit{}) results in poor calibration and unpredictable behavior (e.g., risk scores with more features can perform worse as seen in \mPLRUnit{} models for \textds{income} and \textds{telemarketing} in Figure \ref{Fig::RegPlotsTraditional}). 
\end{itemize}
The effects of rescaled rounding and unit weighting on calibration are reflected by highly suboptimal values of the loss in Table \ref{Table::ExperimentalResults} and Figure \ref{Fig::RegPlotsTraditional}. These issues are often overlooked, perhaps because their effect on AUC is far less severe~\citep[e.g. the rescaling is recommended by][]{usdoj2005riskclassification,psc2012report4}.

Our baseline methods may not match the exact methods used in practice as they do not reflect the significant human input used in risk score development (e.g., domain experts perform preliminary feature selection, round coefficients, or choose a scaling factor before rounding, manually or without validation, as shown in Appendix~\ref{Appendix::RiskScoreBackground}). Nevertheless, these results highlight two major pitfalls of traditional approaches, namely:

\begin{itemize}

\item Traditional approaches heuristically post-process a single model. This means that they fail whenever a heuristic dramatically changes CAL or AUC.

\item Traditional approaches use heuristics that are oblivious to the value of the loss function, which tends to result in poor calibration.

\end{itemize}

\paragraph{On Pooled Approaches}

Our results suggest that risk scores built using our pooled approaches attain considerably better calibration and rank accuracy than those built using traditional approaches. These methods aim to overcome the pitfalls of traditional approaches using two strategies:
\begin{itemize}

\item \emph{Pooling}, which generates a pool of \PLR{} models, post-processes each model to produce a pool of risk scores, and selects the best risk score within the pool. Pooling provides some robustness against failure modes of heuristics that dramatically alter performance (e.g., for rounding, it is very unlikely that the coefficients of all models in the pool will be rounded to zero). The performance gain due to pooling can be seen by comparing the results for \mPLRRd{} to \PLRCR{} in Table \ref{Table::ExperimentalResults}.

\item \emph{Loss-Sensitive Heuristics}, such as \SR{} and \DCD{}, which produce a pool of risk scores that attain lower values of the loss, and thereby let us select a risk score with better CAL and AUC. The performance gain due to loss-sensitive heuristics can be seen by comparing the results for \PLRCR{} to \PLRCRDCD{} in Table \ref{Table::ExperimentalResults}. 

\end{itemize}
The fact that \RiskSLIM{} risk scores have lower loss compared to pooled methods shows that direct optimization can efficiently find solutions that may not be found by exhaustive post-processing (e.g., where we fit all possible $\ell_1$ and $\ell_2$ penalized logistic regression models, and convert them to risk scores with specially-designed heuristics). Here, we have shown that exhaustive post-processing strategies can often produce risk scores that perform well. In Section \ref{Sec::SeizurePrediction}, however, we will see that the performance gap can be significant in the presence of non-trivial constraints.

\paragraph{On Computation}

Although the risk score problem is NP-hard, we trained \RiskSLIM{} models that were certifiably optimal or had small optimality gaps for all datasets in under 20 minutes using an \LCPA{} implementation with the improvements in Section \ref{Sec::AlgorithmicImprovements}. Even when \LCPA{} did not recover a certifiably optimal solution, it produced a risk score that performed well and did not exhibit the calibration issues of models built heuristically. 

In general, the time spent computing cutting planes is a small portion of the overall runtime for \LCPA{} ($<1\%$, for all datasets). Given that \LCPA{} scales linearly with sample size, we expect to obtain similar results even if the datasets had far more samples. Factors that affected the time to obtain a certifiably optimal solution include:
\begin{itemize}

\item \emph{Highly Correlated Features}: Subsets of redundant features produce multiple optima, which increases the size of the B\&B tree.

\item \emph{Feature Encoding}: In particular, the problem is harder when the dataset includes real-valued variables, like those in the \textds{spambase} dataset.

\item \emph{Difficulty of the Learning Problem}: On separable problems such as \textds{mushroom}, it is easy to recover a certifiably optimal solution since many solutions perform well and produce a near-optimal lower bound.

\end{itemize}

\begin{figure}[htbp]
\centering{
\scoringsystem{}
\begin{tabular}{|l l  c | l |}
   \hline
1. & \textssm{Prior\;Arrests $\geq$ 2} & 1 point & $\phantom{+}\prow{}$ \\ 
  2. & \textssm{Prior\;Arrests $\geq$ 5} & 1 point & $+\prow{}$ \\ 
  3. & \textssm{Prior\;Arrests\;for\;Local\;Ordinance} & 1 point & $+\prow{}$ \\ 
  4. & \textssm{Age\;at\;Release\;between\;18\;to\;24} & 1 point & $+\prow{}$ \\ 
  5. & \textssm{Age\;at\;Release\;$\geq$ 40} & -1 point & $+$ \\ 
   \hline
 & \instruction{1}{5} & \scorelabel{} & $=$ \\ 
   \hline
\end{tabular}

\vspace{0.1em}\risktable{}
\begin{tabular}{|l|c|c|c|c|c|c|}
   \hline
\rowcolor{scorecolor}\scorelabel{} & -1 & 0 & 1 & 2 & 3 & 4 \\ 
   \hline
\rowcolor{riskcolor}\risklabel{} & 11.9$\%$ & 26.9$\%$ & 50.0$\%$ & 73.1$\%$ & 88.1$\%$ & 95.3$\%$ \\ 
   \hline
\end{tabular}
}
\caption{\RiskSLIM{} model for \textds{rearrest}. RISK is the predicted probability that a prisoner is arrested within 3 years of release from prison. This model has a 5-CV mean test CAL/AUC of 1.7\%/0.697 and training CAL/AUC of 2.6\%/0.701.}
\label{Fig::model_arrest_RiskSLIM}
\end{figure}

\begin{figure}[htbp]
\centering{
\scoringsystem{}
\begin{tabular}{|l l  c | l |}
   \hline
1. & \textssm{Married} & 3 points & $\phantom{+}\prow{}$ \\ 
  2. & \textssm{Reported\;Capital\;Gains} & 2 points & $+\prow{}$ \\ 
  3. & \textssm{Age\,22\,to\,29} & -1 point & $+\prow{}$ \\ 
  4. & \textssm{Highest\;Level\;of\;Education\;is\;High\;School\;Diploma} & -2 points & $+\prow{}$ \\ 
  5. & \textssm{No\;High\;School\;Diploma} & -3 points & $+$ \\ 
   \hline
 & \instruction{1}{5} & \scorelabel{} & $=$ \\ 
   \hline
\end{tabular}
\vspace{0.25em}\risktable{}
\begin{tabular}{|l|c|c|c|c|c|c|c|}
   \hline
\rowcolor{scorecolor}\scorelabel{} & -4 to -1 & 0 & 1 & 2 & 3 & 4 & 5 \\ 
   \hline
\rowcolor{riskcolor}\risklabel{} & $<$ 5.0$\%$ & 11.9$\%$ & 26.9$\%$ & 50.0$\%$ & 73.1$\%$ & 88.1$\%$ & 95.3$\%$ \\ 
   \hline
\end{tabular}
}
\caption{\RiskSLIM{} model for \textds{income}. RISK is the predicted probability that a US resident earns over \$50\,000. This model has a 5-CV mean test CAL/AUC of 2.4\%/0.854 and training CAL/AUC of 4.1\%/0.860.}
\label{Fig::model_adult_RiskSLIM}
\end{figure}

\begin{figure}[htbp]
\centering{
\scoringsystem{}
\begin{tabular}{|l l  c | l |}
   \hline
1. & \textssm{Call\;between\;January\;and\;March} & 1 point & $\phantom{+}\prow{}$ \\ 
  2. & \textssm{Called\;Previously} & 1 point & $+\prow{}$ \\ 
  3. & \textssm{Previous\;Call\;was\;Successful} & 1 point & $+\prow{}$ \\ 
  4. & \textssm{Employment\;Indicator $<$ 5100} & 1 point & $+\prow{}$ \\ 
  5. & \textssm{3\;Month\;Euribor\;Rate\;$\geq$ 100} & -1 point & $+$ \\ 
   \hline
 & \instruction{1}{5} & \scorelabel{} & $=$ \\ 
   \hline
\end{tabular}
\vspace{0.1em}\risktable{}
\begin{tabular}{|l|c|c|c|c|c|c|}
   \hline
\rowcolor{scorecolor}\scorelabel{} & -1 & 0 & 1 & 2 & 3 & 4 \\ 
   \hline
\rowcolor{riskcolor}\risklabel{} & 4.7$\%$ & 11.9$\%$ & 26.9$\%$ & 50.0$\%$ & 73.1$\%$ & 88.1$\%$ \\ 
   \hline
\end{tabular}
}
\caption{\RiskSLIM{} model for \textds{telemarketing}. RISK is the predicted probability that a client opens a new bank account after a marketing call. This model has a 5-CV mean test CAL/AUC of 1.3\%/0.760 and a training CAL/AUC of 1.1\%/0.760.}
\label{Fig::model_bank_RiskSLIM}
\end{figure}

\begin{figure}[htbp]
\centering{
\scoringsystem{}
\begin{tabular}{|l l  c | l |}
   \hline
1. & \textssm{odor = foul} & 5 points & $\phantom{+}\prow{}$ \\ 
  2. & \textssm{gill\,size = broad} & -3 points & $+\prow{}$ \\ 
  3. & \textssm{odor = almond} & -5 points & $+\prow{}$ \\ 
  4. & \textssm{odor = anise} & -5 points & $+\prow{}$ \\ 
  5. & \textssm{odor = none} & -5 points & $+$ \\ 
   \hline
 & \instruction{1}{5} & \scorelabel{} & $=$ \\ 
   \hline
\end{tabular}

\vspace{0.1em}\risktable{}
\begin{tabular}{|l|c|c|c|c|}
   \hline
\rowcolor{scorecolor}\scorelabel{} & -8 & -5 & -3 & 2 to 5\\ 
   \hline
\rowcolor{riskcolor}\risklabel{} & 1.8$\%$ & 26.9$\%$ & 73.1$\%$ & $>$ 95.0$\%$ \\ 
   \hline
\end{tabular}
}
\caption{\RiskSLIM{} model for \textds{mushroom}. RISK is the predicted probability that a mushroom is poisonous. This model has a 5-CV mean test CAL/AUC of 1.8\%/0.989 and a training CAL/AUC of 1.0\%/0.990.}
\label{Fig::model_mushroom_RiskSLIM}
\end{figure}

\begin{figure}[htbp]
\centering{
\scoringsystem{}
\begin{tabular}{|l l  c | l |}
   \hline
1. & \textssm{IrregularShape} & 1 point & $\phantom{+}\prow{}$ \\ 
  2. & \textssm{Age $\geq$ 60} & 1 point & $+\prow{}$ \\ 
  3. & \textssm{OvalShape} & -1 point & $+\prow{}$ \\ 
  4. & \textssm{ObscuredMargin} & -1 point & $+\prow{}$ \\ 
  5. & \textssm{CircumscribedMargin} & -2 points & $+$ \\ 
   \hline
 & \instruction{1}{5} & \scorelabel{} & $=$ \\ 
   \hline
\end{tabular}
\vspace{0.1em}\risktable{}
\begin{tabular}{|l|*{6}{c|}}
   \hline
\rowcolor{scorecolor}\scorelabel{} & -3 & -2 & -1 & 0 & 1 & 2 \\ 
   \hline
\rowcolor{riskcolor}\risklabel{} & 4.7$\%$ & 11.9$\%$ & 26.9$\%$ & 50.0$\%$ & 73.1$\%$ & 88.1$\%$ \\ 
   \hline
\end{tabular}
}
\caption{\RiskSLIM{} model for \textds{mammo}. RISK is the predicted probability that a mammogram pertains to a patient with breast cancer. This model has a 5-CV mean test CAL/AUC of 5.0\%/0.843 and a training CAL/AUC of 3.1\%/0.849.}
\label{Fig::model_mammo_RiskSLIM}
\end{figure}
\begin{figure}[htbp]
{
\scoringsystem{}
\begin{tabular}{|l l  c | l |}
   \hline
1. & \textssm{CharacterFrequency\_DollarSign} & $\times$ 5 points & $\phantom{+}\prow{}$ \\ 
  2. & \textssm{WordFrequency\_Remove} &  $\times$  4 points & $+\prow{}$ \\ 
  3. & \textssm{WordFrequency\_Free} & $\times$  2 points & $+\prow{}$ \\ 
  4. & \textssm{WordFrequency\_HP} &  $\times$ -2 points & $+\prow{}$ \\ 
  5. & \textssm{WordFrequency\_George} &  $\times$  -5 points & $+$ \\ 
   \hline
 & \instruction{1}{5} & \scorelabel{} & $=$ \\ 
   \hline
\end{tabular}

\vspace{0.2em}
\resizebox{0.9\textwidth}{!}{
\risktable{}
\begin{tabular}{|l|*{10}{c|}}
   \hline
\rowcolor{scorecolor}
\scorelabel{} & $\leq$ -1.2 & -1.2 to -0.4 & -0.4 to 0.2 & 0.2 to 0.6 & 0.6 to 1.0 & 1.0 to 1.4 & 1.4 to 1.8 & 1.9 to 2.4 & 2.4 to 3.2 & $\geq$ 3.2\\ 
   \hline
\rowcolor{riskcolor}\risklabel{} & 1.4$\%$ & 14.8$\%$ & 26.8$\%$ & 35.3$\%$ & 45.1$\%$ & 54.1$\%$ & 65.3$\%$ & 74.7$\%$ & 85.6$\%$ & 97.2$\%$ \\  
   \hline
\end{tabular}
}
}
\caption{\RiskSLIM{} model for \textds{spambase}. RISK is the predicted probability that an e-mail is spam. This model has a 5-CV mean test CAL/AUC of 11.7\%/0.928 and a training CAL/AUC of 12.3\%/0.935.}
\label{Fig::model_spambase_RiskSLIM}
\end{figure}

%CAL (Mean Calibration Error @ Distinct Scores) & \cell{c}{12.3$\%$} & \cell{c}{\scriptsize{11.7$\%$}\\\tiny{8.6 - 15.3$\%$}} \\ 
%   \midrule MCAL (Maximum Calibration Error @ Distinct Scores) & \cell{c}{100.0$\%$} & \cell{c}{\scriptsize{99.3$\%$}\\\tiny{96.4 - 100.0$\%$}} \\ 
%   \midrule Model Size & \cell{c}{5} & \cell{c}{\scriptsize{5}\\\tiny{5 - 5}} \\ 
%   \midrule AUC & \cell{c}{0.935} & \cell{c}{\scriptsize{0.928}\\\tiny{0.918 - 0.940}} \\ 
\clearpage

%% file: section_06_seizure_prediction.tex
In this section, we describe a collaboration with the Massachusetts General Hospital and the University of Wisconsin Hospital where we built a customized risk score for ICU seizure prediction \citep{struck2017association}. Our goal is to discuss practical aspects of our approach on a real-world problem with non-trivial constraints.

\subsection{Problem Description}

\begin{figure}[htbp]
\centering
\begin{tabular}{@{}l>{\quad}r@{}}
\includegraphics[width=0.310\textwidth,trim=0mm 0mm 0mm 0mm,clip=true,page=1]{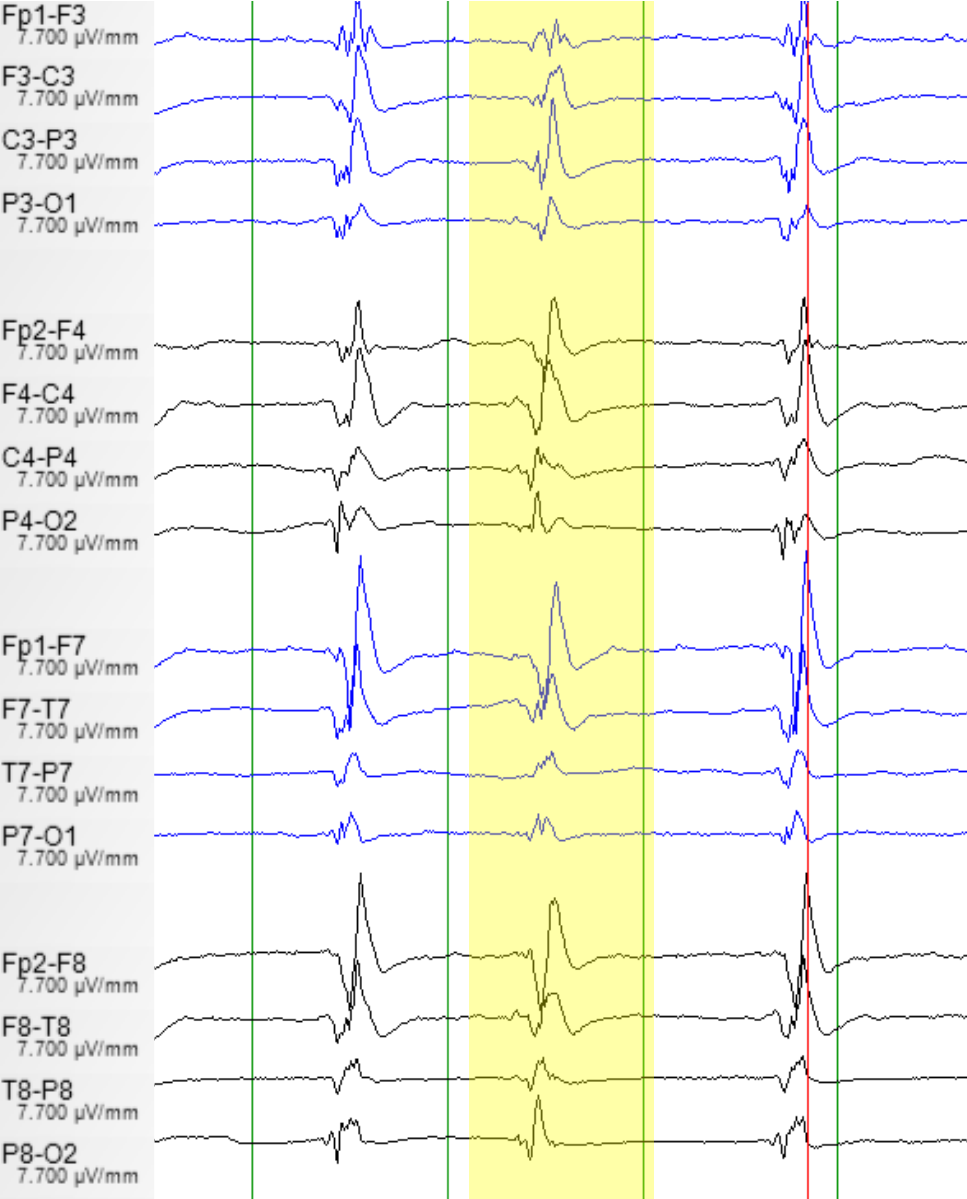}  &
\includegraphics[width=0.310\textwidth,trim=0mm 0mm 0mm 0mm,clip=true,page=2]{seizure_diagrams.pdf} 
\end{tabular}
\caption{cEEG displays electrical activity at 16 standardized locations in a patient's brain using electrodes placed on the scalp. We show two cEEG patterns: a Generalized Periodic Discharge (GPD), which occurs on both sides of the brain (left); and a Lateralized Periodic Discharge (LPD), which occurs on one side of the brain (right). These figures were reproduced from a presentation at a training module by the \citet{acns2012training}.}
\label{Fig::cEEGPatterns}
\end{figure}

Patients who suffer from traumatic brain injury (e.g., due to a ruptured brain aneurysm) often experience seizures while they are in intensive care. Since seizures are not outwardly visible in this setting, but may lead to irreversible brain damage, patients who are brought into the ICU are monitored via \emph{continuous electroencephalography} (cEEG). Based on current clinical standards, neurologists undergo extensive training to recognize a large set of patterns in cEEG output \citep[see e.g., Figure \ref{Fig::cEEGPatterns} and][]{hirsch2013american}. They consider the presence and characteristics of cEEG patterns along with other medical information to evaluate a patient's risk of seizure. These risk estimates are used to decide if a patient can be dismissed from the ICU, kept for further monitoring, or prescribed a medical intervention to avert potential brain injury. In practice, hospitals have a limited number of cEEG monitors that may be poorly assigned between patients in the ICU. Given reliable estimates of seizure risk, patients with a low risk of seizure can be taken off monitoring to free up monitors for new patients.

\paragraph{Data}

Our dataset was derived from cEEG recordings from 41 hospitals, curated by the Critical Care EEG Monitoring Research Consortium. It contains $\n = 5,427$ recordings and $d = 87$ input variables (see Appendix \ref{Appendix::SeizurePrediction} for a list). The outcome is defined as $y_i = +1$ if patient $i$ who has been in the ICU for the past 24 hours will have a seizure in the next 24 hours. There is significant class imbalance as $\prob{y_i = +1}$ = 12.5\%. The input variables include information on patient medical history, secondary neurological symptoms, and the presence and characteristics of 5 standard cEEG patterns: \emph{Lateralized Periodic Discharges} (LPD); \emph{Lateralized Rhythmic Delta} (LRDA); \emph{Generalized Periodic Discharges} (GPD); \emph{Generalized Rhythmic Delta} (GRDA); and  \emph{Bilateral Periodic Discharges} (BiPD).

\paragraph{Model Requirements}

Our collaborators wanted a risk score to help them reliably predict seizure risk by checking the presence and characteristics of cEEG patterns. It was critical for the model to output calibrated risk predictions since physicians would use the predicted risk to choose between multiple treatment options~\citep[i.e., patients may be prescribed different medication based on their predicted risk; see also][for a discussion on how miscalibrated risk predictions can lead to harmful decisions]{van2015calibration,shah2018big}. To be adopted by physicians, it was also important to build a model that could be validated by domain experts and that was aligned with domain expertise.

In Figure~\ref{Fig::2HELPS2B}, we present a \RiskSLIM{} risk score for this problem that satisfies all of these requirements \citep[see][for details on model development]{struck2017association}. This risk score outputs calibrated risk estimates at several operating points. It obeys a model size constraint ($\zeronorm{\lambdab} \leq 6$) to let physicians easily validate the model, and monotonicity constraints to ensure that the signs of some coefficients are aligned with domain knowledge.

\begin{figure}[htbp]
\centering{
\scoringsystem{}
\begin{tabular}{|l l  r | l |}
   \hline
1. & \textssm{Any cEEG Pattern with Frequency \textbf{$>\bm{2}$ Hz}} & 1 point & $\phantom{+}\prow{}$ \\ 
  2. & \textssm{\textbf{E}pileptiform Discharges} & 1 point & $+\prow{}$ \\ 
  3. & \textssm{Patterns include \textbf{L}PD or LRDA or BIPD} & 1 point & $+\prow{}$ \\ 
  4. & \textssm{\textbf{P}atterns Superimposed with Fast, or Sharp Activity} & 1 point & $+\prow{}$ \\ 
  5. &  \textssm{Prior \textbf{S}eizures} & 1 point & $+\prow{}$ \\ 
  6. &  \textssm{\textbf{B}rief Rhythmic Discharges} & 2 points & $+$ \\ 
   \hline
 & \instruction{1}{4} & \scorelabel{} & $=$ \\ 
   \hline
\end{tabular}
\risktable{}
\begin{tabular}{|l|c|c|c|c|c|c|c|c|c|c|c|}
\hline
\rowcolor{scorecolor}\scorelabel{} & 0 & 1 & 2 & 3 & 4 & 5 & 6+\\
\hline
\rowcolor{riskcolor}\risklabel{} &$<$5\% & 12\% & 27\% & 50\% & 73\% & 88 \% & $>$95\% \\
\hline
\end{tabular}
}
\caption{2HELPS2B risk score built by \RiskSLIM{} \citep[see][for details]{struck2017association}. This model has a 5-CV mean test CAL/AUC of 2.7\%/0.819.}
\label{Fig::2HELPS2B}
\end{figure}

As a follow up to our work in \citet[][]{struck2017association}, our collaborators wanted to see if we could improve the usability of the risk score in Figure \ref{Fig::2HELPS2B} without sacrificing too much calibration or rank accuracy. Seeing how the risk score in Figure \ref{Fig::2HELPS2B} includes features that could require a physician to check a large number of patterns (e.g., \textfn{MaxFrequencyOfAnyPattern$\geq$2 Hz} or \textfn{PatternsIncludeLPD\,or\,LRDA\,or\,BIPD}), our collaborators sought to improve usability by specifying \emph{operational constraints} on feature composition and feature encoding. In turn, our goal was to produce the best risk score that satisfied these constraints, so that our collaborators could correctly evaluate the loss in predictive performance due to their requirements, and make an informed choice between competing models. We present a complete list of their constraints in Appendix \ref{Appendix::SeizurePrediction}. They can be grouped as follows:

\begin{itemize}[leftmargin=*]

\item \emph{Limited Model Size}: The model had to use at most 4 input variables, so that it would be easy to validate and use in an ICU.

\item \emph{Monotonicity}: The model had to obey monotonicity constraints for well-known risk factors for seizures (e.g., it could not suggest that having prior seizures lowers seizure risk).

\item \emph{No Redundancy Between Categorical Variables}: The model had to include variables that were linearly independent (e.g., it could include \textfn{Male} or \textfn{Female} but not both).

\item \emph{Specific cEEG Patterns or Any cEEG Pattern}: The dataset had variables for specific cEEG patterns (e.g., \textfn{MaxFrequencyLPD}) and variables for any cEEG pattern (e.g., \textfn{MaxFrequencyAnyPattern}). The model had to use variables for specific patterns or any pattern, but not both.

\item \emph{Frequency in Continuous or Thresholded Form}: The dataset had two kinds of variables related to the frequency of a cEEG pattern: (i) a real-valued variable (e.g., \textfn{MaxFrequencyLPD} $\in [0, 3.0]$); and (ii) 7 binary threshold variables (e.g., \textfn{MaxFrequencyLPD\,$\leq$\,0.5\,Hz}). Models had to use the real-valued variable or the binary variables, not both. 

\item \emph{Limited \# of Thresholds for Thresholded Encoding}: To prevent clinicians from having to check multiple thresholds, the model could include at most 2 binary threshold variables for a given cEEG pattern.

\end{itemize}

\paragraph{Training Setup}

We used the training setup in Section \ref{Sec::ExperimentalSetup}, which we adapted to address constraints as follows. We trained a \RiskSLIM{} model by solving a customized instance of \rsminlp{} with 20 additional constraints and 2 additional variables, which we solved to optimality in $\leq 20$ minutes. The baseline methods had built-in mechanisms to handle monotonicity constraints, but required tuning to handle other constraints. For each method, we trained a final model using all of the training data for the instance of the free parameters that obeyed all constraints and maximized the mean 5-CV test AUC. %We evaluated the predictive performance of all models via nested 5-CV. 

\subsection{Discussion}

\begin{table}[ht]
\tableformat{}
\resizebox{\textwidth}{!}{
\begin{tabular}{l*{8}{c}}
 \toprule

\textbf{Method}
& \bfcell{c}{Constraints\\Violated}
 & \bfcell{c}{Test\\CAL}
 & \bfcell{c}{Test\\AUC}
 & \bfcell{c}{Model\\Size}
 & \bfcell{c}{Loss\\Value}
 & \bfcell{c}{Optimality\\Gap}
 & \bfcell{c}{Train\\CAL}
 & \bfcell{c}{Train\\AUC}
 \\ 
  \toprule
\RiskSLIM{}
& \cell{c}{--}
 & \cell{c}{\scriptsize{2.5$\%$}\\\tiny{1.9 - 3.4$\%$}}
 & \cell{c}{\scriptsize{0.801}\\\tiny{0.758 - 0.841}}
 & \cell{c}{\scriptsize{4}\\\tiny{4 - 4}}
 & \cell{c}{0.293}
 & \cell{c}{0.0$\%$}
 & \cell{c}{2.0$\%$}
 & \cell{c}{0.806}
 \\ 
   
\midrule

\PLRCR{}
& \cell{c}{--}
 & \cell{c}{\scriptsize{5.3$\%$}\\\tiny{3.1 - 7.1$\%$}}
 & \cell{c}{\scriptsize{0.740}\\\tiny{0.712 - 0.757}}
 & \cell{c}{\scriptsize{2}\\\tiny{1 - 3}}
 & \cell{c}{0.350}
 & \cell{c}{--}
 & \cell{c}{6.0$\%$}
 & \cell{c}{0.752}
 \\

\midrule

\PLRCRDCD{}
& \cell{c}{--}
 & \cell{c}{\scriptsize{3.0$\%$}\\\tiny{1.4 - 3.6$\%$}}
 & \cell{c}{\scriptsize{0.745}\\\tiny{0.712 - 0.776}}
 & \cell{c}{\scriptsize{2}\\\tiny{1 - 3}}
 & \cell{c}{0.308}
 & \cell{c}{--}
 & \cell{c}{1.9$\%$}
 & \cell{c}{0.754}
 \\

\midrule

\PLRSRDCD{}
& \cell{c}{--}
 & \cell{c}{\scriptsize{2.8$\%$}\\\tiny{2.4 - 3.1$\%$}}
 & \cell{c}{\scriptsize{0.745}\\\tiny{0.713 - 0.805}}
 & \cell{c}{\scriptsize{3}\\\tiny{2 - 4}}
 & \cell{c}{0.313}
 & \cell{c}{--}
 & \cell{c}{1.9$\%$}
 & \cell{c}{0.767}
 \\

 \midrule

\PLR{}
 & \cell{c}{All}
 & \cell{c}{\scriptsize{2.6$\%$}\\\tiny{1.7 - 3.6$\%$}}
 & \cell{c}{\scriptsize{0.844}\\\tiny{0.829 - 0.869}}
 & \cell{c}{\scriptsize{29}\\\tiny{20 - 35}}
% & \cell{c}{-}
 & \cell{c}{0.272}
 & \cell{c}{--}
 & \cell{c}{2.0$\%$}
 & \cell{c}{0.850}
 \\ 
   
\midrule

% \PLR{}
%  & \cell{c}{Model Size\\Integrality\\Operational}
%  & \cell{c}{\scriptsize{2.7$\%$}\\\tiny{1.6 - 3.1$\%$}}
%  & \cell{c}{\scriptsize{0.845}\\\tiny{0.829 - 0.870}}
%  & \cell{c}{\scriptsize{26}\\\tiny{16 - 41}}
% % & \cell{c}{-}
%  & \cell{c}{0.272}
%  & \cell{c}{--}
%  & \cell{c}{1.9$\%$}
%  & \cell{c}{0.851}
%  \\ 
   
% \midrule

\PLR{}
 & \cell{c}{Integrality\\Operational}
 & \cell{c}{\scriptsize{4.4$\%$}\\\tiny{3.3 - 6.5$\%$}}
 & \cell{c}{\scriptsize{0.742}\\\tiny{0.712 - 0.774}}
 & \cell{c}{\scriptsize{4}\\\tiny{3 - 4}}
% & \cell{c}{-}
 & \cell{c}{0.325}
& \cell{c}{--}
 & \cell{c}{3.9$\%$}
 & \cell{c}{0.771}
 \\ 

\midrule

\PLR{}\mto{}\mRd{}
 & \cell{c}{Operational}
 & \cell{c}{\scriptsize{7.0$\%$}\\\tiny{5.7 - 9.2$\%$}}
 & \cell{c}{\scriptsize{0.743}\\\tiny{0.705 - 0.786}}
 & \cell{c}{\scriptsize{2}\\\tiny{2 - 3}}
% & \cell{c}{3}
 & \cell{c}{0.329}
& \cell{c}{--}
 & \cell{c}{7.0$\%$}
 & \cell{c}{0.735}
 \\ 
   
\midrule

\PLR{}\mto{}\mRsRd{}
 & \cell{c}{Operational}
 & \cell{c}{\scriptsize{12.4$\%$}\\\tiny{11.2 - 13.6$\%$}}
 & \cell{c}{\scriptsize{0.761}\\\tiny{0.733 - 0.815}}
 & \cell{c}{\scriptsize{4}\\\tiny{4 - 4}}
% & \cell{c}{14}
 & \cell{c}{2.109}
& \cell{c}{--}
 & \cell{c}{12.5$\%$}
 & \cell{c}{0.760}
 \\ 
   
\midrule

\PLR{}\mto{}\mUnit{}
 & \cell{c}{Operational}
 & \cell{c}{\scriptsize{24.6$\%$}\\\tiny{23.6 - 25.7$\%$}}
 & \cell{c}{\scriptsize{0.759}\\\tiny{0.732 - 0.813}}
 & \cell{c}{\scriptsize{4}\\\tiny{4 - 4}}
% & \cell{c}{10}
 & \cell{c}{0.520}
& \cell{c}{--}
 & \cell{c}{24.8$\%$}
 & \cell{c}{0.759}
 \\ 
\bottomrule
\end{tabular}
}
\caption{Performance of risk scores for seizure prediction, and feasibility with respect to constraints. We report the 5-CV mean test CAL and 5-CV mean test AUC. The ranges in each cell represent the 5-CV minimum and maximum. We present the risk scores built using each method in Figures \ref{Fig::model_seizure_L0_4_RiskSLIM} to \ref{Fig::model_seizure_L0_4_RSLP_SRDCD}.}
\label{Table::DemoPerformanceTable}
\end{table}

\paragraph{On Performance and Usability in a Constrained Setting}

The results in Table~\ref{Table::DemoPerformanceTable} show the potential performance benefits of training an optimized risk score for problems with non-trivial constraints. Here,  \RiskSLIM{}  has a 5-CV mean test CAL/AUC of 2.5\%/0.801 while the best risk score built using a heuristic method has a 5-CV mean test CAL/AUC of 2.8\%/0.745

In contrast to the experiments in Section~\ref{Sec::Experiments}, only \RiskSLIM{} and our pooled methods were able to find a feasible risk score under the constraints. Traditional methods (e.g., \PLR{}\mto{}\mRd{}, \PLR{}\mto{}\mRsRd{}, and \PLR{}\mto{}\mUnit{}) violate one or more constraints after rounding. In fact, these methods cannot produce a risk score with comparable performance to the \RiskSLIM{} risk score even when these constraints are relaxed. If we consider only simple constraints on model size and monotonicity, then risk scores produced by these methods have a test AUC of at most 0.761 (\PLR{}\mto{}\mRsRd{}) and a test CAL of at least 7.0\% (\PLR{}\mto{}\mRd{}).

As shown in Figures \ref{Fig::model_seizure_L0_4_RiskSLIM} to \ref{Fig::model_seizure_L0_4_RSLP_SRDCD}, risk scores with similar test CAL can still exhibit important differences in terms of calibration. Here, the risk predictions of the \RiskSLIM{} model are monotonic and within the boundaries of the risk predictions of the fold-based models. In comparison, the risk predictions of other models do not monotonically increase with observed risk and vary significantly across test folds (e.g., \PLRSRDCD{}). As noted by our collaborators, such issues affect model adoption: the monotonicity violations of the \PLRSR{} model suggest that patients with a score of 3.5 may have more seizures compared to patients with a score of 4.0, eroding trust in the model's risk predictions.

Although the risk scores in Figures \ref{Fig::model_seizure_L0_4_RiskSLIM} to \ref{Fig::model_seizure_L0_4_RSLP_SRDCD} obey all of the constraints specified by our collaborators, they exhibit differences in terms of usability. The \RiskSLIM{} model requires physicians to check cEEG output for a single cEEG pattern (LPD). In comparison, other risk scores include the feature \emph{PatternsInclude BiPD or LRDA or LPD}, which can require physicians to check cEEG output for 3 patterns in the worst case. The \PLRSRDCD{} risk score also uses \textfn{MaxFrequencyLPD}, which requires estimating the frequency of LPD and thereby requires more time. %Note that it is possible, as a simple extension of our approach, to fine-tune usability by penalizing each feature based on how difficult it is for a physician to evaluate it.

\paragraph{On the Value of the Optimality Gap in Practice}

The results in Table \ref{Table::DemoPerformanceTable} illustrate how heuristics may lead practitioners to overestimate the true impact of real-world constraints on model performance. Here: three traditional methods could not output a feasible risk score; six pooled methods produced feasible risk scores with suboptimal AUC and calibration issues; and a baseline \PLR{} model with real-valued coefficients has an AUC of 0.742. Based on these results, a practitioner might inadvertently conclude that no feasible risk score could achieve a test AUC of 0.801.

In contrast, \RiskSLIM{} models are paired with an optimality gap. In practice, a small optimality gap suggests that we have trained the best possible risk score that satisfies a specific set of constraints. Thus, if a risk score with a small optimality gap performs poorly on training data, and the model generalizes (e.g., its training performance is similar to its K-CV performance), then one can attribute the performance deficit of the model to overly restrictive constraints and improve performance by relaxing them.

This provides a general mechanism to evaluate the effect of constraints on performance. If, for example, that our collaborators were not satisfied with the performance or usability of our model, then we could train certifiably optimal risk scores for different sets of constraints. By comparing the performance of certifiably optimal models, our collaborators could evaluate the true impact of their requirements on predictive performance, and navigate trade-offs in an informed manner. This approach helped our collaborators decide between a model with 4 features or 5 features. Here, we trained a \RiskSLIM{} risk score with 5 features, which has a 5-CV test CAL/AUC of 3.4\%/0.816. However, the slight improvement in test AUC did not outweigh the fact that the larger model included the feature \textfn{MaxFreqFactorAnyPattern$\;\geq\;2$} which could increase the model evaluation time by doctors.
	
\paragraph{On the Challenges of Handling Operational Constraints}

One of the practical benefits of our approach is that it can address constraints without parameter tuning or post-processing. Since our approach can directly incorporate constraints into the MINLP formulation, all \RiskSLIM{} models are feasible. Thus we can produce a feasible risk score and estimate its predictive performance by training 6 models: 1 final model fit on the full dataset for deployment, and 5 models trained on subsets of the training data to produce an unbiased performance estimate for the algorithm via 5-fold CV.

In contrast, the pooled methods produce a feasible model by post-processing a large pool of models and discarding those that are infeasible. Since we must then choose between feasible models on the basis of 5-CV performance, we must use a nested CV setup to pair any model with an unbiased performance estimate. This requires fitting a total of 33,000 models.\footnote{A nested CV setup with 5 outer folds, 5 inner folds, and 1,100 free parameter instances requires fitting $1,100 \times 5 \times (5 + 1) = 33,000$ models.} In general settings, there is no guarantee that pooled methods will produce a feasible risk score. In this case, for instance, only 12\% of the instances for the pooled methods satisfied all constraints (see Table~\ref{Table::DemoFeasibilityTable} in Appendix~\ref{Appendix::SeizurePrediction}). This kind of massive testing can become computationally burdensome.

Our results highlight other issues with methods that aim to address constraints by parameter tuning. Let us say we would use a standard  $K$-fold CV setup to tune parameters. In this case, we would train models on $K$ validation folds for each instance of the free parameters, choose the instance of the free parameters that maximizes the mean $K$-CV test AUC without violating any constraints, and then train a ``final model" for this instance. Unfortunately, there is no guarantee that the final model will obey all constraints. 

\paragraph{On the Benefits of Risk Scores with Small Integer Coefficients}

Figures \ref{Fig::model_seizure_L0_4_RiskSLIM} to \ref{Fig::model_seizure_L0_4_RSLP_SRDCD} illustrate some of the practical benefits of risk scores with small integer coefficients. When input variables belong to a small discrete set, the scores also belong to a small discrete set. This reduces the number of operating points on the ROC curve and reliability diagram, which makes it easier to pick an operating point. Further, when input variables are binary, the decision rules at each operating point can be represented as a Boolean rule. For the \RiskSLIM{} model in Figure~\ref{Fig::model_seizure_L0_4_RiskSLIM}, for example, the decision rule:
{\small
\begin{align*}
\hat{y}_i=+1 ~\text{if}~ &\text{score} \geq 2
\intertext{is equivalent to the Boolean function:}
\hat{y}_i=+1 ~\text{if}~ & \textfn{BriefRhythmicDischarge}\\
& OR~ \textfn{PatternsIncludeLPD}\\ 
& OR~ (\textfn{PriorSeizure} ~AND~ \textfn{EpiletiformDischarge})
\end{align*}
}
Small integer coefficients let users extract such rules by listing conditions when the score exceeds the threshold. This is more challenging when a model uses real-valued coefficients, as shown by the score function of the \PLR{} model from Table~\ref{Table::DemoPerformanceTable}:
{\small
\begin{align*}
\text{score} = & -2.35 + 0.91\cdot\textfn{PatternsIncludeBiPD\,or\,LRDA\,or\,LPD} + 0.03\cdot\textfn{PriorSeizure} \\ 
 & + 0.61\cdot\textfn{MaxFrequencyLPD}.
\end{align*}
}
In this case, extracting a Boolean function is difficult as computing the score involves arithmetic with real-valued coefficients and the real-valued variable \textfn{MaxFrequencyLPD}.

%%%%%%%%%%%%%%%%%%%%%%%%%%%%%%%%%%%%%%%%%%%%%%%%%%%
% RiskSLIM 
%%%%%%%%%%%%%%%%%%%%%%%%%%%%%%%%%%%%%%%%%%%%%%%%%%%

%\begin{table}[htbp]
%\begin{tabular}{lcc}
%  \toprule \bfcell{c}{Metric} & \bfcell{c}{All Data} & \bfcell{c}{5-Fold CV} \\ 
%   \midrule CAL (Mean Calibration Error @ Distinct Scores) & \cell{c}{2.0$\%$} & \cell{c}{\scriptsize{2.5$\%$}\\\tiny{2.0 - 3.4$\%$}} \\ 
%   \midrule MCAL (Maximum Calibration Error @ Distinct Scores) & \cell{c}{9.7$\%$} & \cell{c}{\scriptsize{22.2$\%$}\\\tiny{13.1 - 32.8$\%$}} \\ 
%   \midrule Model Size & \cell{c}{4} & \cell{c}{\scriptsize{4}\\\tiny{4 - 4}} \\ 
%   \midrule AUC & \cell{c}{0.806} & \cell{c}{\scriptsize{0.801}\\\tiny{0.758 - 0.841}} \\ 
%   \midrule Logistic Loss & \cell{c}{0.293} & \cell{c}{\scriptsize{0.295}\\\tiny{0.271 - 0.329}} \\ 
%   \midrule Optimality Gap & \cell{c}{0.0$\%$} & \cell{c}{\scriptsize{0.0$\%$}\\\tiny{0.0 - 0.0$\%$}} \\ 
%   \midrule \# of Distinct Scores & \cell{c}{7} & \cell{c}{\scriptsize{7.0}\\\tiny{6 - 8}} \\ 
%   \midrule \% of Trivial Models & \cell{c}{0.0} & \cell{c}{\scriptsize{ NA}\\\tiny{ NA -  NA}} \\ 
%   \bottomrule
%\end{tabular}
%\label{Table::StatsTable}
%\end{table}
%
\begin{figure}[htbp]
\centering{
\scoringsystem{}
\begin{tabular}{|l l  c | l |}
   \hline
1. & \textssm{BriefRhythmicDischarge} & 2 points & $\phantom{+}\prow{}$ \\ 
  2. & \textssm{PatternsInclude\,LPD} & 2 points & $+\prow{}$ \\ 
  3. & \textssm{PriorSeizure} & 1 point & $+\prow{}$ \\ 
  4. & \textssm{EpiletiformDischarge} & 1 point & $+$ \\ 
   \hline
 & \instruction{1}{4} & \scorelabel{} & $=$ \\ 
   \hline
\end{tabular}
\risktable{}
\begin{tabular}{|l|c|c|c|c|c|c|c|}
   \hline
\rowcolor{scorecolor}\scorelabel{} & 0 & 1 & 2 & 3 & 4 & 5 & 6 \\ 
   \hline
\rowcolor{riskcolor}\risklabel{} & 4.7$\%$ & 11.9$\%$ & 26.9$\%$ & 50.0$\%$ & 73.1$\%$ & 88.1$\%$ & 95.3$\%$ \\ 
   \hline
\end{tabular}

\vspace{1.00em}
\begin{tabular}{@{}lr@{}}
\demoplot{4}{risk_slim}{calibration_plot} & \demoplot{4}{risk_slim}{roc_plot} %& \appdemoplot{4}{RSLP_CR}{calibration_histogram} 
\end{tabular}
}
\caption{\RiskSLIM{} risk score (top), reliability diagram (bottom left), and ROC curve (bottom right) for the \textds{seizure} dataset. We plot results for the final model on training data in black, and the 5 fold models on test data in grey. This model has a 5-CV mean test CAL/AUC of 2.5\%/0.801.}
\label{Fig::model_seizure_L0_4_RiskSLIM}
\end{figure}

%%%%%%%%%%%%%%%%%%%%%%%%%%%%%%%%%%%%%%%%%%%%%%%%%%%
% CR
%%%%%%%%%%%%%%%%%%%%%%%%%%%%%%%%%%%%%%%%%%%%%%%%%%%

%\begin{table}[htbp]
%\begin{tabular}{lcc}
%  \toprule \bfcell{c}{Metric} & \bfcell{c}{All Data} & \bfcell{c}{5-Fold CV} \\ 
%   \midrule CAL (Mean Calibration Error @ Distinct Scores) & \cell{c}{6.0$\%$} & \cell{c}{\scriptsize{5.3$\%$}\\\tiny{3.1 - 7.1$\%$}} \\ 
%   \midrule MCAL (Maximum Calibration Error @ Distinct Scores) & \cell{c}{52.2$\%$} & \cell{c}{\scriptsize{37.7$\%$}\\\tiny{28.4 - 54.1$\%$}} \\ 
%   \midrule Model Size & \cell{c}{2} & \cell{c}{\scriptsize{2}\\\tiny{1 - 3}} \\ 
%   \midrule AUC & \cell{c}{0.752} & \cell{c}{\scriptsize{0.740}\\\tiny{0.712 - 0.757}} \\ 
%   \midrule Logistic Loss & \cell{c}{0.350} & \cell{c}{\scriptsize{0.343}\\\tiny{0.303 - 0.381}} \\ 
%   \midrule Optimality Gap & \cell{c}{-} & \cell{c}{\scriptsize{-}\\\tiny{NA - NA$\%$}} \\ 
%   \midrule \# of Distinct Scores & \cell{c}{3} & \cell{c}{\scriptsize{4.6}\\\tiny{2 - 7}} \\ 
%   \midrule \% of Trivial Models & \cell{c}{0.0} & \cell{c}{\scriptsize{0.00}\\\tiny{0.00 - 0.00}} \\ 
%   \bottomrule
%\end{tabular}
%\end{table}

\begin{figure}[htbp]
\centering{
\scoringsystem{}
\begin{tabular}{|l l  c | l |}
   \hline
1. & \textssm{BriefRhythmicDischarge} & 1 point & $\phantom{+}\prow{}$ \\ 
  2. & \textssm{PatternsIncludeBiPD\,or\,LRDA\,or\,LPD} & 1 point & $+$ \\ 
   \hline
 & \instruction{1}{4} & \scorelabel{} & $={}$ \\ 
   \hline
\end{tabular}
\risktable{}
\begin{tabular}{|l|c|c|c|}
   \hline
\rowcolor{scorecolor}\scorelabel{} & 0 & 1 & 2 \\ 
   \hline
\rowcolor{riskcolor}\risklabel{} & 4.7$\%$ & 11.9$\%$ & 26.9$\%$ \\ 
   \hline
\end{tabular}

\vspace{1.00em}
\begin{tabular}{@{}lr@{}}
\demoplot{4}{RSLP_CR}{calibration_plot} & \demoplot{4}{RSLP_CR}{roc_plot} %& \appdemoplot{4}{RSLP_CR}{calibration_histogram} 
\end{tabular}
}
\caption{\PLRCR{} risk score (top), reliability diagram (bottom left), and ROC curve (bottom right) for the \textds{seizure} dataset. We plot results for the final model on training data in black, and results for the fold models on test data in grey. This model has a 5-CV mean test CAL/AUC of 5.3/0.740\%.}
\label{Fig::model_seizure_L0_4_CR}
\end{figure}

%%%%%%%%%%%%%%%%%%%%%%%%%%%%%%%%%%%%%%%%%%%%%%%%%%%
% CRDCD 
%%%%%%%%%%%%%%%%%%%%%%%%%%%%%%%%%%%%%%%%%%%%%%%%%%%

%\begin{table}[htbp]
%\begin{tabular}{lcc}
%  \toprule \bfcell{c}{Metric} & \bfcell{c}{All Data} & \bfcell{c}{5-Fold CV} \\ 
%   \midrule
%CAL (Mean Calibration Error @ Distinct Scores) & \cell{c}{1.9$\%$} & \cell{c}{\scriptsize{3.0$\%$}\\\tiny{1.4 - 3.6$\%$}} \\ 
%   \midrule MCAL (Maximum Calibration Error @ Distinct Scores) & \cell{c}{9.0$\%$} & \cell{c}{\scriptsize{20.6$\%$}\\\tiny{10.2 - 38.1$\%$}} \\ 
%   \midrule Model Size & \cell{c}{2} & \cell{c}{\scriptsize{2}\\\tiny{1 - 3}} \\ 
%   \midrule AUC & \cell{c}{0.754} & \cell{c}{\scriptsize{0.745}\\\tiny{0.712 - 0.776}} \\ 
%   \midrule Logistic Loss & \cell{c}{0.308} & \cell{c}{\scriptsize{0.319}\\\tiny{0.282 - 0.338}} \\ 
%   \midrule Optimality Gap & \cell{c}{-} & \cell{c}{\scriptsize{-}\\\tiny{NA - NA$\%$}} \\ 
%   \midrule \# of Distinct Scores & \cell{c}{4} & \cell{c}{\scriptsize{5.6}\\\tiny{2 - 9}} \\ 
%   \midrule \% of Trivial Models & \cell{c}{0.0} & \cell{c}{\scriptsize{0.00}\\\tiny{0.00 - 0.00}} \\ 
%   \bottomrule
%\end{tabular}
%\label{Table::StatsTable}
%\end{table}

\begin{figure}[htbp]
\centering{
\scoringsystem{}
\begin{tabular}{|l l  c | l |}
   \hline
1. & \textssm{BriefRhythmicDischarge} & 3 points & $\phantom{+}\prow{}$ \\ 
  2. & \textssm{PatternsIncludeBiPD\,or\,LRDA\,or\,LPD} & 2 points & $+$ \\ 
   \hline
 & \instruction{1}{2} & \scorelabel{} & $=$ \\ 
   \hline
\end{tabular}
\risktable{}
\begin{tabular}{|l|c|c|c|c|}
   \hline
\rowcolor{scorecolor}\scorelabel{} & 0 & 2 & 3 & 5 \\ 
   \hline
\rowcolor{riskcolor}\risklabel{} & 4.7$\%$ & 26.9$\%$ & 50.0$\%$ & 88.1$\%$ \\ 
   \hline
\end{tabular}

\vspace{1.00em}
\begin{tabular}{@{}lr@{}}
\demoplot{4}{RSLP_CRDCD}{calibration_plot} & \demoplot{4}{RSLP_CRDCD}{roc_plot} %& \appdemoplot{4}{RSLP_CR}{calibration_histogram} 
\end{tabular}
}
\caption{\PLRCRDCD{} risk score (top), reliability diagram (bottom left), and ROC curve (bottom right) for the \textds{seizure} dataset. We plot results for the final model on training data in black, and results for the fold models on test data in grey. This model has a 5-CV mean test CAL/AUC of 3.0/0.745\%.}
\label{Fig::model_seizure_L0_4_CRDCD}
\end{figure}

%%%%%%%%%%%%%%%%%%%%%%%%%%%%%%%%%%%%%%%%%%%%%%%%%%%
% SRDCD
%%%%%%%%%%%%%%%%%%%%%%%%%%%%%%%%%%%%%%%%%%%%%%%%%%%
%\begin{table}[htbp]
%\begin{tabular}{lcc}
%  \toprule \bfcell{c}{Metric} & \bfcell{c}{All Data} & \bfcell{c}{5-Fold CV} \\ 
%   \midrule CAL (Mean Calibration Error @ Distinct Scores) & \cell{c}{1.9$\%$} & \cell{c}{\scriptsize{2.8$\%$}\\\tiny{2.4 - 3.1$\%$}} \\ 
%   \midrule MCAL (Maximum Calibration Error @ Distinct Scores) & \cell{c}{18.2$\%$} & \cell{c}{\scriptsize{14.5$\%$}\\\tiny{8.7 - 23.4$\%$}} \\ 
%   \midrule Model Size & \cell{c}{3} & \cell{c}{\scriptsize{3}\\\tiny{2 - 4}} \\ 
%   \midrule AUC & \cell{c}{0.767} & \cell{c}{\scriptsize{0.745}\\\tiny{0.713 - 0.805}} \\ 
%   \midrule Logistic Loss & \cell{c}{0.313} & \cell{c}{\scriptsize{0.320}\\\tiny{0.303 - 0.344}} \\ 
%   \midrule Optimality Gap & \cell{c}{-} & \cell{c}{\scriptsize{-}\\\tiny{NA - NA$\%$}} \\ 
%   \midrule \# of Distinct Scores & \cell{c}{9} & \cell{c}{\scriptsize{5.2}\\\tiny{4 - 7}} \\ 
%   \midrule \% of Trivial Models & \cell{c}{0.0} & \cell{c}{\scriptsize{0.00}\\\tiny{0.00 - 0.00}} \\ 
%   \bottomrule
%\end{tabular}
%\label{Table::StatsTable}
%\end{table}

\begin{figure}[htbp]
\centering{
\scoringsystem{}
\begin{tabular}{|l l  r | l |}
   \hline
1. & \textssm{PriorSeizure} & 1 point & $\phantom{+}\prow{}$ \\ 
  2. & \textssm{PatternsIncludeBiPD\,or\,LRDA\,or\,LPD} & 1 point & $+\prow{}$ \\ 
  3. & \textssm{MaxFreqFactorLPD} & $\times$ 1 point per Hz & $+$ \\ 
   \hline
 & \instruction{1}{3} & \scorelabel{} & $= $ \\ 
   \hline
\end{tabular}
\risktable{}
\begin{tabular}{|l|c|c|c|c|c|c|c|c|c|c|}
   \hline
\rowcolor{scorecolor}\scorelabel{} & 0 & 1 & 2 & 2.5 & 3 & 3.5 & 4 & 4.5 & 5 \\ 
   \hline
\rowcolor{riskcolor}\risklabel{} & 4.7$\%$ & 11.9$\%$ & 26.9$\%$ & 37.8$\%$ & 50.0$\%$ & 62.2$\%$ & 73.1$\%$ & 81.8$\%$ & 88.1$\%$ \\ 
   \hline
\end{tabular}

\vspace{1.00em}
\begin{tabular}{@{}lr@{}}
\demoplot{4}{RSLP_SRDCD}{calibration_plot} & \demoplot{4}{RSLP_SRDCD}{roc_plot} %& \appdemoplot{4}{RSLP_LR}{calibration_histogram} 
\end{tabular}
}
\caption{\PLRSRDCD{} model (top), reliability diagram (bottom left), and ROC curve (bottom right) for the \textds{seizure} dataset. We plot results for fold models on test data in grey, and for the final model on training data in black. This model has a 5-CV mean test CAL/AUC of 2.8\%/0.745.}
\label{Fig::model_seizure_L0_4_RSLP_SRDCD}
\end{figure}

%% file: section_07_discussion.tex
Risk scores are simple tools that are often used to inform consequential decisions, such as mortality prediction and loan approval. Despite the fact that risk scores have been used in such tasks for nearly a century, they are often developed using \emph{ad hoc} methods -- i.e., by combining machine learning methods with heuristics~\citep[see e.g., the pipeline for the TIMI risk score in][in Appendix \ref{Appendix::RiskScoreBackground}]{antman2000timi}. In practice, such approaches may produce a risk score that violates important requirements, leading practitioners to make manual adjustments, or to build models without data~\citep[e.g., via a panel of domain experts as in][]{gage2001validation,mcginley2012national}. Such approaches do not produce risk scores with performance guarantees, and may result in the deployment of models with poor calibration or rank accuracy. Ad hoc model development also has hidden costs. For example, the need to implement a specialized training pipeline slows down model development.

Our goal in this paper was to design a machine learning approach to build risk scores that standardizes model development, reduces the need for domain experts to manually specify and tune models, and pairs models with useful guarantees. Our approach trains risk scores by solving an empirical risk minimization problem that performs exact feature selection while restricting coefficients to small integers and enforcing application-specific constraints. Since commercial solvers did not handle this problem well, we solved it with a new cutting plane algorithm (\LCPA{}), which allows training to scale linearly with the number of samples, and can be used to solve other empirical risk minimization problems with non-convex regularizers or constraints. As shown in Sections \ref{Sec::Experiments} and \ref{Sec::SeizurePrediction}, \LCPA{} can train certifiably optimal risk scores for real-world datasets in minutes. These models often attain best-in-class calibration and rank accuracy, and avoid pitfalls of heuristics used in practice. Our approach simplifies model development, by letting practitioners customize risk scores without post-processing. In addition, it pairs models with a certificate of optimality, which can be used to understand how application-specific constraints affect predictive performance.

%% file: appendix_proofs.tex
\begin{proof}[Remark  \ref{Rem::LCPAMaxCuts}]
We first explain why \LCPA{} attains the optimal objective value, and then justify the upper bounds on the number of cuts and number of nodes.

Observe that \LCPA{} finds the optimal solution to \RSMINLP{} through an exhaustive search over the feasible region $\Lset$. Thus, \LCPA{} is bound to encounter the optimal solution, since it only discards a node $(\bbval{t}, \bbpart{t})$ if: (i) the surrogate problem is infeasible over $\bbpart{t}$ (in which case \rsminlp{} is also infeasible over $\bbpart{t}$); or (ii) the surrogate problem has an objective value that exceeds than $\objvalmax{}$ (in which case, any integer feasible solution $\bbpart{t}$ is also suboptimal).

The bound on the number of cuts follows from the fact that Algorithm \ref{Alg::LCPA} only adds cuts at integer feasible solutions, of which there are at most $|\Lset|$. The bound on the number of processed nodes represents a worst-case limit produced by bounding the depth of the branch-and-bound tree. To do this, we exploit the fact that the splitting rule $\bbsplit{}$ splits a partition into two mutually exclusive subsets by adding integer-valued bounds such $\lambda_j \geq \ceil{\lambda_j}$ and $\lambda_j \leq \floor{\lambda_j} - 1$ on a single coefficient to the feasible solution. Consider applying $\bbsplit{}$ a total of $\maxlambda{j} - \minlambda{j} + 1$ times in succession on a fixed dimension $j$. This results in a total of $\maxlambda{j} - \minlambda{j} + 1$ nodes where each node fixes the coefficient in dimension $j$ to an integer value $\lambda_j \in \{\minlambda{j},\ldots,\maxlambda{j}\}$. Pick any node and repeat this process for coefficients in the remaining dimensions. The resulting B\&B tree will have at most $2^{|\Lset|} - 1$ leaf nodes where $\lambdab$ is restricted to integer feasible solutions.
\end{proof}

\clearpage
\begin{proof}[Proposition \ref{Prop::LossBounds}]
Since the coefficient set $\Lset$ is bounded, the data are $(\xb_i, y_i)_{i=1}^\n$ discrete, and the normalized logistic loss function $\lossfun{\lambdab}$ is continuous, it follows that the value of $\lossfun{\lambdab}$ is bounded:%
$$\lossfun{\lambdab} \in [\min_{\lambdab \in \Lset} \lossfun{\lambdab}, \max_{\lambdab \in \Lset}, \lossfun{\lambdab}] \text{ for all } \lambdab \in \Lset.$$%
Thus we need only to show that $\lossmin \leq \min_{\lambdab \in \Lset} \lossfun{\lambdab}$, and $\lossmax \geq \max_{\lambdab \in \Lset} \lossfun{\lambdab}$. 
For the lower bound, we observe that:
\begin{align}
\min_{\lambdab \in \Lset} \lossfun{\lambdab}  &= \min_{\lambdab \in \Lset} \logloss{\lambdab} \nonumber \\
& = \min_{\lambdab \in \Lset} \frac{1}{\n} \sum_{i:y_i = +1} \log (1+ \exp(- \dotprod{\lambdab}{\xb_i})) + \frac{1}{\n} \sum_{i:y_i = -1} \log (1+ \exp(\dotprod{\lambdab}{\xb_i})) \nonumber \\
& \geq \frac{1}{\n} \sum_{i:y_i = +1}  \min_{\lambdab \in \Lset} \log (1+ \exp(- \dotprod{\lambdab}{\xb_i})) + \frac{1}{\n} \sum_{i:y_i = -1}  \min_{\lambdab \in \Lset} \log (1+ \exp(\dotprod{\lambdab}{\xb_i})) \nonumber \\
& = \frac{1}{\n} \sum_{i:y_i = +1} \log (1+ \exp(- \max_{\lambdab \in \Lset} \dotprod{\lambdab}{\xb_i})) + \frac{1}{\n} \sum_{i:y_i = -1} \log (1+ \exp(\min_{\lambdab \in \Lset} \dotprod{\lambdab}{\xb_i})) \nonumber \\ 
& = \frac{1}{\n} \sum_{i:y_i = +1} \log (1+ \exp(- s_i^{\max}) + \frac{1}{\n} \sum_{i:y_i = -1} \log (1+ \exp(s_i^{\min})) \nonumber\\
& = \lossmin. \nonumber
\end{align}
The upper bound can be derived in a similar manner.
\end{proof}

\clearpage
\begin{proof}[Proposition \ref{Prop::ZeroNormUB}]
We are given that $\objvalmax \geq \objval{\lambdab^\opt}$ where $\objval{\lambdab^\opt} := \lossfun{\lambdab^\opt} + C_0 \zeronorm{\lambdab^\opt}$ by definition. Thus, we can recover the upper bound from Proposition \ref{Prop::ZeroNormUB} as follows:
\begin{align}
\lossfun{\lambdab^\opt} + C_0 \zeronorm{\lambdab^\opt} & \leq \objvalmax,  \nonumber \\
\zeronorm{\lambdab^\opt}  & \leq \frac{\objvalmax - \lossfun{\lambdab^\opt}}{C_0}, \nonumber \\
\zeronorm{\lambdab^\opt}  & \leq \frac{\objvalmax - \lossmin}{C_0}, \label{Eq::ZeroNormUBMin}\\
\zeronorm{\lambdab^\opt}  & \leq \floor{\frac{\objvalmax - \lossmin}{C_0}}. \label{Eq::ZeroNormUBInt}
\end{align}
Here, \eqref{Eq::ZeroNormUBMin} follows from the fact that $\lossmin \leq \lossfun{\lambdab^\opt}$ by definition, and \eqref{Eq::ZeroNormUBInt} follows from the fact that the number of non-zero coefficients is a natural number.
\end{proof}

\begin{proof}[Proposition \ref{Prop::LossUB}]
We are given that $\objvalmax \geq \objval{\lambdab^\opt}$ where $\objval{\lambdab^\opt} := \lossfun{\lambdab^\opt} + C_0 \zeronorm{\lambdab^\opt}$ by definition. Thus, we can recover the upper bound from Proposition \ref{Prop::LossUB} as follows:
\begin{align*}
\lossfun{\lambdab^\opt} + C_0 \zeronorm{\lambdab^\opt} &\leq \objvalmax, \\
\lossfun{\lambdab^\opt}  & \leq \objvalmax - C_0 \zeronorm{\lambdab^\opt}, \\
\lossfun{\lambdab^\opt}  & \leq \objvalmax - C_0 \zeronormmin.
\end{align*}
Here, the last line follows from the fact that $\zeronormmin \leq \zeronorm{\lambdab^\opt}$ by definition.
\end{proof}

\begin{proof}[Proposition \ref{Prop::LossLB}]
We are given that $\objvalmin \leq \objval{\lambdab^\opt}$ where $\objval{\lambdab^\opt} := \lossfun{\lambdab^\opt} + C_0 \zeronorm{\lambdab^\opt}$ by definition. Thus, we can recover the lower bound from Proposition \ref{Prop::LossLB} as follows:
\begin{align*}
\lossfun{\lambdab^\opt} + C_0 \zeronorm{\lambdab^\opt} & \geq \objvalmin, \\
\lossfun{\lambdab^\opt}  & \geq \objvalmin - C_0 \zeronorm{\lambdab^\opt}, \\
\lossfun{\lambdab^\opt}  & \geq \objvalmin - C_0 \zeronormmax.
\end{align*}
Here, the last line follows from the fact that $\zeronormmax \geq \zeronorm{\lambdab^\opt}$ by definition.
\end{proof}

%% file: appendix_tradeoff_parameter.tex
\newcommand{\minlosssym}{L}
\newcommand{\argminlosssym}{\mathcal{M}}
\newcommand{\minloss}{\minlosssym{}}
\newcommand{\argminloss}{\argminlosssym{}}
\newcommand{\minlossk}[1]{L({#1})}
\newcommand{\argminlossk}[1]{\argminlosssym{}({#1})}
\newcommand{\nterms}{k}
\newcommand{\txsparsest}{\textnormal{opt}}
\newcommand{\optterms}{\nterms^\txsparsest{}}
\newcommand{\optlambda}{\lambdab^\txsparsest{}}

In Section \ref{Sec::ProblemStatement}, we state that if the trade-off parameter $C_0$ in the objective of the risk score problem is sufficiently small, then its optimal solution will attain the best possible trade-off between logistic loss and sparsity. In what follows, we formalize this statement. In what follows, we will omit the intercept term for clarity and explicitly show the regularization parameter $C_0$ in the \RiskSLIM{} objective so that $\objvalfun{\lambdab; C_0} := \lossfun{\lambdab} + C_0 \zeronorm{\lambdab}$. We use some new notation shown in Table \ref{Table::NotationForSparsity}.

\begin{table}[htbp]
\renewcommand{\arraystretch}{1.1}
\small\centering
\begin{tabular}{ll}
\toprule
\bfcell{l}{Notation} & \bfcell{l}{Description}\\
\toprule
$\argminloss{} = \argmin_{\lambdab \in \Lset} \lossfun{\lambdab}$
& \cell{l}{minimizers of the logistic loss} \\ \midrule
$\Lset(\nterms{}) = \{\lambdab \in \Lset ~|~ \zeronorm{\lambdab} \leq \nterms{}\}$
& \cell{l}{feasible coefficients of models with model size $\leq \nterms{}$} \\ \midrule
$\argminlossk{\nterms} = \argmin_{\lambdab \in \Lset(\nterms{})}  \lossfun{\lambdab}$
& \cell{l}{minimizers of the logistic loss among models of size $\leq \nterms{}$} \\ \midrule
$\minlossk{k} = \min_{\lambdab \in \Lset(\nterms{})} \lossfun{\lambdab}$
& \cell{l}{logistic loss of minimizers with model size $\leq\nterms{}$ } \\ \midrule
$\optlambda{} \in \argmin_{\lambdab \in \argminloss{}} \zeronorm{\lambdab}$
& \cell{l}{sparsest minimizers among all minimizers of the logistic loss} \\ \midrule
$\optterms{} = \zeronorm{\optlambda}$
& \cell{l}{model size of sparsest minimizer} \\ \bottomrule
\end{tabular}
\caption{Notation used in Remarks \ref{Rem::Minimizers} and \ref{Rem::SmallTradeOff}}
\label{Table::NotationForSparsity}
\end{table}

\begin{remark}[Minimizers of the Risk Score Problem]
\label{Rem::Minimizers}
Any optimal solution to the risk score problem will achieve a logistic loss of $\minlossk{\nterms{}}$ for some $\nterms{} \geq 0$:
$$\min_{\lambdab\in\Lset} \objvalfun{\lambdab; C_0} = \min_{\nterms\in\{0,1,\ldots,\optterms\}} \minlossk{\nterms{}} + C_0\nterms.$$
\end{remark}

\begin{remark}[Small Trade-Off Parameters Do Not Influence Accuracy]
\label{Rem::SmallTradeOff}
There exists an integer $z \geq 1$ such that if $$ C_0 <  \frac{1}{z}\left[\minlossk{\optterms{}-z} - \minlossk{\optterms{}}\right],$$ then $$\lambdab^\txsparsest{} \in \argmin_{\lambdab \in \Lset} \objvalfun{\lambdab;C_0}.$$ 
%For this $z$, the right hand side $\frac{1}{z} [\minlossk{\optterms{}-z} - \minlossk{\optterms{}}]$ is strictly greater than 0. 
\end{remark}
Remark \ref{Rem::SmallTradeOff} states that as long as $C_0$ is sufficiently small, the regularized objective is minimized by any model $\lambdab^\txsparsest{}$ that has the smallest size among the most accurate feasible models. In other words, there exists a small enough value of $C_0$ to guarantee that we will obtain the best possible solution. Since we do not know $z$ in advance and since it is just as difficult to compute $\minloss{}$ as it is to solve the risk score problem, we will not know in advance how small $C_0$ must be to avoid sacrificing sparsity. However, it does not matter which value of $C_0$ we choose as long as it is sufficiently small. In practice, we set $C_0 = 10^{-8}$, which is the smallest value that we can use without running into numerical issues with the MIP solver.

\clearpage
\begin{proof}[Remark \ref{Rem::Minimizers}] 
Since $\minlossk{\nterms}$ is the minimal value of the logistic loss for all models with at most $\nterms{}$ non-zero coefficients, we have:
\begin{align}
\label{Eq:MinLogis}
\minlossk{\nterms{}} +C_0\nterms{}  \leq \lossfun \lambdab +  C_0 \nterms{}  \textrm{ for any } \lambdab \in  \Lset(\nterms{})
\end{align}
Denote a feasible minimizer of $\objvalfun{\lambdab; C_0}$ as $\lambdab' \in \argmin_{\lambda\in\Lset} \objvalfun{\lambdab; C_0}$, and let $\nterms{}' = \zeronorm{\lambdab'}$. Since $\lambdab' \in \Lset(\nterms{}')$, we have that:%
\begin{align}
\minlossk{\nterms{}'}  + C_0\nterms{}'  \leq \objvalfun{\lambdab'; C_0}. \label{Eq::MinLog2}
\end{align}
%
%Consider the minimizer (over any constrained or unconstrained set) of the right side of (\ref{Eq::MinLogis}), where the right side is \RiskSLIM{}'s objective $\objvalfun{\lambdab;C_0}$. The right side then must be minimized by a $\lambdab$ that achieves the minimum of the logistic loss for its size (left side). This is true for all $\nterms{}$. Thus, we need only minimize over $\nterms$.
%
Taking the minimum of the left hand side of \eqref{Eq::MinLog2}:
\begin{align}
\min_{\nterms{} \in \{0,1,2,...,\optterms{}\}} \minlossk{\nterms{}} +C_0\nterms{} \leq \minlossk{\nterms{}'} +C_0\nterms{}'. \label{Eq::MinLog3}
\end{align}
Combining \eqref{Eq::MinLog2} and \eqref{Eq::MinLog3}, we get:
\begin{align*}
\min_{\nterms{} \in\{0,1,2,...,\optterms{}\}} \minlossk{\nterms{}} +C_0\nterms{} \leq \minlossk{\nterms{}'} + C_0\nterms{}' \leq \min_{\lambdab\in\Lset} \objvalfun{\lambdab; C_0}.
\end{align*}
If these inequalities are not all equalities, we have a contradiction with the definition of $\lambdab'$ and $\nterms{}'$ as the minimizer of $\objvalfun{\lambdab; C_0}$ and its size. So all must be equality. This proves the statement.
\end{proof}

\clearpage
\begin{proof}[Remark \ref{Rem::SmallTradeOff}]
Note that if $C_0 = 0$, then any minimizer of $\objvalfun{\lambdab;C_0}$ has $\optterms{}$ non-zero coefficients. Consider increasing the value of $C_0$ starting from zero until a threshold value $C_0^{\txmin{}}$, defined as the smallest value such that the minimizer can sacrifice some loss to remove at least one non-zero coefficient. Let $z \geq 1$ be the number of non-zero coefficients removed. For the threshold value $C_0^{\txmin{}}$ at which we choose the smaller model rather than the one with size $\optterms{}$, we have
$$\minlossk{\optterms{}} + {C_0^{\txmin{}}} \optterms{} \geq \minlossk{\optterms{}-z}+C_0^{\txmin{}}(\optterms{}-z).$$
Simplifying, we obtain:
$$\frac{1}{z}\left[\minlossk{\optterms{}-z} - \minlossk{\optterms{}}\right] \leq C_0^{\txmin{}}.$$
Thus, \RiskSLIM{} does not sacrifice sparsity for logistic loss when:
$$\frac{1}{z}\left[\minlossk{\optterms{}-z} - \minlossk{\optterms{}} \right] > C_0.$$ 
Here, we know that the value on the left hand side is greater than 0 because $\minlossk{\cdot}$ is decreasing in its argument, and if there was no strict decrease, there would be a contradiction with the definition of $\optterms{}$ as the 
smallest number of terms of an optimal model. In particular, 
$\minlossk{\optterms{}-z} = \minlossk{\optterms{}}$ implies that:
\begin{align*}
\min_{\lambdab} V(\lambdab;0) 
&= \minlossk{\optterms{}} + {C_0^{\txmin{}}} \optterms{}\\
&= \minlossk{\optterms{}-z}+C_0^{\txmin{}}(\optterms{})\\
&\geq \minlossk{\optterms{}-z}+C_0^{\txmin{}}(\optterms{}-z)\\
&\geq \min_{\lambdab} V(\lambdab;0).
\end{align*}
Thus all inequalities are equalities, which implies that $z=0$ and contradicts $z\geq 1$.
\end{proof}

%% file: appendix_background_material.tex
In Table \ref{Appendix::Table::FormatQuotes}, we present quotes from authors in medicine, criminal justice, and finance to support the claim that risk scores are used because they are easy to use, understand, and validate.

\begin{table}[htbp]
\tableformat{}
\begin{tabularx}{\textwidth}{ll>{\itshape\footnotesize}X}
\toprule
\textbf{Domain} & \textbf{Reference} & \textnormal{\small\textbf{Quote}} \\ 
\toprule

Medicine & \citet{than2014development} & ``Ease of use might be facilitated by presenting a rule developed from logistic regression as a score, where the original predictor weights have been converted to integers that are easy to add together... Though less precise than the original regression formula, such presentations are less complex, easier to apply by memory and usable without electronic assistance." \\ 
\midrule

\cell{l}{Criminal Justice} & \cite{duwe2016sacrificing} & ``It is commonplace... for fine-tuned regression coefficients to be replaced with a simple-point system... to promote the easy implementation, transparency, and interpretability of risk-assessment instruments." \\ 
\midrule

Finance & \cite{finlay2012credit} & ``presenting a linear model in the form of a scorecard is attractive because it's so easy to explain and use. In particular, the score can be calculated using just addition to add up the relevant points that someone receives"\\

\bottomrule
\end{tabularx}
\caption{Quotes on why risk scores with small integer coefficients are used in different domains.}
\label{Appendix::Table::FormatQuotes}
\end{table}

\subsection*{Importance of Operational Constraints}

The approaches used to create risk scores vary significantly for each problem. There is no standard approach within a given domain, \citep[see e.g., the different techniques proposed in criminal justice by][]{gottfredson2005mathematics,bobko2007usefulness,duwe2016sacrificing}, or a given application \citep[see, e.g., the different approaches used to create risk scores for cardiac illness,][]{six2008chest,antman2000timi,than2014development}.  

A key reason for the lack of a standardized approach is because risk scores in domains such as medicine and criminal justice need to obey additional \textit{operational constraints} to be used and accepted. In some cases, these constraints can be explicitly stated. \cite{reilly2006translating}, for example, describe the requirements put forth by physicians when building a model to detect major cardiac complications for patients with chest pain: 

\begin{quote}
``\textit{Our physicians... insisted that a new left bundle-branch block be considered as an electrocardiographic predictor of acute ischemia. In addition, they argued that patients who are stratified as low risk by the prediction rule could inappropriately include patients presenting with acute pulmonary edema, ongoing ischemic pain despite maximal medical therapy, or unstable angina after recent coronary revascularization (52). They insisted that such emergent clinical presentations be recommended for coronary care unit admission, not telemetry unit admission}."
\end{quote}

In other cases, however, operational constraints may depend on qualities that are difficult to define a priori. Consider for example, the following statement of \citet{than2014development}, that describes the importance of sensibility for deployment: 

\begin{quote}
``\textit{An important consideration during development is the clinical sensibility of the resulting prediction rule [...] Evaluation of sensibility requires judgment rather than statistical methods. A sensible rule is easy to use, and has content and face validities. Prediction rules are unlikely to be applied in practice if they are not considered sensible by the end-user, even if they are accurate.}"
\end{quote}

\subsection*{Approaches to Model Development}

Common heuristics used in model development include:

\begin{itemize}

\item \textit{Heuristic Feature Selection}: Many approaches use heuristic feature selection to reduce the number of variables in the model. Model development pipelines can often involve multiple rounds of feature selection, and may use different heuristics at each stage (e.g., \citealt{antman2000timi} use a significance test to remove weak predictors, then use approximate feature selection via forward stepwise regression).

\item \textit{Heuristic Rounding}: Many approaches use rounding heuristics to produce models with integer coefficients. In the simplest case, this involves scaling and rounding the coefficients from a logistic regression model \citep[e.g.,][]{goel2015precinct} or a linear probability model \citep[e.g.,][]{usdoj2005riskclassification}. The SAPS II score \citep{le1993new}, for example, was built in this way (``\textit{the general rule was to multiply the $\beta$ for each range by 10 and round off to the nearest integer.}")

%\item \textit{Empirical Calibration}: In risk assessment applications, many approaches determine the coefficients of the model using logistic regression. Although these models are capable of generating a predicted risk estimates for each score through the logit function, the final model typically uses empirical risk estimates determined using an out-of-sample population \citep{six2008chest}. 

\item \textit{Expert Judgement}: A common approach to model development involves having a panel experts build a model by hand, and using data to validate the model after it is built (e.g., for the CHADS$_2$ score for stroke prediction of \citealt{gage2001validation}, and the National Early Warning Score to assess acute illness in the ICU of \citealt{mcginley2012national}). Expert judgement can also be used in data-driven approaches. In developing the EDACS score \citep{than2014development}, for example, expert judgement was used to: (i) determine a scaling factor for model coefficients (``\textit{The beta coefficients were multiplied by eight, which was the smallest common multiplication factor possible to obtain a sensible score that used whole numbers and facilitated clinical ease of use.}") (ii) convert a continuous variable into a binary variables (``\textit{Age was the only continuous variable to be included in the final score. It was converted to a categorical variable, using 5-year age bands with increasing increments of +2 points.}")

\item \textit{Unit Weighting}: This technique aims to produce a score by simply adding together all variables that are significantly correlated with the outcome of interest. Unit weighting is prevalent in criminal justice \citep[see, e.g.,][]{bobko2007usefulness,duwe2016sacrificing}, where it is referred to as the \textit{Burgess} method \citep[as it was first proposed by][]{burgess1928factors}. The use of this technique is frequently motivated by empirical work showing that linear models with unit weights may perform surprisingly well \cite[see, e.g.,][]{einhorn1975unit,dawes1979robust,holte1993very,Holte:2006fp,bobko2007usefulness}.%As an example, consider the work of  \citet{dawes1979robust}, where Dawes provides several examples of how classifiers where the coefficients are set intuitively as $\pm 1$ outperform models where the coefficients were ``obtained upon cross-validating... upon half the sample."

\end{itemize}

\subsection*{Critical Analysis of a Real-World Training Pipeline}

Many risk scores are built using sequential \textit{training pipelines} that combine traditional statistical techniques, heuristics, and expert judgement. The TIMI Risk Score of \citet{antman2000timi}, for example, was built as follows:
\begin{enumerate}[itemsep=6pt]\itshape
\item ``A total of 12 baseline characteristics arranged in a dichotomous fashion were screened as candidate predictor variables of risk of developing an end-point event"

\item ``After each factor was tested independently in a univariate logistic regression model, those that achieved a significance level of $p < .20$ were [retained]."

\item ``[The remaining factors]... selected for testing in a multivariate step-wise (backward elimination) logistic regression model. Variables associated with $p < .05$ were retained in the final model." 

\item ``After development of the multivariate model, the [risk predictions were determined]... for the test cohort using those variables that had been found to be statistically significant predictors of events in the multivariate analysis." 

\item ``The score was then constructed by a simple arithmetic sum of the number of variables present."
\end{enumerate}
Although this pipeline uses several established statistical techniques (e.g. stepwise regression, significance testing), it is unlikely to output a risk score that attains the best possible performance because:
\begin{itemize}

\item Decisions involving feature selection and rounding are made sequentially rather than globally (e.g., Steps 1-3 involve feature selection, Step 5 involves rounding).

\item The objective function that is optimized at each step differs from the global performance metric of interest.  

\item Some steps optimize conflicting objective functions (e.g., backward elimination typically optimizes the AIC or BIC, while the final model is fit to optimize the logistic loss). 

\item Some steps do not fully optimize their own objective function (i.e., backward elimination does not return a globally optimal feature set).

\item Some steps depend free parameters that are set without validation (e.g., the threshold significance level of $p < .20$ used in Step 2)

\item The final model is not trained using all the available training data. Here, Steps 4 and 5 use data from a ``test cohort" that could have been used to improve the fit of the final model. 

%\item The final model uses an empirical risk estimate for each score (i.e., the predicted risk for each score simply represents the fraction of patients in the test cohort with $y_i = +1$).

\item The coefficients of the final model are set to +1 and not optimized (that is, the model uses unit weights).
\end{itemize}

%% file: appendix_computational_experiments.tex
In this appendix, we provide additional details on the computational experiments in Sections \ref{Sec::MethodComparison} and \ref{Sec::AlgorithmicImprovements}. 

\subsection{Simulation Procedure for Synthetic Datasets}

We ran the computational experiments in Sections \ref{Sec::MethodComparison} and \ref{Sec::AlgorithmicImprovements} using a collection of synthetic datasets that we generated from the \textds{breastcancer} dataset of \citet[][]{mangasarian1995breast}, which contains $\n = 683$ samples and $d = 9$ features $x_{ij} \in \{0,\ldots, 10\}$. This choice was based off the fact that the \textds{breastcancer} dataset produces a \RSMINLP{} instance that can be solved using many algorithms, the data have been extensively studied in the literature, and can be obtained from the UCI ML repository \citep{bache2013uci}.  

We show the simulation procedure in Algorithm \ref{Alg::DataSimulation}. Given the original dataset, this procedure generates a collection of nested synthetic datasets in two steps. First, it generates the largest dataset (with $\Nmax{} = 5\times10^6$ and $\dmax{} = 30$) by replicating features and samples from the original dataset and adding normally distributed noise. Second, it produces smaller datasets by taking nested subsets of the samples and features. This ensures that a synthetic dataset with $d$ features and $\n$ samples contains the same features and examples as a smaller synthetic dataset (i.e., with $d' < d$ features and $\n' < \n$ samples). 

The resulting collection of synthetic datasets has two properties that are useful for benchmarking the performance of algorithms for \RSMINLP{}:
\begin{enumerate}[leftmargin=*,itemsep=0.5em]
\item They produce difficult instances of the risk score problem. Here, the \RSMINLP{} instances for synthetic datasets with $d > 9$ are challenging because the dataset contains replicates of the original 9 features. In particular, feature selection becomes exponentially harder when the dataset contains several copies of highly correlated features, as it means that there are an exponentially larger number of slightly suboptimal solutions.

\item They can be used to make inferences about the optimal objective value of \RSMINLP{} instances we may not be able to solve. Say, for example, that we could not solve an instance of the risk score problem for a synthetic dataset with $(d, \n) = (20, 10^6)$, but could solve an instance for a smaller synthetic dataset with $(d, \n{}) = (10, 10^6)$. In this case, we know that the optimal value of an \RSMINLP{} instance where $(d, \n) = (20, 10^6)$ must be less than or equal to the optimal value of an \RSMINLP{} instance where $(d, \n) = (10, 10^6)$. This is because the $(d, \n) = (20, 10^6)$ dataset contains all of the features as the $(d, \n) = (10, 10^6)$ dataset.
\end{enumerate}

\begin{algorithm}[htbp]
\caption{Simulation Procedure to Generate Nested Synthetic Datasets}
\label{Alg::DataSimulation}
\begin{algorithmic}[1]
\vspace{0.15em}
\small
\INPUT
\alginput{$X^\textrm{original} \gets [x_{ij}]_{i=1\ldots\Noriginal{}, j=1\ldots d^\text{original}}$}{feature matrix of original dataset}
\alginput{$Y^\textrm{original} \gets [y_i]_{i=1\ldots\Noriginal{}}$}{label vector of original dataset}
\alginput{$d^1\ldots \dmax$}{dimensions for synthetic datasets (increasing order)}
\alginput{$\n^1 \ldots \Nmax$}{sample sizes for synthetic datasets (increasing order)}
\algrule{0.5\textwidth}{0.5pt}  	
\INITIALIZE
\alginitialize{$\J^\text{original} \gets [1,\ldots, d^\text{original}]$}{index array for original features}
\alginitialize{$\J^{\max} \gets []$}{index array of features for largest synthetic dataset}
\alginitialize{$m^\text{full} \gets \lfloor \dmax / d^\text{original} \rfloor$}{}
\alginitialize{$m^\text{remainder} \gets \dmax - m^\text{full} \cdot d^\text{original}$}{}
\algrule{0.5\textwidth}{0.5pt}
\STEP{I}: Generate Largest Dataset
\vspace{0.5em} 

\For {$m{} = 1,\ldots, m^\textrm{full}$}
\State $\J^{\max} \leftarrow [\J^{\max}, \text{\textsf{RandomPermute}}(\J^{\text{\textrm{original}}})]$
\EndFor
    
\State  $\J^{\max} \leftarrow [\J^{\max}, \text{\textsf{RandomSampleWithoutReplacement}}(\J^{\text{original}}, m^\text{remainder})]$

  	  	\For {$i = 1,\ldots, \Nmax$}
  	  		    \State sample $l$ with replacement from $1,\ldots,\Noriginal$
				\State $y^{\max}_{i} \leftarrow y_i$  	  		  	  	  	
  	  	  	\For {$j = 1,\ldots, \dmax$}
	  	  	  	\State $k \leftarrow \J^{\max}[j]$
	  	  	  	\State sample $\varepsilon$ from $\text{Normal}(0, 0.5)$
	  	  	  	\State $x^{\max}_{ij} \leftarrow \round{x_{l,k} + \varepsilon}$ \Comment{new features are noisy versions of original features}
	  	  	  	\State $x^{\max}_{ij} \leftarrow \min(10,\max(0, x^{\max}_{ij}))$ \Comment{new features have same bounds as old features}
		\EndFor
	\EndFor
	\State $X^{\max} \leftarrow [x^{\max}_{ij}]_{i=1\ldots\Nmax, j=1\ldots \dmax}$
	\State $Y^{\max} \leftarrow [y^{max}_i]_{i=1\ldots\Nmax}$	
\vspace{0.5em}
\STEP{II}: Generate Smaller Datasets
\vspace{0.5em} 
    
		\For {$d = [d^1, \ldots, \dmax]$}
      	  	    \For {$\n{} = [\n^1, \ldots, \Nmax]$}
      	  	    \State      	  	    $X^{(\n{},d)} = X^{\max}[1:\n,1:d]$
      	  	    \State      	  	    $Y^{\n} = Y^{\max}[1:\n]$
      	  	    \EndFor
		\EndFor

  \Ensure synthetic datasets $(X^{(\n,d)},Y^{\n})$ for all $\n^1 \ldots \Nmax$ and $d^1\ldots \dmax$.
  \end{algorithmic}
\end{algorithm}

\clearpage
\subsection{Setup on the Performance Comparison in Section \ref{Sec::MethodComparison}}
\label{Appendix::PerformanceComparisonSetup}

We consider an instance of \RSMINLP{} where $C_0 = 10^{-8}$, $\lambda_0 \in \{-100, 100\}$, and $\lambda_j = \{-10,\ldots,10\}$ for $j = 1,\ldots,d$.  We solved this instance on a 3.33 GHz Intel Xeon CPU with 16GB of RAM for up to 6 hours using the following algorithms: 
\begin{romanlist}
\item \CPA{}, as described in Algorithm \ref{Alg::CPA}; 
\item \LCPA{}, as described in Algorithm \ref{Alg::LCPA};
\item \textsf{ActiveSetMINLP}, an active set MINLP algorithm;
\item \textsf{InteriorMINLP}, an interior point MINLP algorithm;
\item \textsf{InteriorCGMINLP}, an interior point MINLP algorithm where the primal-dual KKT system is solved with a conjugate gradient method;
\end{romanlist}
All MINLP algorithms were implemented in a state-of-the-art commercial solver (i.e., Artelsys Knitro 9.0, which is an updated version of the solver described in \citealt{byrd2006knitro}). If an algorithm did not return a certifiably optimal solution for a particular instance within the 6 hour time limit, we reported results for the best feasible solution. Both \CPA{} and \LCPA{} were implemented using CPLEX 12.6.3. 

%\subsubsection*{\rsmip{} Formulation for \CPA{}}

Our \CPA{} implementation uses the solves the following formulation of \rsmip{}.

\begin{definition}[\rsmip{}]
\label{Def::RSMIP}
Given a finite coefficient set $\Lset \subset \Z^{d+1}$, trade-off parameter $C_0 > 0$, and cutting plane approximation $\cutloss{k}: \R^{d+1} \to \R_+$ with cut parameters $\{\lossfun{\lambdab^t},\lossgrad{\lambdab^t}\}_{t=1}^k$, the surrogate optimization problem $\RSMIP{\cutloss{k}}$ can be formulated as the mixed integer program:
{%
\small%\setstretch{1.25}
\begin{subequations}
\label{Formulation::RSMIP}
\begin{equationarray}{@{}crcl>{\qquad}l>{\quad}r@{}}
\min_{\lossval{},\lambdab,\bf{\alpha}} & V & & \notag \\
\st  & V & = &  \lossval{} + C_0 R    & & \mipwhat{objective value} \label{Con::RSMIPObjVar} \\
& \lossval{} & \geq & \lossfun{\lambdab^t} + \dotprod{\lossgrad{\lambdab^t}}{\lambdab{} - \lambdab^t} & \miprange{t}{1}{k} & \mipwhat{cut constraints} \label{Con::RSMIPCuts} \\
& R & = & \sum_{j=1}^d \alpha_j & & \mipwhat{$\ell_0$--norm} \label{Con::RSMIPL0Var} \\
&  \lambda_j  & \leq &  \maxlambda{j} \alpha_j  & \miprange{j}{1}{d} & \mipwhat{$\ell_0$ indicator constraints} \label{Con::RSMIPL0UB}  \\
&  \lambda_j  & \geq & - \minlambda{j} \alpha_j  & \miprange{j}{1}{d} & \mipwhat{$\ell_0$ indicator constraints} \label{Con::RSMIPL0LB} \\
& V  & \in & [\objvalmin{}, \objvalmax{}] & & \mipwhat{bounds on objective value} \label{Con::RSMIPObjBds} \\ 
& \lossval{} & \in & [\lossmin{}, \lossmax{}] &  & \mipwhat{bounds on loss value} \label{Con::RSMIPLossBds} \\
& R & \in & \{\zeronormmin, \ldots, \zeronormmax\} & & \mipwhat{bounds on $\ell_0$--norm} \label{Con::RSMIPL0Bds}  \\ 
&  \lambda_j  & \in & \{\minlambda{j}, \ldots, \maxlambda{j}\} & \miprange{j}{1}{d} & \mipwhat{coefficient bounds}\label{Con::RSMIPCoefBds} \\ 
& \alpha_j  & \in & \{0,1\}  & \miprange{j}{1}{d} & \mipwhat{$\ell_0$ indicator variables}\label{Con::RSMIPL0VarBds} \notag%
\end{equationarray}
\end{subequations}
}
\end{definition}
The formulation in Definition \ref{Def::RSMIP} contains $2d + 3$ variables and $k + 2d + 2$ constraints (excluding bounds on the variables). Here, the cutting plane approximation is represented via the cut constraints in \eqref{Con::RSMIPCuts} and the auxiliary variable $\lossval{} \in \R_{+}$. The $\ell_0$--norm is computed using the indicator variables $\alpha_j = \indic{\lambda_j \neq 0}$ set in constraints \eqref{Con::RSMIPL0UB} and \eqref{Con::RSMIPL0LB}. The $\lambda_j$ are restricted to a bounded set of integers in constraints \eqref{Con::RSMIPCoefBds}. The formulation includes two additional auxiliary variables: $V$, defined as the objective value in \eqref{Con::RSMIPObjVar}; and $R$, defined as the $\ell_0$-norm in \eqref{Con::RSMIPL0Var}. %These variables will be useful for implementing the techniques in Section \ref{Sec::AlgorithmicImprovements}. In particular, by including $V$ and $R$, we can use a control callback in a MIP solver to set bounds on these quantities during branch-and-bound without adding new constraints to the formulation.

\subsection{Results for MINLP Algorithms}
\label{Appendix::PerformanceComparison}

\begin{figure}[htbp]
\centering\setlength{\tabcolsep}{0pt}\renewcommand{\arraystretch}{0.0}
\begin{tabular}{>{\;}l>{\quad}c>{\quad}c@{}c@{}c@{}}

\toprule

& &  
\cell{c}{\scriptsize\textsf{ActiveSetMINLP}} & 
\cell{c}{\scriptsize\textsf{InteriorMINLP}}  & 
\cell{c}{\scriptsize\textsf{InteriorCGMINLP}}\\

\midrule

{\itshape\scriptsize\renewcommand{\arraystretch}{1.05}
\begin{tabularx}{0.25\textwidth}{X}
{\bf{Time to Train a Good Risk Score}}\\[0.25em]
i.e., the time for an algorithm to find a solution whose loss $\leq10\%$ of the optimal loss. It reflects the time to obtain a risk score with good calibration without a proof of optimality.
\end{tabularx}
} & 
\cell{c}{\includenewheatlegend{algorithm_comparison_good_time_legend.pdf}} &
\cell{c}{\includenewheatmapNOYAXIS{algorithm_comparison_good_time_MINLP_3.pdf}} &
\cell{c}{\includenewheatmapNOYAXIS{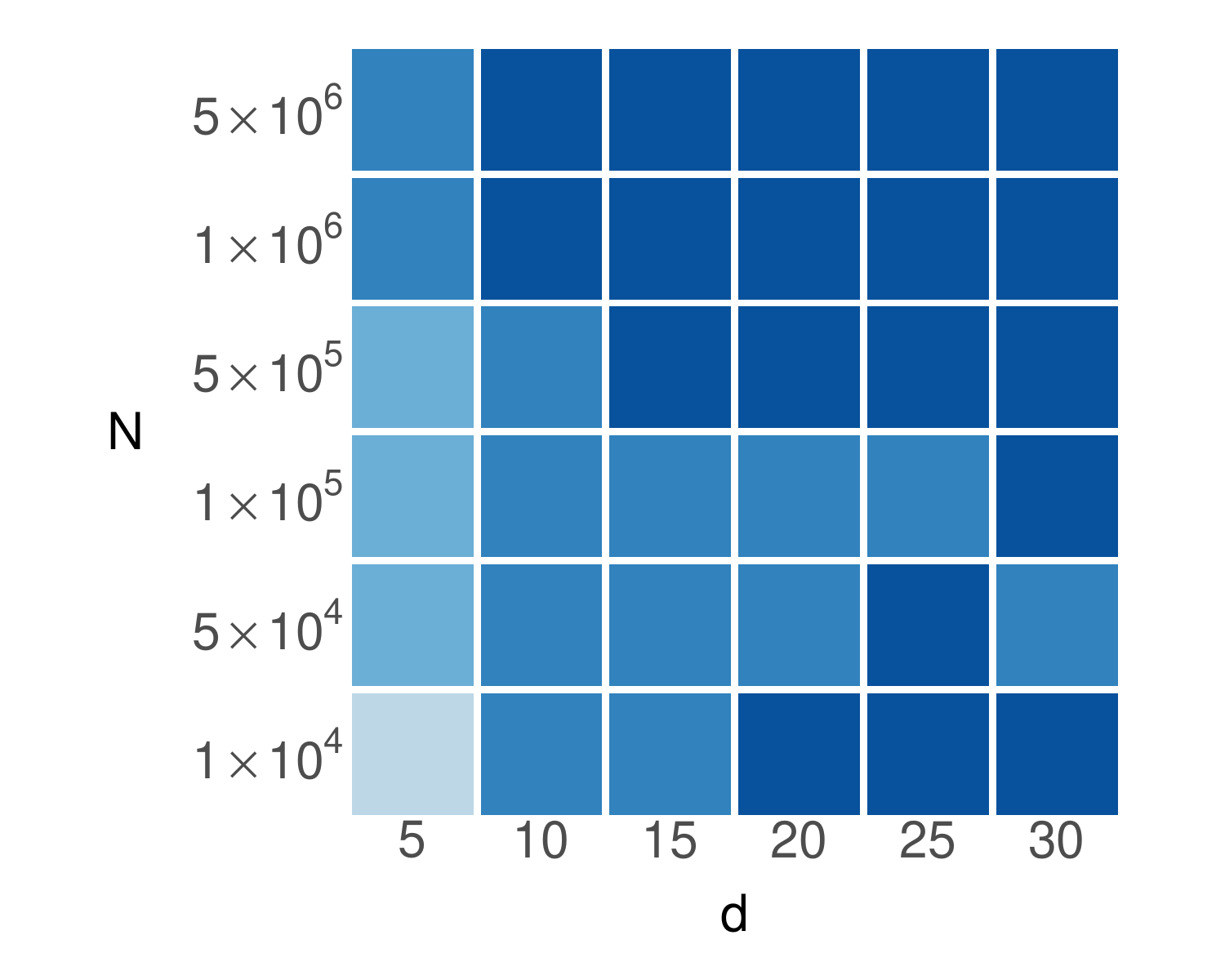}} & 
\cell{c}{\includenewheatmapNOYAXIS{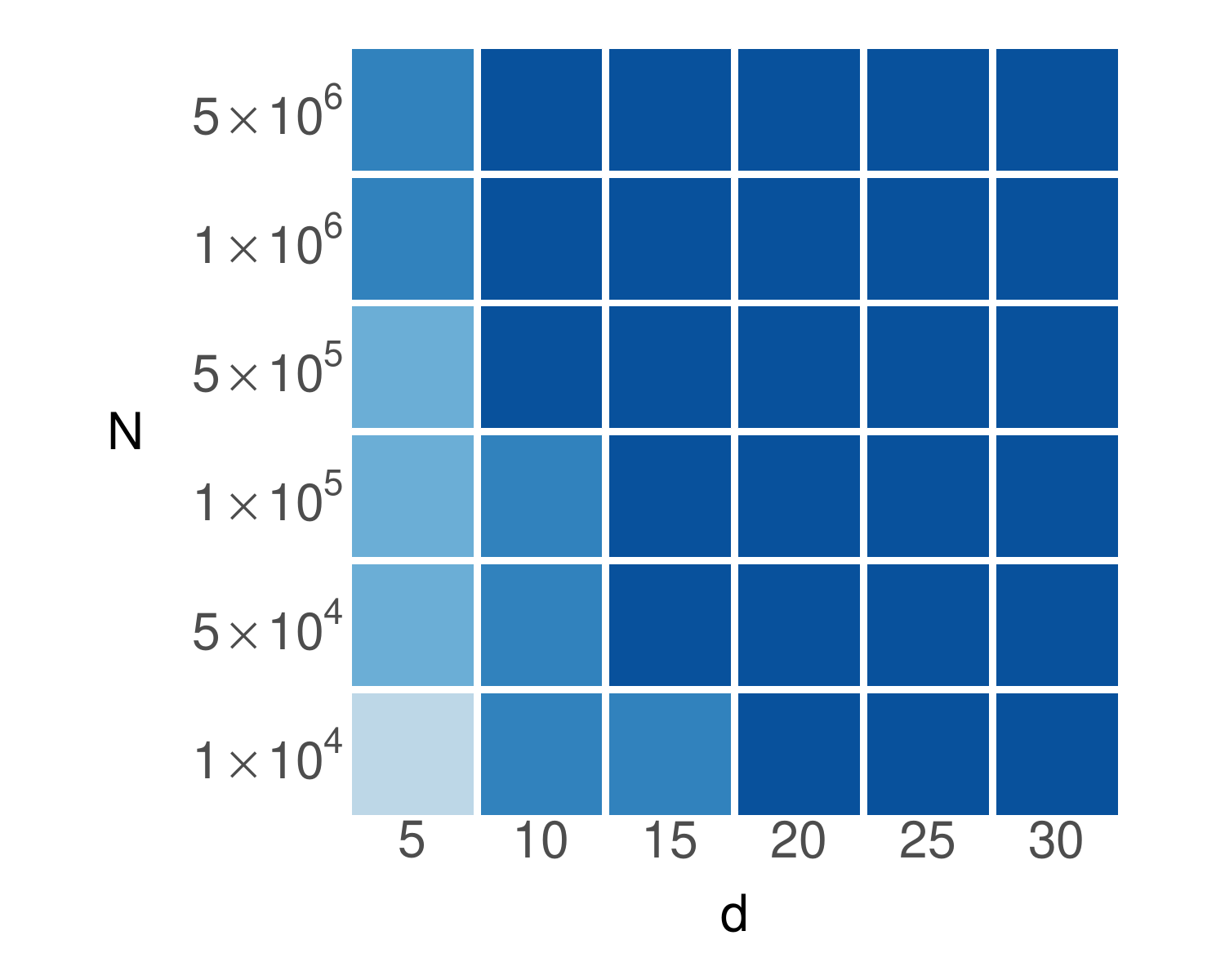}} \\

\midrule

{\renewcommand{\arraystretch}{1.05}%
\itshape\scriptsize
\begin{tabularx}{0.25\textwidth}{X}
\bf{Optimality Gap of Best}\\
\bf{Solution at Termination}\\[0.25em]
i.e., $(\objvalmax{}-\objvalmin{})/\objvalmax{}$, where $\objvalmax{}$ is the objective value of the best solution found at termination. A gap of 0.0\% means an algorithm has found the optimal solution and provided a proof of optimality within 6 hours.
\end{tabularx}} & 
\cell{c}{\includenewheatlegend{algorithm_comparison_relgap_legend.pdf}} &
\cell{c}{\includenewheatmapNOYAXIS{algorithm_comparison_relgap_MINLP_3.pdf}} &
\cell{c}{\includenewheatmapNOYAXIS{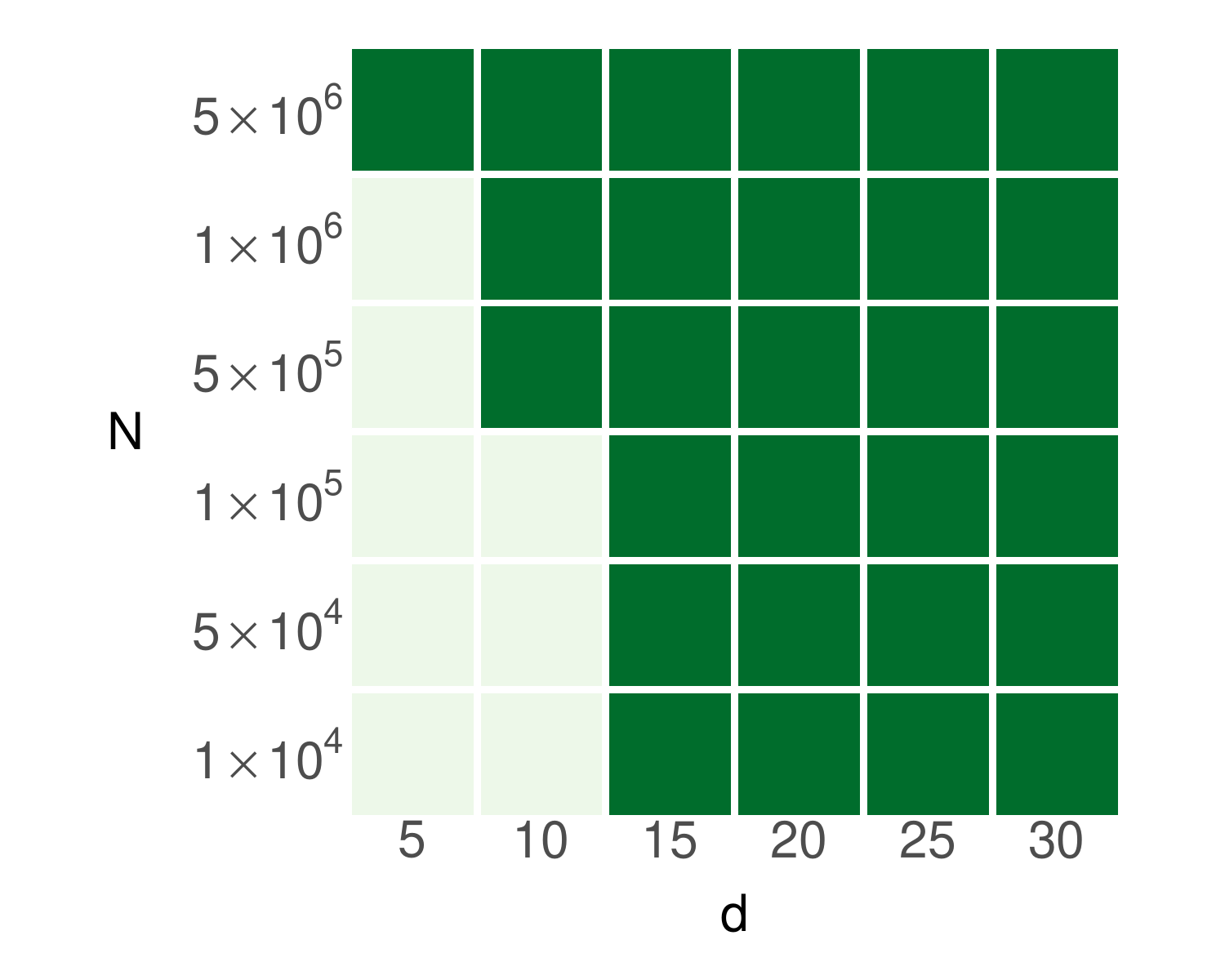}} & 
\cell{c}{\includenewheatmapNOYAXIS{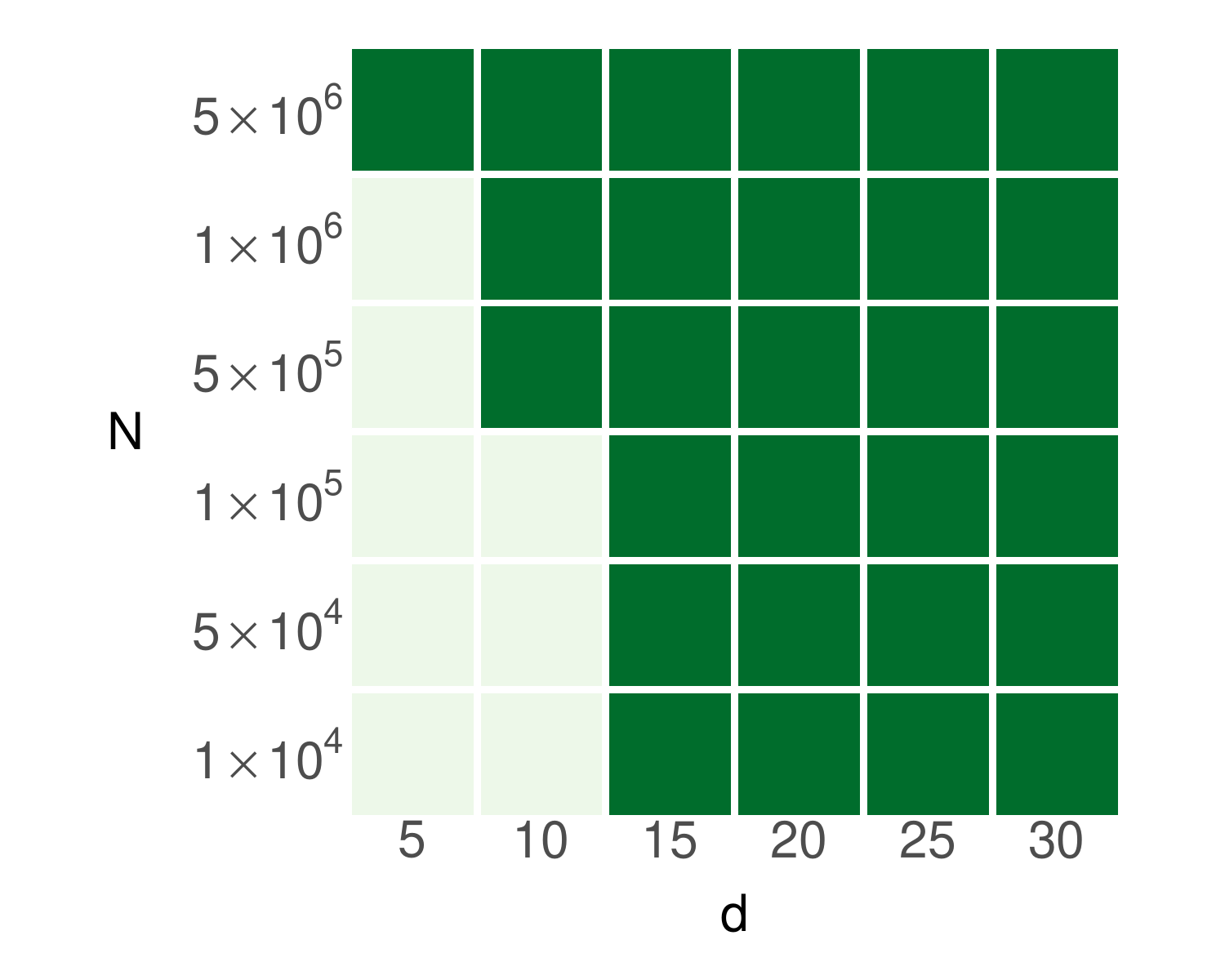}} \\

\midrule

{\renewcommand{\arraystretch}{1.05}%
\itshape\scriptsize
\begin{tabularx}{0.25\textwidth}{X}
\bf{\% Time Spent on Data-}\\
\bf{Related Computation}\\[0.25em]
i.e., the proportion of total runtime that an algorithm spends computing the value, gradient, or Hessian of the loss function.
\end{tabularx}} & 
\cell{c}{\includenewheatlegend{algorithm_comparison_data_time_legend.pdf}} &
\cell{c}{\includenewheatmapNOYAXIS{algorithm_comparison_data_time_MINLP_3.pdf}} &
\cell{c}{\includenewheatmapNOYAXIS{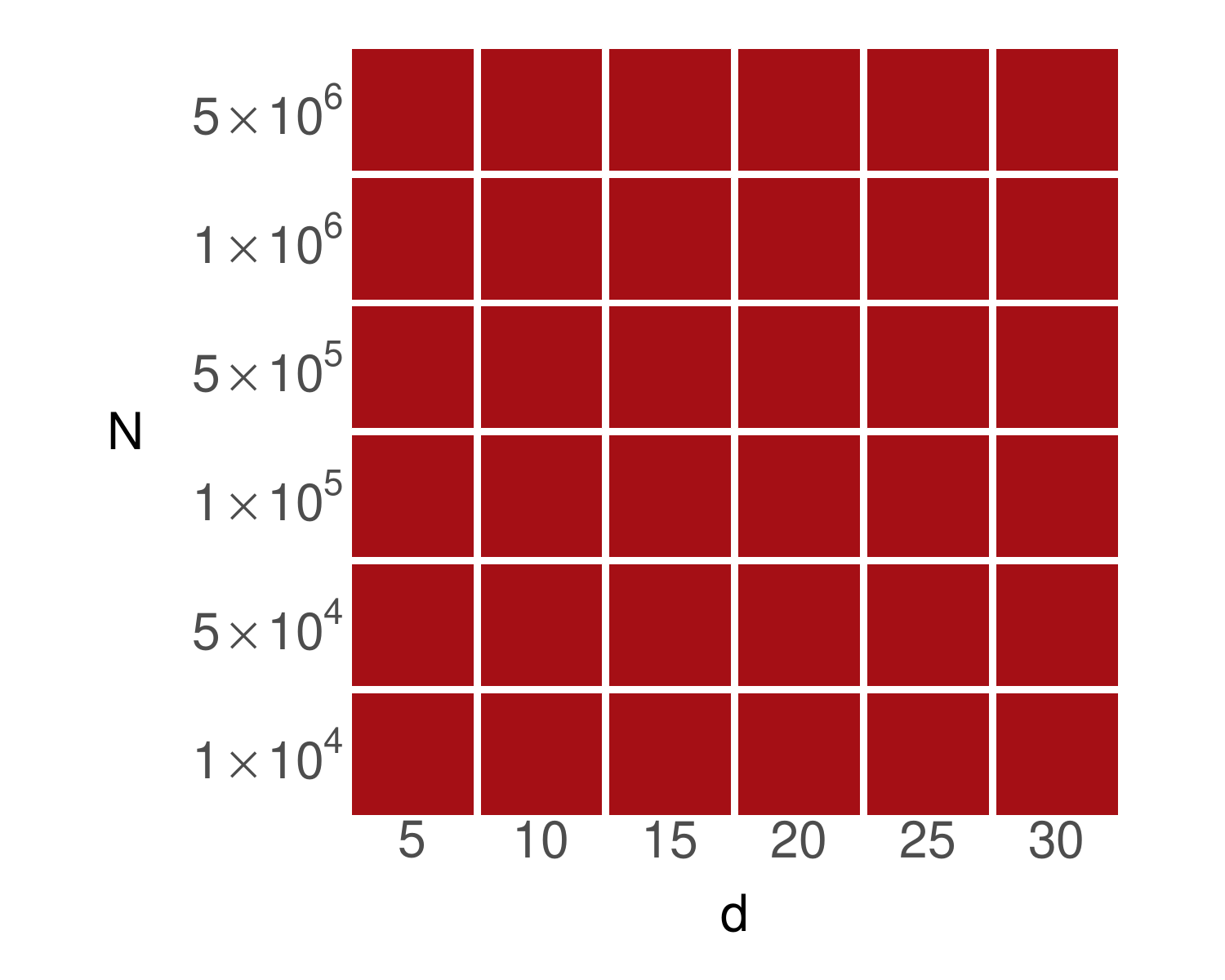}} & 
\cell{c}{\includenewheatmapNOYAXIS{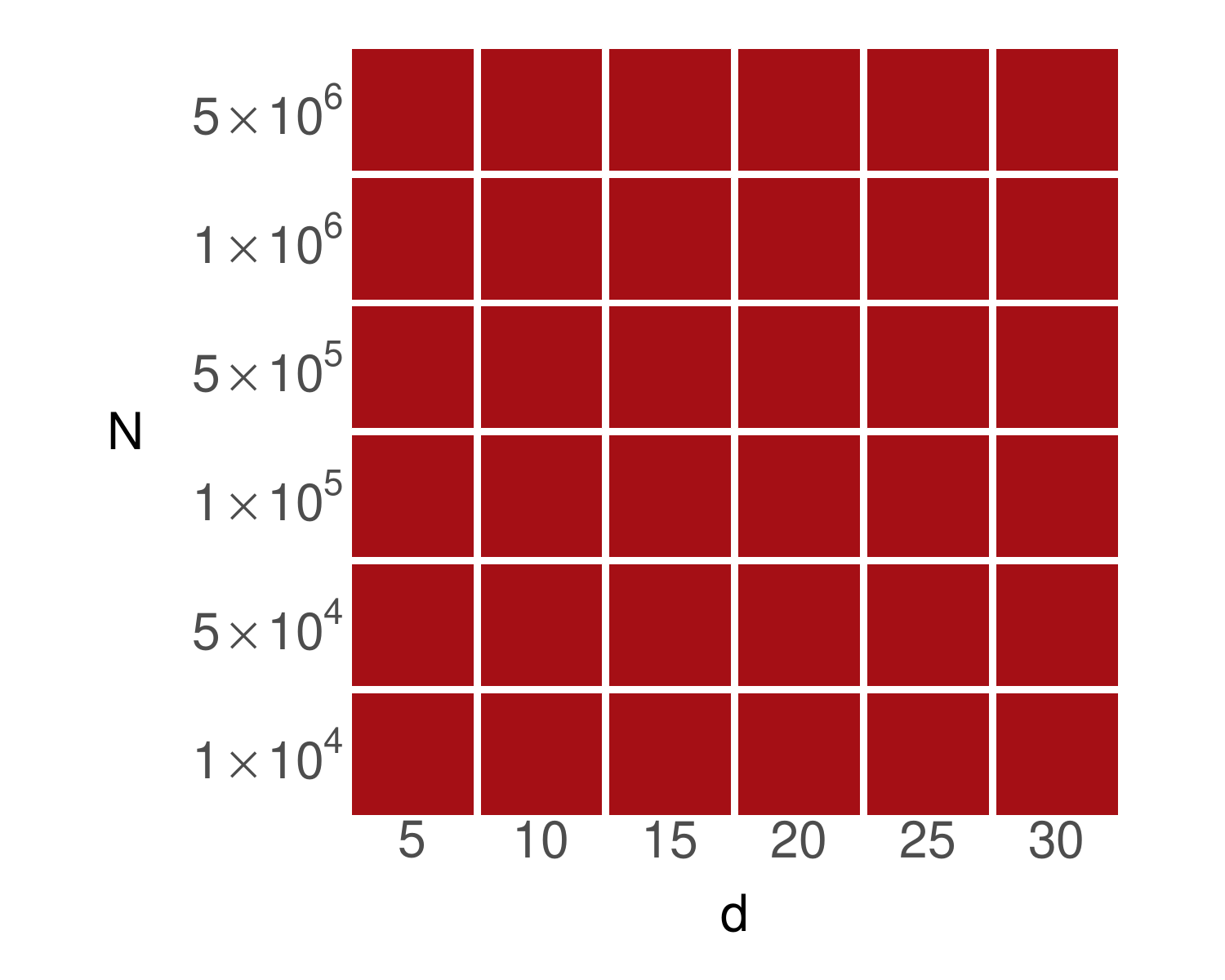}} \\

\bottomrule
\end{tabular}
\caption{Performance of MINLP algorithms on difficult instances of \RSMINLP{} for synthetic datasets with varying dimensions $d$ and sample sizes $\n$. All algorithms perform similarly. We report results for \textsf{ActiveSetMINLP} in Section \ref{Sec::MethodComparison} because it solves the most instances to optimality.}
\label{Fig::MethodComparisonMINLPOnly}
\end{figure}

In Figure \ref{Fig::MethodComparisonMINLPOnly}, we plot results for all MINLP algorithms that we benchmarked against \CPA{} and \LCPA{}: \textsf{ActiveSetMINLP}, \textsf{InteriorMINLP}, and \textsf{InteriorCGMINLP}. We reported results for \textsf{ActiveSetMINLP} in Section \ref{Sec::MethodComparison} because it solved the largest number of instances to optimality.

%% file: appendix_algorithmic_improvements.tex
\subsection{Initialization Procedure}
\label{Sec::Initialization}

In Algorithm \ref{Alg::InitializationProcedure}, we present an initialization procedure for \LCPA{}. The procedure aims to speed up \LCPA{} by generating: a collection of cutting planes; a good integer solution; and non-trivial bounds on the values of the objective, loss, and number of non-zero coefficients. It combines all of the techniques from this section, as follows:
\begin{enumerate}[leftmargin=*]
\item \textit{Run \textnormal{\CPA{}} using \rslp{}}: We apply traditional \CPA{} to produce a cutting-plane approximation of the loss function using a convex relaxation of \RSMINLP{}. We can achieve this by replacing $\RSMIP{k}$ in Step \ref{AlgStep::CPASolveMIP} of \CPA{} with 
$\RSLP{k}{\conv{\Lset}}$, and running \CPA{} 
until a user-specified stopping condition is met. We store: (i) the cuts from \rslp{} (which will be used as an initial cutting plane approximation for \LCPA{}); (ii) the lower bound from \CPA{} on the objective value of \rslp{} (which represents a lower bound $\objvalmin$ on the optimal value of \RSMINLP{}).

\item \textit{Sequential Rounding and Polishing}: We collect the solutions produced at each iteration of \CPA{}. For each solution, we run \SR{} to obtain an integer solution for $\RSMINLP{}$, and then polish it using \DCD{}. We use the best solution found so far to update the upper bound $\objvalmax$ on the optimal value to \RSMINLP{}. 

\item \textit{Chained Updates}: Having obtained non-trivial bounds on $\objvalmin$ and $\objvalmax$, we update the bounds on key quantities using \ChainedUpdates{}.
\end{enumerate}
In Figure \ref{Fig::WarmStartPerformanceProfile}, we show how the initialization procedure in Algorithm \ref{Alg::InitializationProcedure} improves the lower bound and the optimality gap over the course of \LCPA{}.

\begin{algorithm}
\caption{Initialization Procedure for \LCPA{}}
\label{Alg::InitializationProcedure}
\begin{algorithmic}[1]\small
\vspace{0.15em}
\INPUT
\alginput{$(\xb_i, y_i)_{i=1}^\n$}{training data}
\alginput{$\Lset$}{coefficient set}
\alginput{$C_0$}{$\ell_0$ penalty parameter}
\alginitialize{$\objvalmin$, $\objvalmax$, $\lossmin$, $\lossmax$, $\zeronormmin$, $\zeronormmax$}{initial bounds on $\objval{\lambdab^\opt}$, $\lossfun{\lambdab^\opt}$ and $\zeronorm{\lambdab^\opt}$}
\alginput{$T^\textrm{max}$}{time limit for \CPA{} on \rslp{}}
\vspace{-0.5em}\algrule{0.5\textwidth}{0.5pt}\vspace{-0.5em}
\INITIALIZE
\alginitialize{$\cutlossfun{0}{\lambdab} \gets \left\{0\right\}$}{initial approximation of loss function}
 \alginitialize{$\mathcal{P}^\textrm{int} \gets \emptyset$}{initial collection of integer solutions}
\vspace{1em}
\STEP{I:} \textbf{Run}  \CPA{} \textbf{using subroutine} \rslp{} 
\vspace{1em}
\State Run \CPA{} \Comment{Algorithm \ref{Alg::CPA} using $\RSLP{\cutloss{0}}{\conv{\Lset}}$}
\State $k \gets$ number of cuts added by \CPA{} within time limit $T^\textrm{max}$
\State $\cutloss{\textrm{initial}} \gets \cutloss{k}$ \Comment{store cuts from each iteration}
\State $\mathcal{P}^\textrm{real} \gets \{\lambdab^t\}_{t=1}^k$ \Comment{store solutions from each iteration}
\State  $\objvalmin \gets$ lower bound from \CPA{} \Comment{lower bound for \rslp{} is lower bound for \rsminlp{}}
\vspace{0.75em}
\STEP{II:} \textbf{Round and Polish Non-Integer Solutions from} \CPA{}
\vspace{0.75em}
\For{\textbf{each} $\lambdab^\textrm{real} \in \mathcal{P}^\textrm{real}$}
\State $\lambdab^\textrm{sr} \gets \SR\left(\lambdab^\textrm{real}, \Lset, C_0\right)$ \Comment{Algorithm \ref{Alg::SequentialRounding}}
\State $\lambdab^\textrm{dcd} \gets \DCD\left(\lambdab^\textrm{sr},\Lset, C_0\right)$ \Comment{Algorithm \ref{Alg::DCD}}
\State $\mathcal{P}^\textrm{int} \gets \mathcal{P}^\textrm{int} \cup \{ \lambdab^\textrm{dcd} \}$ \Comment{store polished integer solutions}
\EndFor
\State $\lambdab^\textrm{best} \gets \argmin_{\lambdab \in \mathcal{P}^\textrm{int}} \objval{\lambdab}$
\State $\objvalmax \gets \objval{\lambdab^\textrm{best}}$  \Comment{best integer solution produces upper bound for $\objval{\lambdab^\opt}$}
\vspace{0.75em}
\STEP{III: Update Bounds on Objective Terms}
\vspace{0.75em}
\State $(\objvalmin,\ldots,\zeronormmax) \gets \ChainedUpdates\left(\objvalmin,\ldots,\zeronormmax,C_0\right)$ \Comment{Algorithm \ref{Alg::ChainedUpdates}}
\vspace{0.75em}
\Ensure $\lambdab^\textrm{best}$,  $\cutlossfun{\textrm{initial}}{\lambdab}$, $\objvalmin$, $\objvalmax$, $\lossmin$, $\lossmax$, $\zeronormmin$, $\zeronormmax$
\vspace{0.25em}
\end{algorithmic}
\end{algorithm}
%
%There are several options for how to stop \CPA{} in Algorithm \label{Alg::InitializationProcedure}. In addition to imposing time limits on the total runtime, we note two other stopping strategies:
%%
%\begin{itemize} 
%\item \textit{Stopping Rule for Best Initial Lower Bound}: We can stop by specifying a threshold $\varepsilon \geq 0$ on the relative optimality gap for \CPA{}. This stopping rule aims to run \CPA{} until the lower bound from the initialization procedure is within $\varepsilon$ of the best lower bound that can be achieved by solving \rslp{}. As per Theorem 2 in \citet{teo2007scalable}, \CPA{} will produce an $\varepsilon$-optimal solution after at most $K \leq \log_2{C_0 L} - 2 \log_2 M + 8 \frac{M^2}{C_0\epsilon}$ iterations where $L = \lossfun{\mathbf{0}}$ and $M = \max_{\lambdab \in \Lset} \lossgrad{\lambdab}.$
%%
%\item \textit{Stopping Rule for Best Initial Upper Bound}: We can stop \CPA{} once the non-integer solutions from each iteration are rounded in the same way. Let $K := \min t ~\st~ \lambdab^t = \round{\lambdab^t}$, then note that $\round{\lambdab^{k}} = \round{\lambdab^{k+m}}$ for all $m \in \N$, and that $\SR(\lambdab^k) = \SR(\lambdab^{K+m})$ for all $m \in\N$. Since the output from sequential rounding and polishing is ultimately return a solution $\lambdab^\textrm{best}$ that is used to set the upper bound $\objvalmax$, this stopping condition stops \CPA{} as soon as the non-integer solution will no longer improve the upper bound obtained by the initialization procedure.
%\end{itemize}

\begin{figure}[htbp]
\centering
%\begin{tabular}{c}
%\setlength{\tabcolsep}{0pt}
%\includeAlgLegend{WARMSTART}{legend.pdf} \\ 
\begin{tabular}{@{}lr@{}}
%\includeAlgPlot{WARMSTART}{lb_vs_time.pdf} & \includeAlgPlot{WARMSTART}{gap_vs_time.pdf} 
\includeAlgPlot{WARMSTART}{lb_vs_nodes.pdf} & \includeAlgPlot{WARMSTART}{gap_vs_nodes.pdf} 
\end{tabular}
%\end{tabular}
\caption{Performance profile of \LCPA{} in a basic implementation (black) and with the initialization procedure in Algorithm \ref{Alg::InitializationProcedure} (red). Results reflect performance on a \RSMINLP{} instance for a synthetic dataset with $d = 30$ and $\n =$ 50,000 (see Appendix \ref{Appendix::SimulationDetails} for details).}
\label{Fig::WarmStartPerformanceProfile}
\end{figure}

\subsection{Reducing Data-Related Computation}
\label{Sec::DataRelatedComputation}
In this section, we present techniques to reduce data-related computation for the risk score problem. 
\subsubsection{Fast Loss Evaluation via a Lookup Table}
\label{Sec::FastLossEvaluation}

The first technique is designed to speed up the evaluation of the loss function and its gradient, which reduces runtime when we compute cut parameters \eqref{Eq::CuttingPlaneParameters} and call the rounding and polishing procedures in Section \ref{Sec::AlgorithmicImprovements}. The technique requires that the features $\xb_i$ and coefficients $\lambdab$ belong to sets that are bounded, discrete, and regularly spaced, such as $\xb_i \in \X \subseteq \{0,1\}^d$ and $\lambdab \in \Lset \subseteq \{-10,\ldots,10\}^{d+1}$. 

Evaluating the logistic loss, $\log(1 + \exp(-\dotprod{\lambdab}{y_i\xb_i})$, is a relatively expensive operation because it involves exponentiation and must be carried out in multiple steps to avoid numerical overflow/underflow when the \textit{scores} $s_i  = \dotprod{\lambdab}{\xb_i y_i}$ are too small or large%
\footnote{The value of $\exp(s)$ can be computed reliably using IEEE 754 double precision floating point numbers for $s \in [-700,700]$. The term will overflow to $\infty$ when $s < -700$, and underflow to $0$ when when $s > 700$.}. When the training data and coefficients belong to discrete bounded sets,  the scores $s_i = \dotprod{\lambdab}{\xb_iy_i}$ belong to a discrete and bounded set $$\Sset = \left \{\dotprod{\lambdab}{\xb_iy_i} ~ \big | ~  i = 1,\ldots,\n \text{ and } \lambdab \in \Lset\right\}.$$ If the elements of the feature set $\X$ and the coefficient set $\Lset$ are regularly spaced, then the scores belong to the set of integers $\Sset \subseteq \Z \cap [s^{\min}, s^{\max}]$ where:
\begin{align*}
s^{\min} &= \min_{i, \lambdab} \left\{ \dotprod{\lambdab}{\xb_i y_i} \text{ for all } (\xb_i, y_i) \in \data \text{ and } \lambdab \in \Lset \right\},\\ 
s^{\max} &= \max_{i, \lambdab} \left\{ \dotprod{\lambdab}{\xb_i y_i} \text{ for all } (\xb_i, y_i) \in \data \text{ and } \lambdab \in \Lset \right\}.
\end{align*}
Thus, we can precompute and store all possible values of the loss function in a lookup table with $s^{\max} - s^{\min} + 1$ rows, where row $m$ contains the value of $[\log(1+\exp(-(m + s^{\min} -1)))]$.

This strategy can reduce the time to evaluate the loss as it replaces a computationally expensive operation with a fast lookup. In practice, the lookup table is small enough to be cached in memory, which yields a substantial speedup. In addition, when $\zeronormmax$ is updated over the course of \LCPA{}, the lookup table can be further reduced by recomputing $s^{\min}$ and $s^{\max}$, and limiting the entries to values between $s^{\min}$ and $s^{\max}$. The values of $s^{\min}$ and $s^{\max}$ can be computed in O($\n$) time, so the update is not expensive.

\subsubsection{Faster Rounding Heuristics via Subsampling}
\label{Sec::Subsampling}

We now describe a subsampling technique to reduce computation for rounding heuristics that require multiple evaluations of the loss function (e.g., \SR{}).

Ideally, we would want to run such heuristics frequently as possible because each run may output a solution that updates the incumbent solution (i.e., the current best solution to the risk score problem). In practice, however, this may slow down \LCPA{} since each run requires multiple evaluations of the loss, and runs that fail to update the incumbent amount to wasted computation. If, for example, we ran \SR{} each time the MIP solver found a set of non-integer coefficients in \LCPA{} (i.e., Step \ref{AlgStep::LCPANonIntegerSolution} of Algorithm \ref{Alg::LCPA}), many rounded solutions would not update the incumbent, and we would have wasted too much time rounding, without necessarily finding a better solution. 

Our technique aims to reduce the overhead of calling heuristics by running them on a smaller dataset $\data_\m{}$ built by sampling $\m$ points without replacement from the training dataset $\data_\n{}$. In what follows, we present probabilistic guarantees to choose the number of samples $m$ so that an incumbent update using $\data_\m{}$ guarantees an incumbent update using $\data_n{}$. To clarify when the loss and objective are computed using $\data_\m{}$ or $\data_\n{}$, we let $\samplelossfun{i}{\lambdab} = \log(1+\exp(\dotprod{\lambdab}{y_i\xb_i}))$ and define:
\begin{align*}
\samplelossfun{\m}{\lambdab} &= \frac{1}{\m}\sum_{i=1}^\m \samplelossfun{i}{\lambdab}, &\sampleobjval{\m}{\lambdab} &= \samplelossfun{\m}{\lambdab} + C_0 \zeronorm{\lambdab},\\
\samplelossfun{\n}{\lambdab} &= \frac{1}{\n}\sum_{i=1}^\n \samplelossfun{i}{\lambdab}, &\sampleobjval{\n}{\lambdab} &= \samplelossfun{\n}{\lambdab} + C_0 \zeronorm{\lambdab}.
\end{align*}
Consider a case where a heuristic returns a promising solution $\lambdab^\textrm{hr}$ such that:
\begin{align}
V_\m{}(\lambdab^\textrm{hr}) < \objvalmax \label{Eq::IncumbentUpdateSubsample}.
\end{align}
In this case, we compute the objective on the full training dataset $\data_\n{}$ by evaluating the loss for each of the $\n - \m$ points that were not included in $\data_\m{}$.  We then update the incumbent solution if $\lambdab^\textrm{hr}$ attains an objective value that is less than the current upper bound:
\begin{align}
V_\n{}(\lambdab^\textrm{hr}) < \objvalmax \label{Eq::IncumbentUpdateFull}.
\end{align}
Although this strategy requires evaluating the loss for the full training dataset to validate an incumbent update, it still reduces data-related computation since rounding heuristics typically require multiple evaluations of the loss (e.g., \SR{}, which requires $\frac{d}{3}(d^2+3d+2)$ evaluations). %In the interest of reducing computation, our strategy ignores ``false negative" solutions $\lambdab^\textrm{hr}$ that do poorly on the heuristics dataset $V_\m{}(\lambdab^\textrm{hr}) \geq \objvalmax$ but may update the incumbent on the true dataset $V_\n{}(\lambdab^\textrm{hr}) < \objvalmax$.

To guarantee that any solution that updates the incumbent when the objective is evaluated with $\data_\m{}$ will also update the incumbent when the objective is evaluated with $\data_\n{}$ (i.e., that any solution that satisfies \eqref{Eq::IncumbentUpdateSubsample} will also satisfy \eqref{Eq::IncumbentUpdateFull}), we can use the generalization bound from Theorem \ref{Thm::SubsamplingGeneralizationBound}.
\begin{theorem}[Generalization of Sampled Loss on Finite Coefficient Set]
\label{Thm::SubsamplingGeneralizationBound}
Let $\data_\n{} = (\xb_i,y_i)_{i=1}^\n$ denote a training dataset with $\n{} > 1$ points, $\data_\m{} = (\xb_i,y_i)_{i=1}^\m$  denote a sample of $\m$ points drawn without replacement from $\data_\n{}$, and $\lambdab$ denote the coefficients of a linear classifier from a finite set $\mathcal{\Lset}$. For all $\varepsilon > 0$, it holds that
\begin{align*}
\prob{ \max_{\lambdab \in \Lset} \Big( \samplelossfun{\n}{\lambdab} -  \samplelossfun{\m}{\lambdab} \Big) \geq \varepsilon} &%
\leq |\Lset| \exp \left(  -\frac{2 \varepsilon^2}{ (\frac{1}{\m}) (1 - \frac{\m^2}{\n^2})\Delta^{\max}(\Lset,\data_\n{})^2}  \right),
\end{align*}
where $\Delta^{\max}(\Lset,\data_\n{}) = \max_{\lambdab \in \Lset} ~ \left( \max_{i=1,\ldots,\n} \samplelossfun{i}{\lambdab} - \min_{i=1,\ldots,\n} \samplelossfun{i}{\lambdab} \right).$
\end{theorem}

\clearpage
\begin{proof}[Theorem \ref{Thm::SubsamplingGeneralizationBound}]
For a fixed set coefficient vector $\lambdab \in \Lset$, consider a sample of $\n$ points composed of the values for the loss function $\samplelossfun{i}{\lambdab}$ for each example in the full training dataset $\data_\n{} = (\xb_i,y_i)_{i=1}^\n$. Let $\samplelossfun{\n}{\lambdab} = \frac{1}{\n}\sum_{i=1}^\n \samplelossfun{i}{\lambdab}$ and $\samplelossfun{\m}{\lambdab} = \frac{1}{\m}\sum_{i=1}^\m \samplelossfun{i}{\lambdab}$. Then, the Hoeffding-Serfling inequality \citep[see e.g., Theorem 2.4 in][]{bardenet2015concentration} guarantees the following for all $\varepsilon > 0$:
\begin{align*}
\prob{ \samplelossfun{\n}{\lambdab} - \samplelossfun{\m}{\lambdab} \geq \varepsilon} &\leq \exp \left(  -\frac{2 \varepsilon^2}{
 (\frac{1}{\m{}}) (1 - \frac{\m}{\n}) (1 + \frac{\m}{\n}) \Delta(\lambdab,\data_\n{})^2}  \right),\\
 \intertext{where}
\Delta(\lambdab,\data_\n{}) &= \max_{i=1,\ldots,\n} \samplelossfun{i}{\lambdab} - \min_{i=1,\ldots,\n} \samplelossfun{i}{\lambdab}.
\end{align*}
We recover the desired inequality by generalizing this bound to hold for all $\lambdab \in \Lset$ as follows.
\begin{align}
\prob{ \max_{\lambdab \in \Lset} \Big( \samplelossfun{\n}{\lambdab} -  \samplelossfun{\m}{\lambdab} \Big) \geq \varepsilon} 
& = \prob{ \bigcup_{\lambdab \in \Lset} \left( \samplelossfun{\n}{\lambdab} - \samplelossfun{\m}{\lambdab} \geq \varepsilon \right) }, \nonumber \\ 
& \leq \sum_{\lambdab \in \Lset} \prob{\samplelossfun{\n}{\lambdab} - \samplelossfun{\m}{\lambdab} \geq \varepsilon}, \label{Eq::GB2} \\
& \leq \sum_{\lambdab \in \Lset} \exp \left(  -\frac{2 \varepsilon^2}{(\frac{1}{\m{}}) (1 - \frac{\m}{\n}) (1 + \frac{\m}{\n}) \Delta(\lambdab,\data_\n{})^2}  \right), \label{Eq::GB3} \\
& \leq |\Lset| \exp \left(  -\frac{2 \varepsilon^2}{(\frac{1}{\m{}}) (1 - \frac{\m}{\n}) (1 + \frac{\m}{\n}) \Delta^{\max}(\Lset,\data_\n{})^2}  \right) \label{Eq::GB4}.
\end{align}
Here, \eqref{Eq::GB2} follows from the union bound, \eqref{Eq::GB3} follows from the Hoeffding Serling inequality, \eqref{Eq::GB4} follows from the fact that $\Delta(\lambdab,\data_\n{}) \leq \Delta^{\max}(\Lset,\data_\n{})$ given that $\lambdab \in \Lset$.
\end{proof}

Theorem \ref{Thm::SubsamplingGeneralizationBound} is derived from a concentration inequality for sampling without replacement, called the Hoeffding-Serfling inequality \citep[see][]{bardenet2015concentration}. The Hoeffding-Serfling inequality is tighter than the classical Hoeffding inequality as it ensures that $\prob{\samplelossfun{\n}{\lambdab} - \samplelossfun{\m}{\lambdab} \geq \epsilon} \rightarrow 0$ as $\m \rightarrow \n$ for all $\epsilon > 0$. Here, $\Delta^{\max}(\Lset,\data_\n{})$ is a normalization term that represents the maximum range of the loss and can be computed quickly using the training dataset $\data_\n{}$ and coefficient set $\Lset$ as shown in Proposition \ref{Prop::LossBounds} in Section \ref{Sec::ChainedUpdates}.

In general machine learning settings, the $|\Lset|$ term in Theorem \ref{Thm::SubsamplingGeneralizationBound} would yield a vacuous bound. In this setting, however, rounding ensures that $\Lset$ contains at most $2^d$ elements, which produces a well-defined bound on the difference between $\samplelossfun{\n}{\lambdab}$ and $\samplelossfun{\m}{\lambdab}$. As a result, Theorem \ref{Thm::SubsamplingGeneralizationBound} can be used to assess the probability that a proposed incumbent update leads to an actual incumbent update (see Corollary \ref{Cor::SubsamplingDelta}). Alternatively, it can be used to set the sample size $\m$ so that an incumbent update on $\data_\m{}$ is likely to yield an incumbent update on $\data_\n{}$. In practice, the bound in Theorem \ref{Thm::SubsamplingGeneralizationBound} can be tightened by recomputing the normalization term $\Delta^{\max}(\Lset, \data_\n{})$ over the course of \LCPA{}. This can be done for each real-valued solution $\rho$, or when the MIP solver restricts the set of feasible coefficients $\Lset$.

\begin{corollary}[Update Probabilities of Rounding Heuristics on Subsampled Data]
\label{Cor::SubsamplingDelta}

Consider a rounding heuristic that takes as input a vector of real-valued coefficients $\rhob = (\rho_1,\ldots,\rho_d) \in \textnormal{conv}(\Lset)$ and outputs a vector of rounded coefficients $\lambdab \in \Lset_{|\rhob}$ where
$$\Lset_{\rhob} = \left(\lambdab \in \Lset ~\big|~ \lambda_j \in \left\{\ceil{\rho_j}, \floor{\rho_j}\right\} \;\text{for}\; j = 1,\ldots,d\right).$$
If we evaluate the rounding heuristic using $\m$ points $\data_\m{} = (\xb_i,y_i)_{i=1}^\m$ drawn without replacement from $\data_\n{} = (\xb_i,y_i)_{i=1}^\n$ and the rounded coefficients $\lambdab \in \Lset_{\rhob}$ attain an objective value $V_\m{}(\lambdab)$, then for any $\delta$, with probability at least $1 - \delta$, we have that $$V_\m{}(\lambdab) < \objvalmax{} - \varepsilon_{\delta} \Longrightarrow V_\n{}(\lambdab) \leq \objvalmax{},$$ where $$\varepsilon_{\delta} = \Delta^{\max}(\Lset_{\rhob}, \data_\n{}) \sqrt{ \frac{ \log(1/\delta) + d\log(2)}{2} \left(\frac{1}{\m{}}\right) \left(1 - \frac{\m{}^2}{\n{}^2}\right)}.$$
\end{corollary}

\begin{proof}[Corollary \ref{Cor::SubsamplingDelta}]
We will first show that for any tolerance $\delta > 0$ that we pick, the prescribed choice of $\varepsilon_\delta$ will ensure that $V_\n{}(\lambdab) - V_\m{}(\lambdab) \leq \varepsilon_\delta$ w.p$.$ at least $1 - \delta$. Restating the result of Theorem \ref{Thm::SubsamplingGeneralizationBound}, we have that for any $\varepsilon > 0$:
\begin{align}
\prob{ \max_{\lambdab \in \Lset} \Big( \samplelossfun{\n}{\lambdab} -  \samplelossfun{\m}{\lambdab} \Big) \geq \varepsilon} &\leq |\Lset| \exp \left(  -\frac{2 \varepsilon^2}{
 (\frac{1}{\m{}}) (1 - \frac{\m}{\n}) (1 + \frac{\m}{\n}) \Delta^{\max}(\Lset,\data_\n{})^2}  \right). \label{Eq::C0}
 \end{align}
Note that $\samplelossfun{\n}{\lambdab} - \samplelossfun{\m}{\lambdab} = V_\n{}(\lambdab) - V_\m{}(\lambdab)$ for any fixed $\lambdab$. In addition, note that the set of rounded coefficients $\Lset(\bm{\rho})$ contains at most $|\Lset(\bm{\rho})| \leq 2^d$ coefficient vectors. Therefore, in this setting, \eqref{Eq::C0} implies that for any $\varepsilon > 0$, 
\begin{align}
\prob{ V_\n(\lambdab) - V_\m(\lambdab) \geq \varepsilon} &\leq    2^d \exp \left(  -\frac{2 \varepsilon^2}{(\frac{1}{\m{}})(1-\frac{\m}{\n})(1+\frac{\m}{\n}) \Delta(\Lset(\bm{\rho}),\data_\n{})^2}  \right). \label{Eq::C1}
 \end{align}
By setting $\varepsilon = \varepsilon_\delta$ and simplifying the terms on the right hand side in \eqref{Eq::C1}, we see that
 \begin{align*}
\prob{ V_\n(\lambdab) - V_\m(\lambdab) \geq \varepsilon_\delta} \leq \delta. 
 \end{align*}
Thus, the prescribed value of $\varepsilon_\delta$ ensures that $V_\n{}(\lambdab) - V_\m{}(\lambdab) \leq \varepsilon_\delta$ w.p. at least $1 - \delta$. 

Since we have set $\varepsilon_\delta$ so that $V_\n{}(\lambdab) - V_\m{}(\lambdab) \leq \varepsilon_\delta$ w.p. at least $1 - \delta$, we now need only to show that any $\lambdab$ satisfying $V_\m{}(\lambdab) < \objvalmax - \varepsilon_\delta$ will also satisfy $V_\n{}(\lambdab) \leq \objvalmax$ to complete the proof. To see this, observe that:
\begin{align}
 V_\n{}(\lambdab) - V_\m{}(\lambdab) & \leq \varepsilon_\delta, \nonumber \\
 V_\n{}(\lambdab) &\leq V_\m{}(\lambdab) + \varepsilon_\delta, \nonumber \\
 V_\n{}(\lambdab) & <  \objvalmax. \label{Eq::C2}
 \end{align}
Here, \eqref{Eq::C2} follows from the fact that $V_\m{}(\lambdab) < \objvalmax - \varepsilon_\delta \implies V_\m{}(\lambdab) + \varepsilon_\delta < \objvalmax$. 
\end{proof}

%% file: appendix_extra_experiments.tex
\begin{table}[htbp]
\centering
\resizebox{\textwidth}{!}{
\begin{tabular}{ll*{3}{C{2.2cm}}*{3}{C{3.0cm}}C{1.6cm}}
\toprule
& 
& \multicolumn{3}{c}{\textsc{Traditional Approaches}}
& \multicolumn{3}{c}{\textsc{Pooled Approaches}}
& \\
%\cmidrule(lr){3-3} 
\cmidrule(lr){3-5} 
\cmidrule(lr){6-8}

\textbf{Dataset} & 
\textbf{Metric}
& \perfheader{\mPLRRd{}}
& \perfheader{\mPLRRsRd{}}
& \perfheader{\mPLRUnit{}}
& \perfheader{\PLRCR{}}
& \perfheader{\PLRCRDCD{}}
& \perfheader{\PLRSRDCD{}}
& \RiskSLIM{}  \\ 

\toprule

\cell{l}{\textds{income}\\$n$~=~32561\\$d$~=~36}
 & \bfcell{l}{test cal\\train cal\\test auc\\train auc}
 & \cell{c}{10.5$\%$\\10.5$\%$\\0.787\\0.787}
 & \cell{c}{19.5$\%$\\19.8$\%$\\0.813\\0.811}
 & \cell{c}{25.4$\%$\\25.8$\%$\\0.814\\0.815}
 & \cell{c}{3.0$\%$\\2.6$\%$\\0.845\\0.848}
 & \cell{c}{3.1$\%$\\2.5$\%$\\0.854\\0.857}
 & \cell{c}{4.2$\%$\\4.4$\%$\\0.832\\0.827}
 & \cell{c}{2.6$\%$\\4.2$\%$\\0.854\\0.860}
 \\ 
   
\midrule

\cell{l}{\textds{mammo}\\$n$~=~961\\$d$~=~14}
 & \bfcell{l}{test cal\\train cal\\test auc\\train auc}
 & \cell{c}{10.5$\%$\\12.2$\%$\\0.832\\0.846}
 & \cell{c}{16.2$\%$\\14.2$\%$\\0.846\\0.852}
 & \cell{c}{8.5$\%$\\7.2$\%$\\0.842\\0.850}
 & \cell{c}{10.9$\%$\\10.1$\%$\\0.845\\0.847}
 & \cell{c}{7.1$\%$\\5.4$\%$\\0.841\\0.847}
 & \cell{c}{7.4$\%$\\5.4$\%$\\0.845\\0.847}
 & \cell{c}{5.0$\%$\\3.1$\%$\\0.843\\0.849}
 \\ 
   
\midrule

\cell{l}{\textds{mushroom}\\$n$~=~8124\\$d$~=~113}
 & \bfcell{l}{test cal\\train cal\\test auc\\train auc}
 & \cell{c}{22.1$\%$\\28.6$\%$\\0.890\\0.890}
 & \cell{c}{8.0$\%$\\5.6$\%$\\0.951\\0.942}
 & \cell{c}{19.9$\%$\\18.3$\%$\\0.969\\0.978}
 & \cell{c}{12.6$\%$\\11.9$\%$\\0.984\\0.984}
 & \cell{c}{4.6$\%$\\4.2$\%$\\0.986\\0.984}
 & \cell{c}{5.4$\%$\\3.3$\%$\\0.978\\0.983}
 & \cell{c}{1.8$\%$\\1.0$\%$\\0.989\\0.990}
 \\ 
   
\midrule

\cell{l}{\textds{rearrest}\\$n$~=~22530\\$d$~=~48}
 & \bfcell{l}{test cal\\train cal\\test auc\\train auc}
 & \cell{c}{7.3$\%$\\4.8$\%$\\0.555\\0.640}
 & \cell{c}{24.2$\%$\\24.3$\%$\\0.692\\0.700}
 & \cell{c}{21.8$\%$\\14.1$\%$\\0.698\\0.698}
 & \cell{c}{5.2$\%$\\1.1$\%$\\0.676\\0.676}
 & \cell{c}{1.4$\%$\\1.1$\%$\\0.676\\0.676}
 & \cell{c}{3.8$\%$\\3.7$\%$\\0.677\\0.682}
 & \cell{c}{2.4$\%$\\2.6$\%$\\0.699\\0.701}
 \\ 
   
\midrule

\cell{l}{\textds{spambase}\\$n$~=~4601\\$d$~=~57}
 & \bfcell{l}{test cal\\train cal\\test auc\\train auc}
 & \cell{c}{15.0$\%$\\18.7$\%$\\0.620\\0.772}
 & \cell{c}{29.5$\%$\\26.8$\%$\\0.875\\0.872}
 & \cell{c}{33.4$\%$\\23.8$\%$\\0.861\\0.876}
 & \cell{c}{26.5$\%$\\25.1$\%$\\0.910\\0.917}
 & \cell{c}{16.3$\%$\\14.6$\%$\\0.913\\0.921}
 & \cell{c}{17.9$\%$\\19.3$\%$\\0.908\\0.926}
 & \cell{c}{11.7$\%$\\12.3$\%$\\0.928\\0.935}
 \\ 

\midrule

\cell{l}{\textds{telemarketing}\\$n$~=~41188\\$d$~=~57}
 & \bfcell{l}{test cal\\train cal\\test auc\\train auc}
 & \cell{c}{2.6$\%$\\0.7$\%$\\0.574\\0.500}
 & \cell{c}{11.2$\%$\\11.3$\%$\\0.700\\0.685}
 & \cell{c}{6.2$\%$\\4.9$\%$\\0.715\\0.685}
 & \cell{c}{1.9$\%$\\1.7$\%$\\0.759\\0.759}
 & \cell{c}{1.3$\%$\\1.1$\%$\\0.760\\0.760}
 & \cell{c}{1.3$\%$\\1.1$\%$\\0.760\\0.760}
 & \cell{c}{1.3$\%$\\1.1$\%$\\0.760\\0.760}
 \\

\bottomrule

\end{tabular}
}
\caption{Summary statistics of risk scores with integer coefficients $\lambda_j \in \{-5,\ldots,5\}$ with model size of $\zeronorm{\lambdab} \leq 5$. Here: \emph{test cal} and \emph{test auc}, which are the 5-CV mean test CAL / AUC; \emph{train cal} and \emph{train auc} which are the CAL and AUC of the final model fit using the entire dataset.}
\label{Table::GeneralizationResults}
\end{table}

%% file: appendix_seizure_prediction.tex
In this appendix, we provide supporting material for the seizure prediction application in Section \ref{Sec::SeizurePrediction}. 

\subsection{List of Input Variables}

In Table \ref{Appendix::Table::SeizureInputVariables}, we list all input variables in the training dataset. %In Appendix \ref{Appendix::SeizurePrediction}, we show risk scores, ROC curves, and reliability diagrams for that were omitted from the main text.
\begin{table}[htbp]
\centering
\resizebox{0.875\linewidth}{!}{
\tiny
\begin{tabular}{@{}l|r@{}}
{\begin{tabular}{lcc}
\toprule
\bfcell{l}{Input Variable} & \bfcell{c}{Values} & \bfcell{c}{Sign} \\ 
\toprule
\textfn{Male} & $\{0,1\}$ & \\ \midrule
\textfn{Female} & $\{0,1\}$ & \\ \midrule
\textfn{PriorSeizure} & $\{0,1\}$ & $+$\\ \midrule
\textfn{PosteriorDominantRhythmPresent} & $\{0,1\}$ & $-$\\ \midrule
\textfn{BriefRhythmicDischarge} & $\{0,1\}$ & $+$\\ \midrule
\textfn{NoReactivityToStimulation} & $\{0,1\}$ & \\ \midrule
%$Epileptiform$ & $\{0,1\}$ & $+$\\ \midrule
%$Discharges$ & $\{0,1\}$ & $+$\\ \midrule
\textfn{EpileptiformDischarges} & $\{0,1\}$ & $+$\\ \midrule
\textfn{SecondaryDXIncludesMentalStatusFirst} & $\{0,1\}$ & \\ \midrule
\textfn{SecondaryDXIncludesCNSInfection} & $\{0,1\}$ & \\ \midrule
\textfn{SecondaryDXIncludesCNSInflammatoryDisease} & $\{0,1\}$ & \\ \midrule
\textfn{SecondaryDXIncludesCNSNeoplasm} & $\{0,1\}$ & \\ \midrule
\textfn{SecondaryDXIncludesHypoxisIschemicEncephalopathy} & $\{0,1\}$ & \\ \midrule
\textfn{SecondaryDXIncludesIntracerebralHemorrhage} & $\{0,1\}$ & \\ \midrule
\textfn{SecondaryDXIncludesIntraventricularHemorrhage} & $\{0,1\}$ & \\ \midrule
\textfn{SecondaryDXIncludesMetabolicEncephalopathy} & $\{0,1\}$ & \\ \midrule
\textfn{SecondaryDXIncludesIschemicStroke} & $\{0,1\}$ & \\ \midrule
\textfn{SecondaryDXIncludesSubarachnoidHemmorage} & $\{0,1\}$ & \\ \midrule
\textfn{SecondaryDXIncludesSubduralHematoma} & $\{0,1\}$ & \\ \midrule
\textfn{SecondaryDXIncludesTraumaticBrainInjury} & $\{0,1\}$ & \\ \midrule
\textfn{SecondaryDXIncludesHydrocephalus} & $\{0,1\}$ & \\ \midrule
\textfn{PatternIsStimulusInducedAny} & $\{0,1\}$ & \\ \midrule
\textfn{PatternIsStimulusInducedBiPD} & $\{0,1\}$ & \\ \midrule
\textfn{PatternIsStimulusInducedGPD} & $\{0,1\}$ & \\ \midrule
\textfn{PatternIsStimulusInducedGRDA} & $\{0,1\}$ & \\ \midrule
\textfn{PatternIsStimulusInducedLPD} & $\{0,1\}$ & \\ \midrule
\textfn{PatternIsStimulusInducedLRDA} & $\{0,1\}$ & \\ \midrule
\textfn{PatternIsSuperImposedAny} & $\{0,1\}$ & $+$\\ \midrule
\textfn{PatternIsSuperImposedBiPD} & $\{0,1\}$ & $+$\\ \midrule
\textfn{PatternIsSuperImposedGPD} & $\{0,1\}$ & $+$\\ \midrule
\textfn{PatternIsSuperImposedGRDA} & $\{0,1\}$ & $+$\\ \midrule
\textfn{PatternIsSuperImposedLPD} & $\{0,1\}$ & $+$\\ \midrule
\textfn{PatternIsSuperImposedLRDA} & $\{0,1\}$ & $+$\\ \midrule
\textfn{PatternsInclude\_BiPD} & $\{0,1\}$ & $+$\\ \midrule
\textfn{PatternsInclude\_GPD} & $\{0,1\}$ & $+$\\ \midrule
\textfn{PatternsInclude\_GRDA} & $\{0,1\}$ & $+$\\ \midrule
\textfn{PatternsInclude\_LPD} & $\{0,1\}$ & $+$\\ \midrule
\textfn{PatternsInclude\_LRDA} & $\{0,1\}$ & $+$\\ \midrule
\textfn{PatternsInclude\_GRDA\_or\_GPD} & $\{0,1\}$ & $+$\\ \midrule
\textfn{PatternsInclude\_BiPD\_or\_LRDA\_or\_LPD} & $\{0,1\}$ & $+$\\ \midrule
\textfn{MaxFrequencyAnyPattern} & $\{0.0,0.5,\ldots,3.0\}$ & $+$\\ \midrule
\textfn{MaxFrequencyBiPD} & $\{0.0,0.5,\ldots,3.0\}$ & $+$\\ \midrule
\textfn{MaxFrequencyGPD} & $\{0.0,0.5,\ldots,3.0\}$ & $+$\\ \midrule
\textfn{MaxFrequencyLPD} & $\{0.0,0.5,\ldots,3.0\}$ & $+$\\ \midrule
\textfn{MaxFrequencyLRDA} & $\{0.0,0.5,\ldots,3.0\}$ & $+$\\ \bottomrule
\end{tabular}
}
& 
{\begin{tabular}{lcc}
\toprule
\bfcell{l}{Input Variable} & \bfcell{c}{Values} & \bfcell{c}{Sign} \\
\toprule
%\textfn{MaxFrequencyAnyPattern} & $\{0.0,0.5,1.0,1.5,2.0,2.5,3.0\}$ & $+$\\ \midrule
\textfn{MaxFrequencyAnyPattern\, =  \, 0.0Hz} & $\{0,1\}$ & \\ \midrule
\textfn{MaxFrequencyAnyPattern\,$\geq$\,0.5Hz} & $\{0,1\}$ & $+$\\ \midrule
\textfn{MaxFrequencyAnyPattern\,$\geq$\,1.0Hz} & $\{0,1\}$ & $+$\\ \midrule
\textfn{MaxFrequencyAnyPattern\,$\geq$\,1.5Hz} & $\{0,1\}$ & $+$\\ \midrule
\textfn{MaxFrequencyAnyPattern\,$\geq$\,2.0Hz} & $\{0,1\}$ & $+$\\ \midrule
\textfn{MaxFrequencyAnyPattern\,$\geq$\,2.5Hz} & $\{0,1\}$ & $+$\\ \midrule
\textfn{MaxFrequencyAnyPattern\,$\geq$\,3.0Hz} & $\{0,1\}$ & $+$\\ \midrule
%
%\textfn{MaxFrequencyBiPD} & $\{0.0,0.5,1.0,1.5,2.0,2.5,3.0\}$ & $+$\\ \midrule
\textfn{MaxFrequencyBiPD\,=\,0.0} & $\{0,1\}$ & \\ \midrule
\textfn{MaxFrequencyBiPD\,$\geq$\,0.5Hz} & $\{0,1\}$ & $+$\\ \midrule
\textfn{MaxFrequencyBiPD\,$\geq$\,1.0Hz} & $\{0,1\}$ & $+$\\ \midrule
\textfn{MaxFrequencyBiPD\,$\geq$\,1.5Hz} & $\{0,1\}$ & $+$\\ \midrule
\textfn{MaxFrequencyBiPD\,$\geq$\,2.0Hz} & $\{0,1\}$ & $+$\\ \midrule
\textfn{MaxFrequencyBiPD\,$\geq$\,2.5Hz} & $\{0,1\}$ & $+$\\ \midrule
\textfn{MaxFrequencyBiPD\,$\geq$\,3.0Hz} & $\{0,1\}$ & $+$\\ \midrule
%
%\textfn{MaxFrequencyGPD} & $\{0.0,0.5,1.0,1.5,2.0,2.5,3.0\}$ & $+$\\ \midrule
\textfn{MaxFrequencyGPD\,=\,0.0} & $\{0,1\}$ & \\ \midrule
\textfn{MaxFrequencyGPD\,$\geq$\,0.5Hz} & $\{0,1\}$ & $+$\\ \midrule
\textfn{MaxFrequencyGPD\,$\geq$\,1.0Hz} & $\{0,1\}$ & $+$\\ \midrule
\textfn{MaxFrequencyGPD\,$\geq$\,1.5Hz} & $\{0,1\}$ & $+$\\ \midrule
\textfn{MaxFrequencyGPD\,$\geq$\,2.0Hz} & $\{0,1\}$ & $+$\\ \midrule
\textfn{MaxFrequencyGPD\,$\geq$\,2.5Hz} & $\{0,1\}$ & $+$\\ \midrule
\textfn{MaxFrequencyGPD\,$\geq$\,3.0Hz} & $\{0,1\}$ & $+$\\ \midrule
%
%\textfn{MaxFrequencyGRDA} & $\{0.0,0.5,1.0,1.5,2.0,2.5,3.0\}$ & $+$\\ \midrule
\textfn{MaxFrequencyGRDA\,=\,0.0} & $\{0,1\}$ & \\ \midrule
\textfn{MaxFrequencyGRDA\,$\geq$\,0.5Hz} & $\{0,1\}$ & $+$\\ \midrule
\textfn{MaxFrequencyGRDA\,$\geq$\,1.0Hz} & $\{0,1\}$ & $+$\\ \midrule
\textfn{MaxFrequencyGRDA\,$\geq$\,1.5Hz} & $\{0,1\}$ & $+$\\ \midrule
\textfn{MaxFrequencyGRDA\,$\geq$\,2.0Hz} & $\{0,1\}$ & $+$\\ \midrule
\textfn{MaxFrequencyGRDA\,$\geq$\,2.5Hz} & $\{0,1\}$ & $+$\\ \midrule
\textfn{MaxFrequencyGRDA\,$\geq$\,3.0Hz} & $\{0,1\}$ & $+$\\ \midrule
%
%\textfn{MaxFrequencyLPD} & $\{0.0,0.5,1.0,1.5,2.0,2.5,3.0\}$ & $+$\\ \midrule
\textfn{MaxFrequencyLPD\,=\,0.0} & $\{0,1\}$ & \\ \midrule
\textfn{MaxFrequencyLPD\,$\geq$\,0.5Hz} & $\{0,1\}$ & $+$\\ \midrule
\textfn{MaxFrequencyLPD\,$\geq$\,1.0Hz} & $\{0,1\}$ & $+$\\ \midrule
\textfn{MaxFrequencyLPD\,$\geq$\,1.5Hz} & $\{0,1\}$ & $+$\\ \midrule
\textfn{MaxFrequencyLPD\,$\geq$\,2.0Hz} & $\{0,1\}$ & $+$\\ \midrule
\textfn{MaxFrequencyLPD\,$\geq$\,2.5Hz} & $\{0,1\}$ & $+$\\ \midrule
\textfn{MaxFrequencyLPD\,$\geq$\,3.0Hz} & $\{0,1\}$ & $+$\\ \midrule
%
%\textfn{MaxFrequencyLRDA} & $\{0.0,0.5,1.0,1.5,2.0,2.5,3.0\}$ & $+$\\ \midrule
\textfn{MaxFrequencyLRDA\,=\,0.0} & $\{0,1\}$ & \\ \midrule
\textfn{MaxFrequencyLRDA\,$\geq$\,0.5Hz} & $\{0,1\}$ & $+$\\ \midrule
\textfn{MaxFrequencyLRDA\,$\geq$\,1.0Hz} & $\{0,1\}$ & $+$\\ \midrule
\textfn{MaxFrequencyLRDA\,$\geq$\,1.5Hz} & $\{0,1\}$ & $+$\\ \midrule
\textfn{MaxFrequencyLRDA\,$\geq$\,2.0Hz} & $\{0,1\}$ & $+$\\ \midrule
\textfn{MaxFrequencyLRDA\,$\geq$\,2.5Hz} & $\{0,1\}$ & $+$\\ \midrule
\textfn{MaxFrequencyLRDA\,$\geq$\,3.0Hz} & $\{0,1\}$ & $+$\\ 
\bottomrule \\[2.25em]
\end{tabular}
}
\end{tabular}
}
\caption{Names, values, and sign constraints for input variables in the \textds{seizure} dataset.}
\label{Appendix::Table::SeizureInputVariables}
\end{table}

\clearpage
\subsection{List of Operational Constraints}
\label{Appendix::ListOfOperationalConstraints}

\vspace{0.5em}
\noindent\textbf{No Redundant Categorical Variables}
\begin{enumerate}[leftmargin=1.9em, itemsep=0.1em]
\item Use either \textfn{Male} or \textfn{Female}.
%\item Use either \textfn{EpileptiformDischarge} or any one of (\textfn{Epileptiform}, \textfn{Discharges}).
\item Use either \textfn{PatternsInclude\_GRDA\_or\_GPD} or any one of \\(\textfn{PatternsInclude\_GRDA}, \textfn{PatternsInclude\_GPD}).
\item Use either \textfn{PatternsInclude\_BiPD\_or\_LRDA\_or\_LPD} or any one of \\(\textfn{PatternsInclude\_BiPD}, \textfn{PatternsInclude\_LRDA}, \textfn{PatternsInclude\_LPD}).
\item Use either \textfn{MaxFrequencyAnyPattern$\,= \,0.0$} or \textfn{MaxFrequencyAnyPattern$\,\geq\,0.5$} or neither.
\item Use either \textfn{MaxFrequencyLPD$\,= \,0.0$} or \textfn{MaxFrequencyLPD$\,\geq\,0.5$} or neither.
\item Use either \textfn{MaxFrequencyGPD$\,= \,0.0$} or \textfn{MaxFrequencyGPD$\,\geq\,0.5$} or neither.
\item Use either \textfn{MaxFrequencyGRDA$\,= \,0.0$} or \textfn{MaxFrequencyGRDA$\,\geq\,0.5$} or neither.
\item Use either \textfn{MaxFrequencyBiPD$\,= \,0.0$} or \textfn{MaxFrequencyBiPD$\,\geq\,0.5$} or neither.
\item Use either \textfn{MaxFrequencyLRDA$\,= \,0.0$} or \textfn{MaxFrequencyLRDA$\,\geq\,0.5$} or neither.
\end{enumerate}
\noindent\textbf{Frequency in Continuous Encoding or Thresholded Encoding}
\begin{enumerate}[leftmargin=1.9em,itemsep=0.1em,resume]
\item Choose between \textfn{MaxFrequencyAnyPattern} or \\(\textfn{MaxFrequencyAnyPattern$\,=\,0.0$} $\ldots$ \textfn{MaxFrequencyAnyPattern$\,\geq\,3.0$}).

\item Choose between  \textfn{MaxFrequencyGPD} or \\(\textfn{MaxFrequencyGPD$\,=\,0.0$} $\ldots$ \textfn{MaxFrequencyGPD$\,\geq\,3.0$}).

\item Choose between  \textfn{MaxFrequencyLPD} or \\(\textfn{MaxFrequencyLPD$\,=\,0.0$} $\ldots$ \textfn{MaxFrequencyLPD$\,\geq\,3.0$}).

\item Choose between \textfn{MaxFrequencyGRDA} or \\(\textfn{MaxFrequencyGRDA$\,=\,0.0$} $\ldots$ \textfn{MaxFrequencyGRDA$\,\geq\,3.0$}).

\item Choose between \textfn{MaxFrequencyBiPD} or \\(\textfn{MaxFrequencyBiPD$\,=\,0.0$} $\ldots$ \textfn{MaxFrequencyBiPD$\,\geq\,3.0$}).

\item Choose between \textfn{MaxFrequencyLRDA} or \\(\textfn{MaxFrequencyLRDA$\,=\,0.0$} $\ldots$ \textfn{MaxFrequencyLRDA$\,\geq\,3.0$}).

\end{enumerate}
\noindent\textbf{Limited \# of Thresholds for Thresholded Variables}
\begin{enumerate}[leftmargin=1.9em,itemsep=0.1em,resume]
\item Use at most 2 of: \textfn{MaxFrequencyAnyPattern$\,=\,0.0$}, \textfn{MaxFrequencyAnyPattern$\,\geq\,0.5$} $\ldots$ \textfn{MaxFrequencyAnyPattern$\,\geq\,3.0$}.

\item Use at most 2 of: \textfn{MaxFrequencyLPD$\,=\,0.0$}, \textfn{MaxFrequencyLPD$\,\geq\,0.5$}$\ldots$ \textfn{MaxFrequencyLPD$\,\geq\,3.0$}.

\item Use at most 2 of: \textfn{MaxFrequencyGPD$\,=0.0$}, \textfn{MaxFrequencyGPD$\,\geq\,0.5$}$\ldots$ \textfn{MaxFrequencyGPD$\,\geq\,3.0$}.

\item Use at most 2 of: \textfn{MaxFrequencyGRDA$=0.0$}, \textfn{MaxFrequencyGRDA$\,\geq\,0.5$}$\ldots$ \textfn{MaxFrequencyGRDA$\,\geq\,3.0$}.

\item Use at most 2 of: \textfn{MaxFrequencyBiPD$\,=0.0$}, \textfn{MaxFrequencyBiPD$\,\geq\,0.5$}$\ldots$ \textfn{MaxFrequencyBiPD$\,\geq\,3.0$}.

\item Use at most 2 of: \textfn{MaxFrequencyLRDA$\,=\,0.0$}, \textfn{MaxFrequencyLRDA$\,\geq\,0.5$} $\ldots$\textfn{MaxFrequencyLRDA$\,\geq\,3.0$}.

\end{enumerate}
\noindent\textbf{Any cEEG Pattern or Specific cEEG Patterns}
\begin{enumerate}[leftmargin=1.9em,itemsep=0.1em,resume]
\item Use either \textfn{PatternIsStimulusInducedAny} or any of (\textfn{PatternIsStimulusInducedBiPD}, \textfn{PatternIsStimulusInducedGRDA}, \textfn{PatternIsStimulusInducedGPD}, \textfn{PatternIsStimulusInducedLPD}, \textfn{PatternIsStimulusInducedLRDA}) or none of the variables.
\item Use either \textfn{PatternIsSuperImposed} or any of (\textfn{PatternIsSuperImposedBiPD}, \textfn{PatternIsSuperImposedGPD}  \textfn{PatternIsSuperImposedGRDA}, \textfn{PatternIsSuperImposedLPD},  \textfn{PatternIsSuperImposedLRDA}), or none of the variables. 
\item Use either \textfn{MaxFrequencyAnyPattern} (or its thresholded versions) or any of \textfn{MaxFrequencyBiPD}, \textfn{MaxFrequencyGRDA}, \textfn{MaxFrequencyGPD}, \textfn{MaxFrequencyLPD}, \textfn{MaxFrequencyLRDA}, (or their thresholded versions), or none of the variables.
\end{enumerate}

\subsection{Additional Experimental Results}

\begin{table}[htbp]
\centering
\renewcommand{\arraystretch}{1.2}\resizebox{\textwidth}{!}{%
\begin{tabular}{l*{6}{r}}

\toprule
& 
\multicolumn{2}{c}{\textsc{Training Requirements}} & 
\multicolumn{4}{c}{\textsc{\% of Instances That Satisfy Constraints on}}\\
\cmidrule(lr){2-3} \cmidrule(lr){4-7}

\textbf{Method}
& \bfcell{r}{\# Instances}
& \bfcell{r}{\# Models}
& \bfcell{r}{Monotonicity}
& \bfcell{r}{Model Size}
& \bfcell{r}{Operational}
& \bfcell{r}{All Constraints}

\\ \midrule

\RiskSLIM{}
& 1 
& 6
& 100$\%$
& 100$\%$
& 100$\%$
& 100$\%$

\\ \midrule

\PLRCR{}
& 1,100
& 33,000
& 100$\%$
& 22$\%$
& 20$\%$
& 20$\%$

\\ \midrule

\PLRCRDCD{}
& 1,100
& 33,000
& 100$\%$
& 22$\%$
& 20$\%$
& 20$\%$

\\ \midrule

\PLRLR{}
& 1,100
& 33,000
& 100$\%$
& 9$\%$
& 5$\%$
& 2$\%$

\\ \midrule

\PLRLRDCD{}
& 1,100
& 33,000
& 23$\%$
& 9$\%$
& 5$\%$
& 5$\%$

\\ \midrule

\PLRSR{}
& 1,100
& 33,000
& 100$\%$
& 10$\%$
& 7$\%$
& 4$\%$

\\ \midrule

\PLRSRDCD{}
& 1,100
& 33,000
& 98$\%$
& 10$\%$
& 8$\%$
& 4$\%$

\\ \bottomrule
\end{tabular}
}
\caption{Training requirements and constraint violations for the methods in Table \ref{Table::DemoPerformanceTable}. Each instance is a unique combination of free parameters. 
\# models represents the total number of models that we must train to (1) choose parameters of the final risk score \emph{and} (2) pair this model an unbiased estimate of performance. We need to train 33K for other methods since they require nested cross validation.}
\label{Table::DemoFeasibilityTable}
\end{table}